\documentclass{article}

\usepackage{microtype}
\usepackage{graphicx}
\usepackage{subfigure}
\usepackage{booktabs} 
\usepackage{microtype}
\usepackage{graphicx}
\usepackage{subfigure}
\usepackage{booktabs} 

\usepackage{hyperref}
\usepackage{tcolorbox}
\usepackage{longtable}
\usepackage{hyperref}
\usepackage{amsmath}
\usepackage{amssymb}
\usepackage{mathtools}
\usepackage{amsthm}
\usepackage{enumitem}
\usepackage{pifont}    
\usepackage{makecell}
\usepackage{algorithm}
\usepackage{algorithmic}
\usepackage{textcomp}
\usepackage{csquotes}
\usepackage{microtype}
\usepackage{graphicx}
\usepackage{booktabs}
\usepackage{hyperref}
\usepackage{amsmath}
\usepackage{amssymb}
\usepackage{mathtools}
\usepackage{amsthm}
\usepackage[capitalize,noabbrev]{cleveref}
\usepackage[title]{appendix}
\usepackage{subcaption}
\usepackage{wrapfig}
\usepackage[utf8]{inputenc} 
\usepackage[T1]{fontenc}    
\usepackage{url}            
\usepackage{xspace}
\usepackage[export]{adjustbox}
\usepackage{amsfonts}       
\usepackage{nicefrac}       
\usepackage{microtype}      
\usepackage{wasysym}
\usepackage{enumitem}
\usepackage[american]{babel}
\usepackage{multirow}
\usepackage{adjustbox}
\usepackage{hyperref}       
\usepackage{booktabs}       
\usepackage{tikz} 
\usepackage{amssymb}
\usepackage{amsmath,amsfonts}
\usepackage{amsopn}
\usepackage{bm} 
\usepackage{multirow}
\usepackage{multicol}
\usepackage{flushend}
\usepackage{tabularx}
\usepackage{longtable}
\usepackage{xcolor}
\definecolor{lightblue}{RGB}{173,216,230}
\definecolor{mintgreen}{RGB}{152,251,152}
\definecolor{cherrypink}{RGB}{225,127,88}
\usepackage[textsize=tiny]{todonotes}
\definecolor{darkgreen}{rgb}{0.0, 0.4, 0.26}  
\definecolor{darkred}{rgb}{0.7, 0.15, 0.15}   
\newcommand{\greencheck}{{\color{darkgreen}\ding{51}}}  
\newcommand{\redx}{{\color{darkred}\ding{55}}}  
\newcommand{\name}{\textsc{Beta}}
\usepackage[capitalize,noabbrev]{cleveref}

\usepackage[accepted]{ICML/icml2025}

\usepackage{amsmath}
\usepackage{amssymb}
\usepackage{mathtools}
\usepackage{amsthm}

\usepackage[capitalize,noabbrev]{cleveref}

\theoremstyle{plain}

\theoremstyle{definition}

\theoremstyle{remark}

\usepackage[textsize=tiny]{todonotes}

\icmltitlerunning{TabPFN Unleashed: A Scalable and Effective Solution to Tabular Classification Problems}

\begin{document}

\twocolumn[
\icmltitle{TabPFN Unleashed: A Scalable and Effective Solution to \\ Tabular Classification Problems}

\begin{icmlauthorlist}
\icmlauthor{Si-Yang Liu}{yyy}
\icmlauthor{Han-Jia Ye}{yyy}
\end{icmlauthorlist}

\icmlaffiliation{yyy}{State Key Laboratory for Novel Software Technology, Nanjing University}

\icmlcorrespondingauthor{Han-Jia Ye}{yehj@lamda.nju.edu.cn}

\icmlkeywords{Machine Learning, ICML}

\vskip 0.3in
]

\printAffiliationsAndNotice{\icmlEqualContribution} 

\begin{abstract}
TabPFN has emerged as a promising in-context learning model for tabular data, capable of directly predicting the labels of test samples given labeled training examples. It has demonstrated competitive performance, particularly on small-scale classification tasks. However, despite its effectiveness, TabPFN still requires further refinement in several areas, including handling high-dimensional features, aligning with downstream datasets, and scaling to larger datasets.
In this paper, we revisit existing variants of TabPFN and observe that most approaches focus either on reducing bias or variance, often neglecting the need to address the other side, while also increasing inference overhead. 
To fill this gap, we propose~\name~(\textbf{B}agging and \textbf{E}ncoder-based Fine-tuning for \textbf{T}abPFN \textbf{A}daptation), a novel and effective method designed to \textit{minimize both bias and variance}. To reduce bias, we introduce a lightweight encoder to better align downstream tasks with the pre-trained TabPFN. By increasing the number of encoders in a lightweight manner,~\name~mitigate variance, thereby further improving the model’s performance. Additionally, bootstrapped sampling is employed to further reduce the impact of data perturbations on the model, all while maintaining computational efficiency during inference. Our approach enhances TabPFN’s ability to handle high-dimensional data and scale to larger datasets. Experimental results on over 200 benchmark classification datasets demonstrate that~\name~either outperforms or matches state-of-the-art methods.
\end{abstract}

\section{Introduction}
\label{sec:intro} 
Tabular data is one of the most widely used data formats across various domains, including finance~\cite{Financial_SVM}, healthcare~\cite{HassanAHK20}, e-commerce~\cite{nederstigt2014floppies}, and medical analysis~\cite{schwartz2007drug,subasi2012medical}. Despite its ubiquity, modeling tabular data with deep learning methods remains a challenge due to its heterogeneous nature~\cite{BorisovLSHPK24TabularSurvey}. 
Yet recent advancements have led to the development of tabular foundation models~\cite{WhyTFM}~, such as TabPFN (Tabular Prior-Fitted Networks)~\cite{Hollmann2022TabPFN,hollmann2025tabpfn}. TabPFN operates in two stages: \textit{pre-training} and \textit{inference}. During the pre-training stage, the model is pre-trained on a diverse set of synthetic datasets. In the inference stage, given a new task and a set of labeled examples as a ``prompt,'' TabPFN directly predicts the labels of test samples using in-context learning, without requiring further parameter updates. This approach enables TabPFN to achieve performance comparable to or even surpass tree-based methods, particularly on small tabular datasets~\cite{McElfreshKVCRGW23when}. 

TabPFN has shown potential across a wide range of applications, including tabular data generation~\cite{TabPFGen}, data augmentation~\cite{TabMDA}, and time series forecasting~\cite{TabPFN-TS}. These use cases highlight the versatility of TabPFN, positioning it as a model worthy of further exploration. However, alongside application-driven studies, there is a growing interest in improving TabPFN's performance from various perspectives~\cite{TuneTables,LocalPFN,MixturePFN,TabForestPFN,MaTabDPT}. While previous research has reported performance improvements, the underlying reasons driving these gains remain unclear.  
These advancements are often fragmented, with each approach focusing on a specific aspect, and some methods even sacrifice efficiency for enhanced performance, without a comprehensive understanding of how to improve TabPFN systematically.

In this paper, we adopt the bias-variance decomposition framework introduced by~\citet{SF_PFN} to analyze the generalization error of TabPFN and its variants. This framework allows us to revisit and categorize existing methods, revealing that performance improvements typically arise from addressing either bias or variance. However, these approaches often neglect the other aspect, leading to suboptimal performance.  
Therefore, there is a need for a method that simultaneously addresses both bias and variance to improve performance. 
To this end, we propose a novel, efficient, and scalable approach,~\name. Our method enhances TabPFN’s performance by introducing a \textit{fine-tuning} stage, enabling parameter-efficient adaptation that aligns the downstream dataset distribution with the pre-trained TabPFN to mitigate bias. To further reduce variance~\name~maps raw data into multiple latent spaces and employs computationally efficient bootstrapped sampling during inference. Beyond performance improvements,~\name~offers a lightweight and scalable solution that effectively handles high-dimensional data and large datasets while maintaining inference efficiency.\looseness=-1


\name~improves performance by refining both the fine-tuning and inference stages to 
reduce bias and variance. In the fine-tuning phase, we employ a lightweight encoder module as an input feature adapter, which transforms datasets with arbitrary dimensionality into multiple fixed-dimensional representations, thereby naturally enabling dimensionality reduction. To further enhance generalization, we integrate Batch Ensemble \cite{WenTB20BatchEnsemble,Yury2024TabM} to increase the number of encoders in a lightweight manner, which introduces diversity in learned representations and reduces variance. These enhancements enable \name~to improve robustness and scalability, ensuring better adaptation to downstream datasets. 
In the inference phase, we introduce bootstrapped sampling, a technique that has been largely overlooked in previous TabPFN variants. By generating multiple subsets of the dataset as support sets for the context composition, \name~reduces variance and improves robustness. Furthermore, \name~seamlessly integrates with Error-Correcting Output Codes (ECOC) to effectively handle multiclass classification tasks with more than 10 classes.

Experimental results on multiple benchmark datasets, including over 200 classification tasks, demonstrate that our method significantly improves TabPFN's performance. We describe our main contributions below.
\begin{enumerate}[noitemsep,topsep=0pt,leftmargin=*]
    \item We introduce an adaptation method for TabPFN that addresses key limitations related to dataset size, high-dimensional features, and multiclass classification tasks.
    \item By analyzing the generalization error of existing TabPFN variants through bias-variance decomposition and experiments on real-world datasets, we developed~\name, a method that effectively mitigates both bias and variance. 
    \item We achieve state-of-the-art performance on the largest benchmark datasets to date, demonstrating the robustness and scalability of our method for real-world tabular tasks.
\end{enumerate}

\noindent{\bf Remark}.
We have noticed that the latest release of TabPFN-v2~\cite{hollmann2025tabpfn}~has partially alleviated some of the limitations previously discussed. Specifically, TabPFN-v2 incorporates design improvements that enable it to handle larger datasets and more features. However, it is important to highlight that TabPFN-v2 is a concurrent work, and while it partially mitigates these limitations, it does not fully resolve the challenges posed by dataset size and feature count. Thus, many of the improvements proposed in this paper are general enhancements that can complement TabPFN-v2 and potentially further its applicability to a broader range of tasks.\looseness=-1

\section{Preliminary}
\label{sec:preliminary}

In this section, we provide a brief overview of TabPFN and analyze its properties, while also revisiting existing variants.\looseness=-1

\subsection{TabPFN}
We consider a tabular dataset consisting of $N$ examples and $d$ features. Each instance $\boldsymbol{x}_i \in \mathbb{R}^d$ is represented by $d$ feature values, where $\boldsymbol{x}_{i,j}$ denotes the $j$-th feature of instance $\boldsymbol{x}_i$. These features can be numerical ($\boldsymbol{x}_{i,j}^{\text{num}} \in \mathbb{R}$) or categorical ($\boldsymbol{x}_{i,j}^{\text{cat}}$), with categorical values often encoded as integers. Each instance is associated with a label $\boldsymbol{y}_i$, where $\boldsymbol{y}_i \in [C] = \{1, \dots, C\}$ for classification task, and $\boldsymbol{y}_i \in \mathbb{R}$ for regression task. Given a training dataset, $D_{\text{train}} = \{(\boldsymbol{x}_{\text{train}}^{(i)}, \boldsymbol{y}_{\text{train}}^{(i)})\}_{i=1}^{N_{\text{train}}}$, and test samples, $\boldsymbol{X}_{\text{test}} = [\boldsymbol{x}_{\text{test}}^{(i)}]_{i=1}^{N_{\text{test}}}$, the goal is to predict the corresponding labels, $\boldsymbol{Y}_{\text{test}} = [\boldsymbol{y}_{\text{test}}^{(i)}]_{i=1}^{N_{\text{test}}}$, as accurately as possible.\looseness=-1

TabPFN follows a two-stage process: a pre-training stage, where the model is trained on synthetic datasets by minimizing the discrepancy between the predicted label of the test instance and its true label, and an inference stage, where it directly predicts the labels of test samples given a set of labeled training examples.

\begin{figure}[t]
    \centering
    \includegraphics[width=0.48\textwidth]{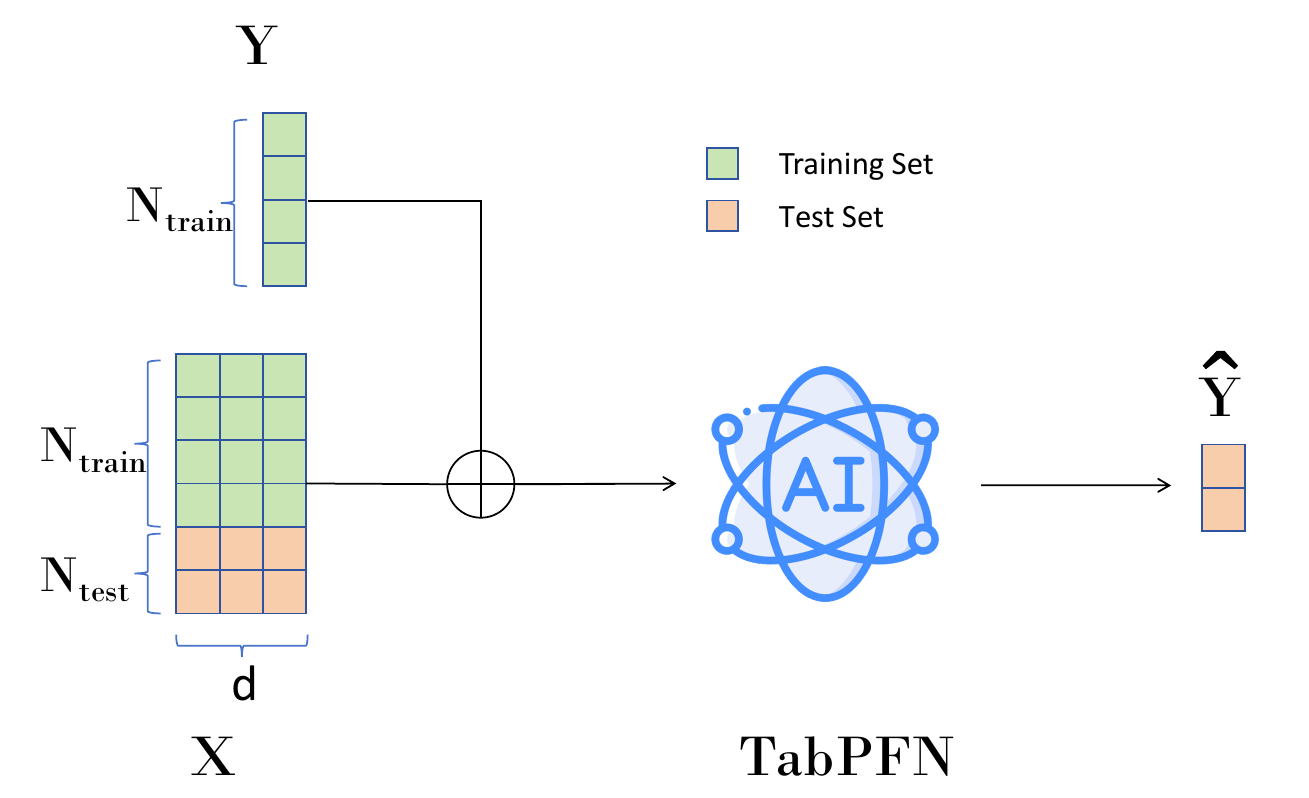}
    \vspace{-5mm}
    \caption{The prediction process of TabPFN. The training set and test set are concatenated, with the training labels added to the input sequence. TabPFN then performs a single forward pass to generate predictions for all test samples.}
    \label{fig:TabPFN}
    \vspace{-5mm}
\end{figure}

Following pre-training, TabPFN takes the entire training dataset \(D_{\text{train}}=(\boldsymbol{X}_{\text{train}} ,\boldsymbol{y}_{\text{train}})\) and the features of query points \(\boldsymbol{X}_{\text{q}}\) as context to make predictions. This is done using \textbf{PFN-Style Batching}~\cite{Hollmann2022TabPFN}, where \(N_{\text{test}}\) test samples are batched into a single ``prompt,'' as illustrated in~\autoref{fig:TabPFN}. 
Since TabPFN is trained with a fixed input dimensionality, typically set to a predefined value \(d_{\max}\) (e.g., 100), datasets with fewer features (\(d < d_{\max}\)) need to be extended via zero-padding before being processed. Formally, given an input instance \(\boldsymbol{x}_i \in \mathbb{R}^d\), the padded input \(\tilde{\boldsymbol{x}}_i \in \mathbb{R}^{d_{\max}}\) is obtained as:
\begin{equation}
\tilde{\boldsymbol{x}}_i = \left[ \boldsymbol{x}_i, \boldsymbol{0} \right] \in \mathbb{R}^{d_{\max}},
\end{equation}
where \(\boldsymbol{0} \in \mathbb{R}^{d_{\max} - d}\) represents the zero-padding applied to match the required input dimensionality. However, if a dataset exceeds this limit (\(d > d_{\max}\)), TabPFN cannot directly process such datasets, as its architecture is not designed to accommodate higher-dimensional inputs.

TabPFN outputs a probability distribution over possible labels \(\boldsymbol{y}_{\text{q}} \in \{1, \dots, C\}\). Specifically, let \( q_\theta \) denote the logits produced by TabPFN, where \( \theta \) represents the parameters of the pre-trained model. The posterior predictive distribution is given by:
\vspace{-3mm}
\begin{equation} 
\label{eq:tabpfn_post_pred}
p_\theta(\boldsymbol{y}_{\text{q}} \mid \boldsymbol{X}_{\text{q}}, D_{\text{train}}) =  \frac{\exp(q_\theta(\tilde{\boldsymbol{X}}_{\text{q}}, \tilde{D}_{\text{train}})_{[\boldsymbol{y}_{\text{q}}]})}{\sum_{c=1}^C \exp(q_\theta(\tilde{\boldsymbol{X}}_{\text{q}}, \tilde{D}_{\text{train}})_{[c]})},  
\end{equation}

\vspace{-2mm}
where \(\tilde{D}_{\text{train}}\) denotes the zero-padded training dataset, acting as the support set for the context composition. Since TabPFN performs inference directly, the model typically applies multiple rounds of feature shuffling on the original features, followed by prediction. The final prediction is obtained by averaging the results from these multiple inferences. For more details about how the model processes inputs, please refer to Appendix~\ref{appendix:details_of_pfn}.
\subsection{TabPFN Variants} 
Existing research to improve the performance of TabPFN has primarily focused on two different strategies, stemming from its dependency on $q_{\theta}$ and \(D_{\text{train}}\) in~\autoref{eq:tabpfn_post_pred}. One approach optimizes context selection, refining \(D_{\textbf{train}}\) to provide more informative support examples. The other concentrates on fine-tuning TabPFN, improving \(q_{\theta}\) to better adapt to downstream tasks. These two strategies offer complementary solutions for enhancing TabPFN's overall performance. 
\begin{figure*}[h]
\centering
\begin{minipage}{0.32\textwidth}
    \centering
    \includegraphics[width=\linewidth]{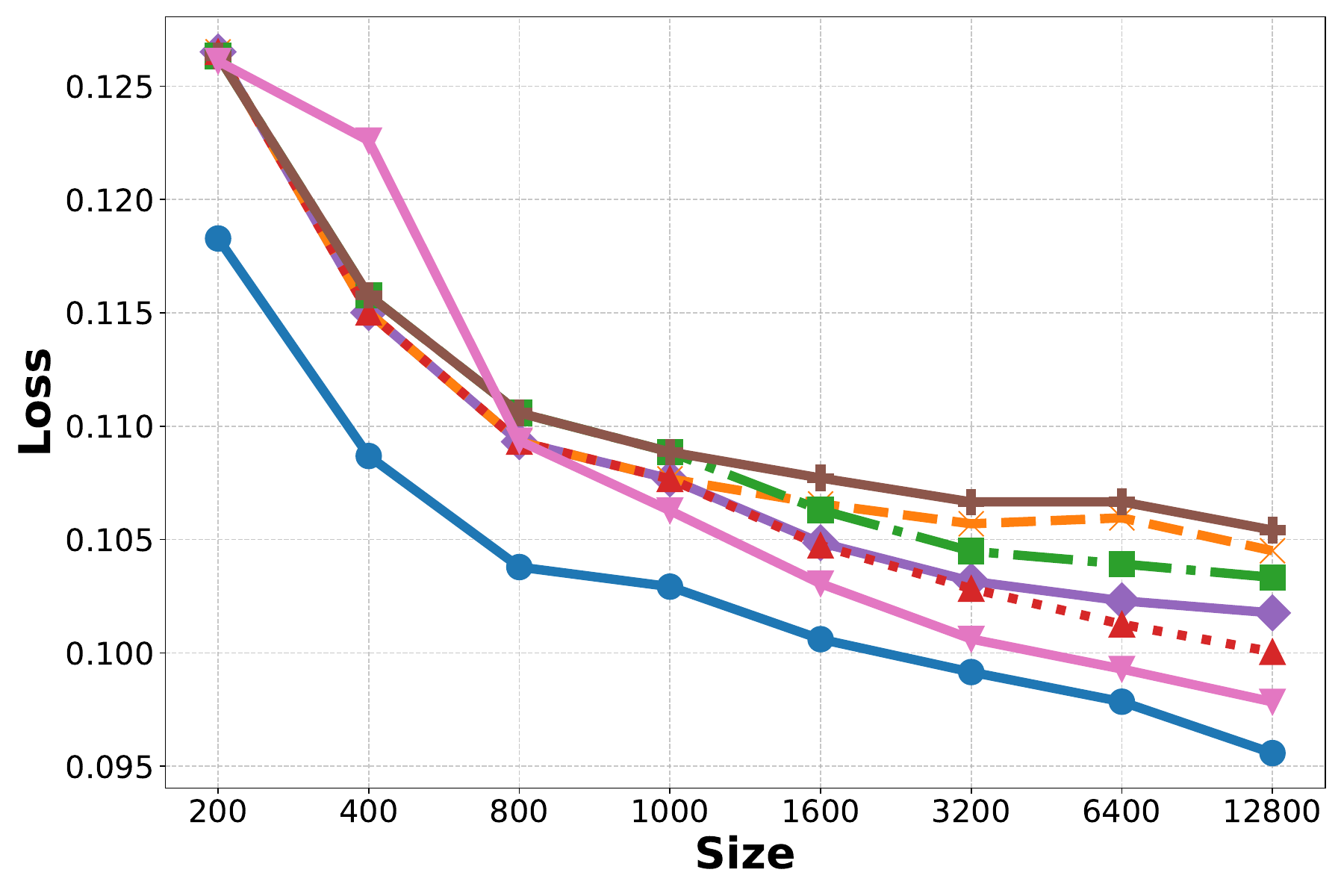}
    {\scriptsize \mbox{(a) {\textit{Generalization error} on the Adult Dataset.}}}
    \label{fig:adult_generalization_error}
\end{minipage}%
\begin{minipage}{0.32\textwidth}
    \centering
    \includegraphics[width=\linewidth]{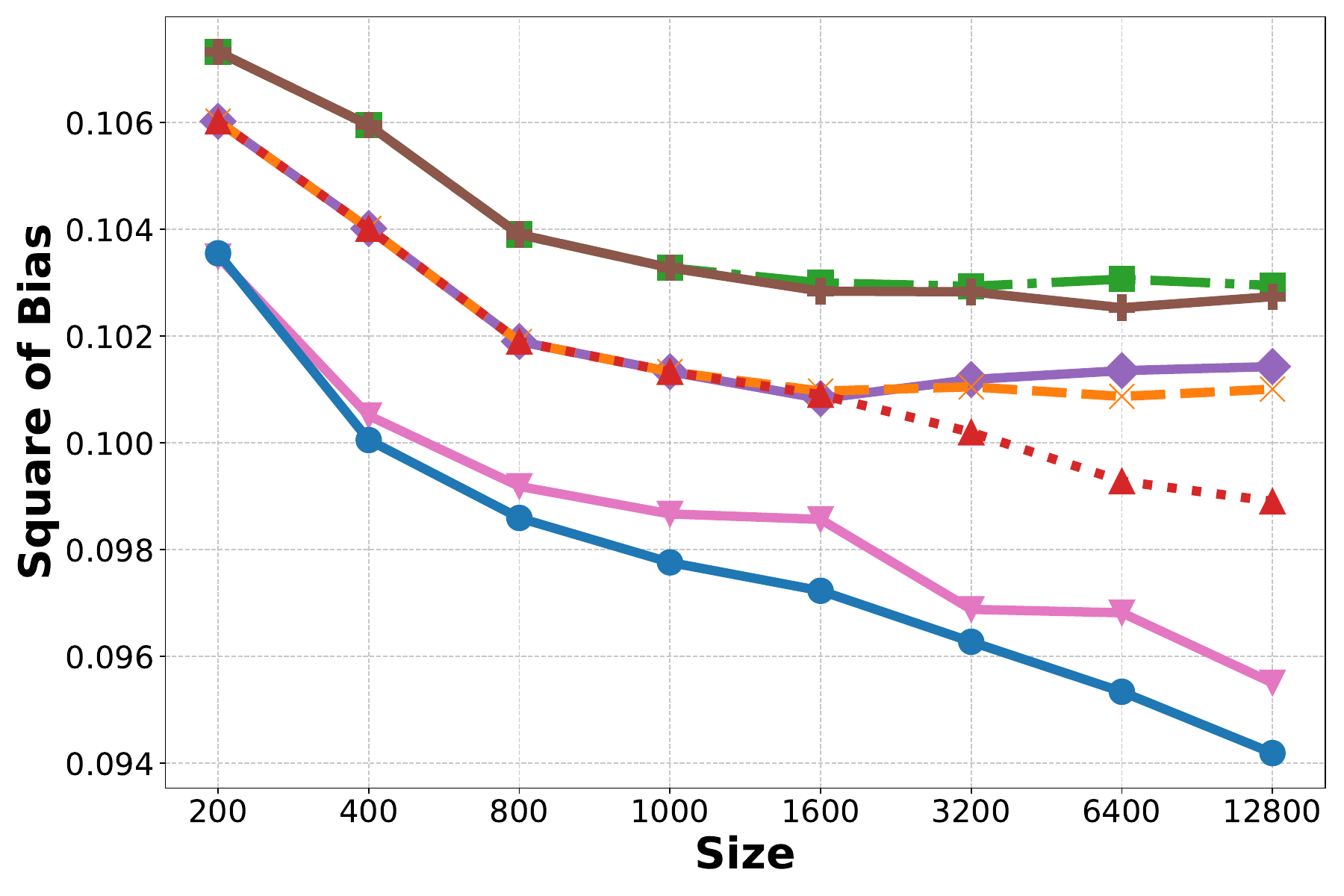}
    {\scriptsize \mbox{(b) {\textit{Bias} on the Adult Dataset.}}}
    \label{fig:adult_bias}
\end{minipage}%
\begin{minipage}{0.32\textwidth}
    \centering
    \includegraphics[width=\linewidth]{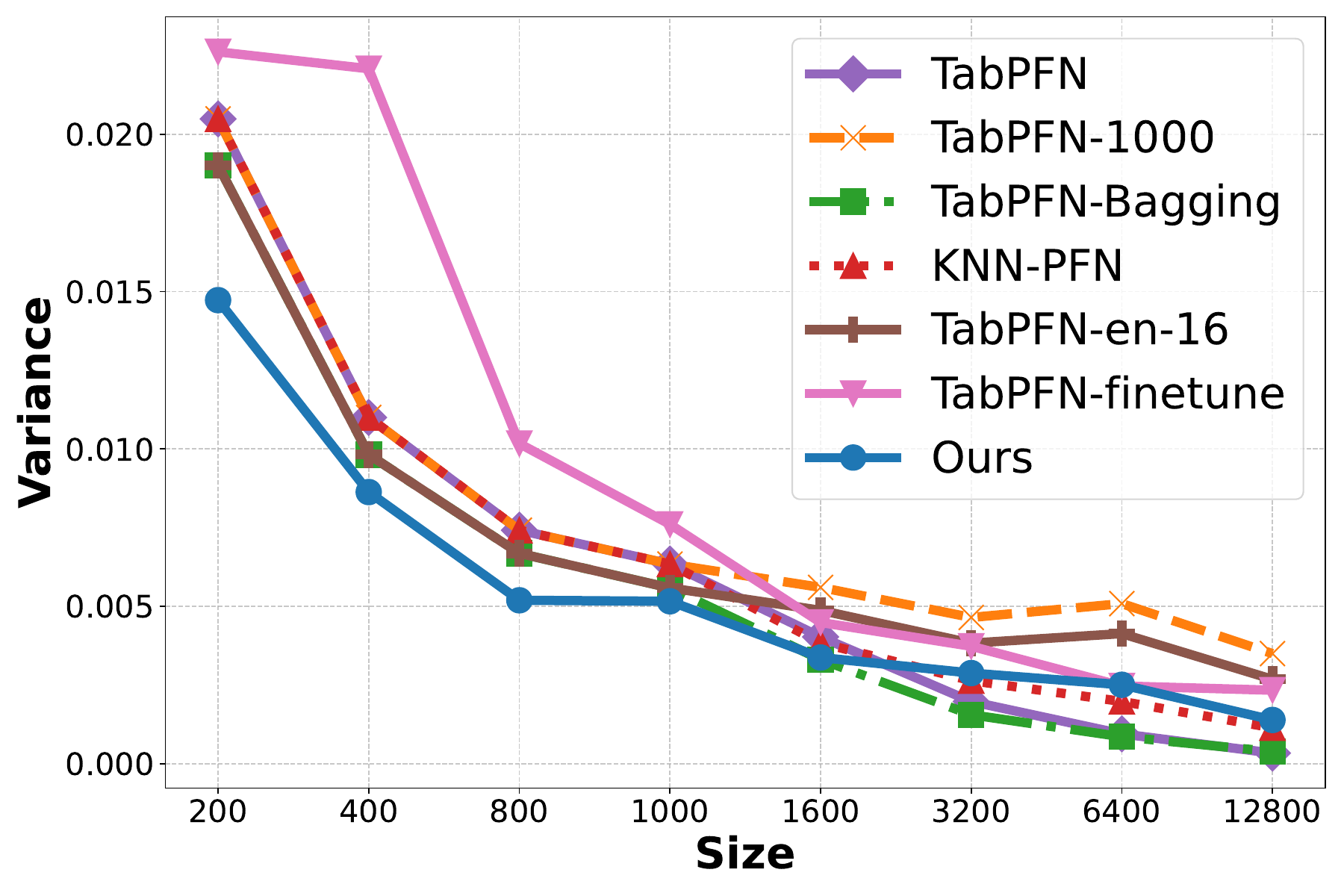}
    {\scriptsize \mbox{(c) {\textit{Variance} on the Adult Dataset.}}}
    \label{fig:adult_variance}
\end{minipage}


\begin{minipage}{0.32\textwidth}
    \centering
    \includegraphics[width=\linewidth]{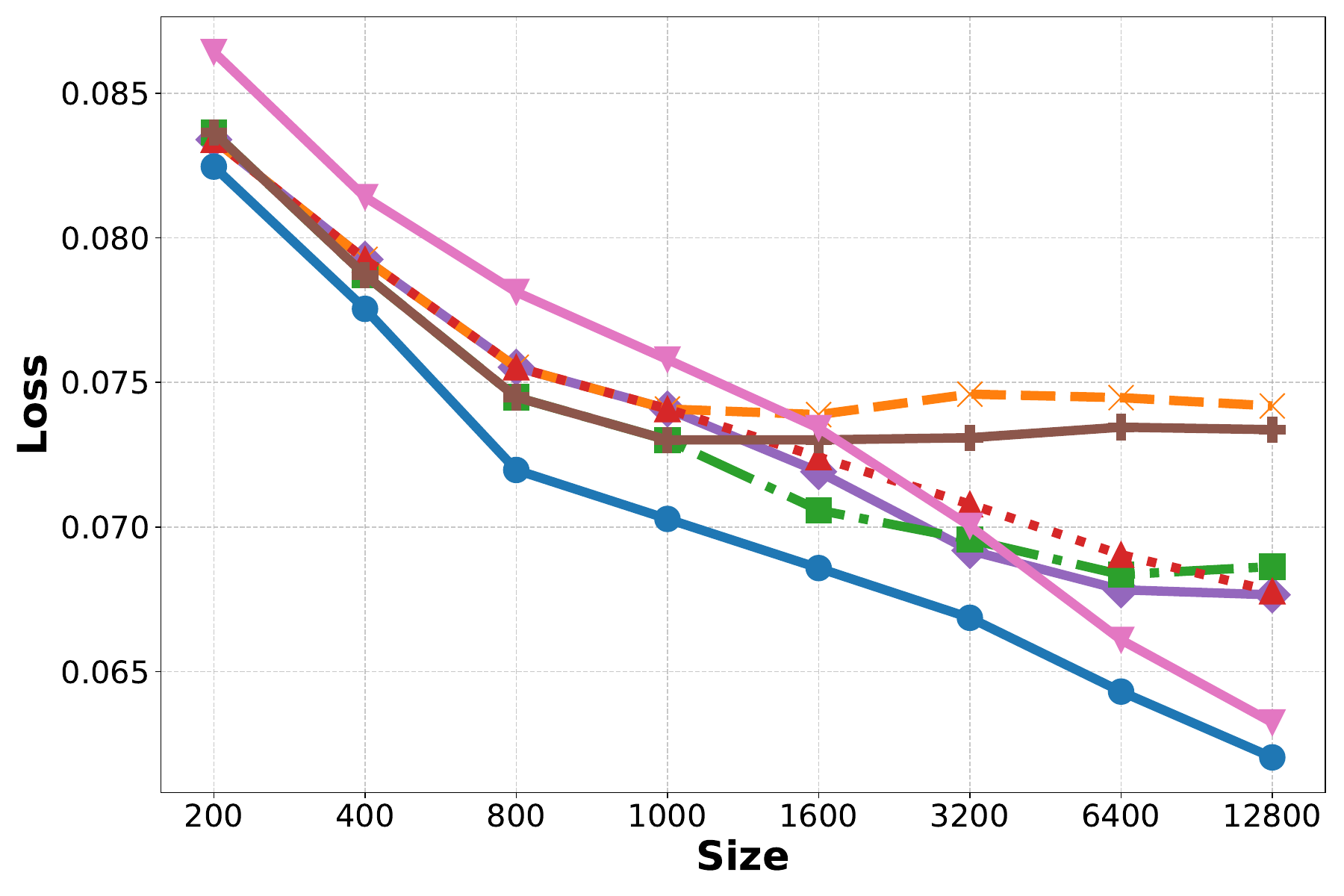}
    {\scriptsize \mbox{(d) {\textit{Generalization Error} on the Bank Dataset.}}}
    \label{fig:bank_generalization_error}
\end{minipage}%
\begin{minipage}{0.32\textwidth}
    \centering
    \includegraphics[width=\linewidth]{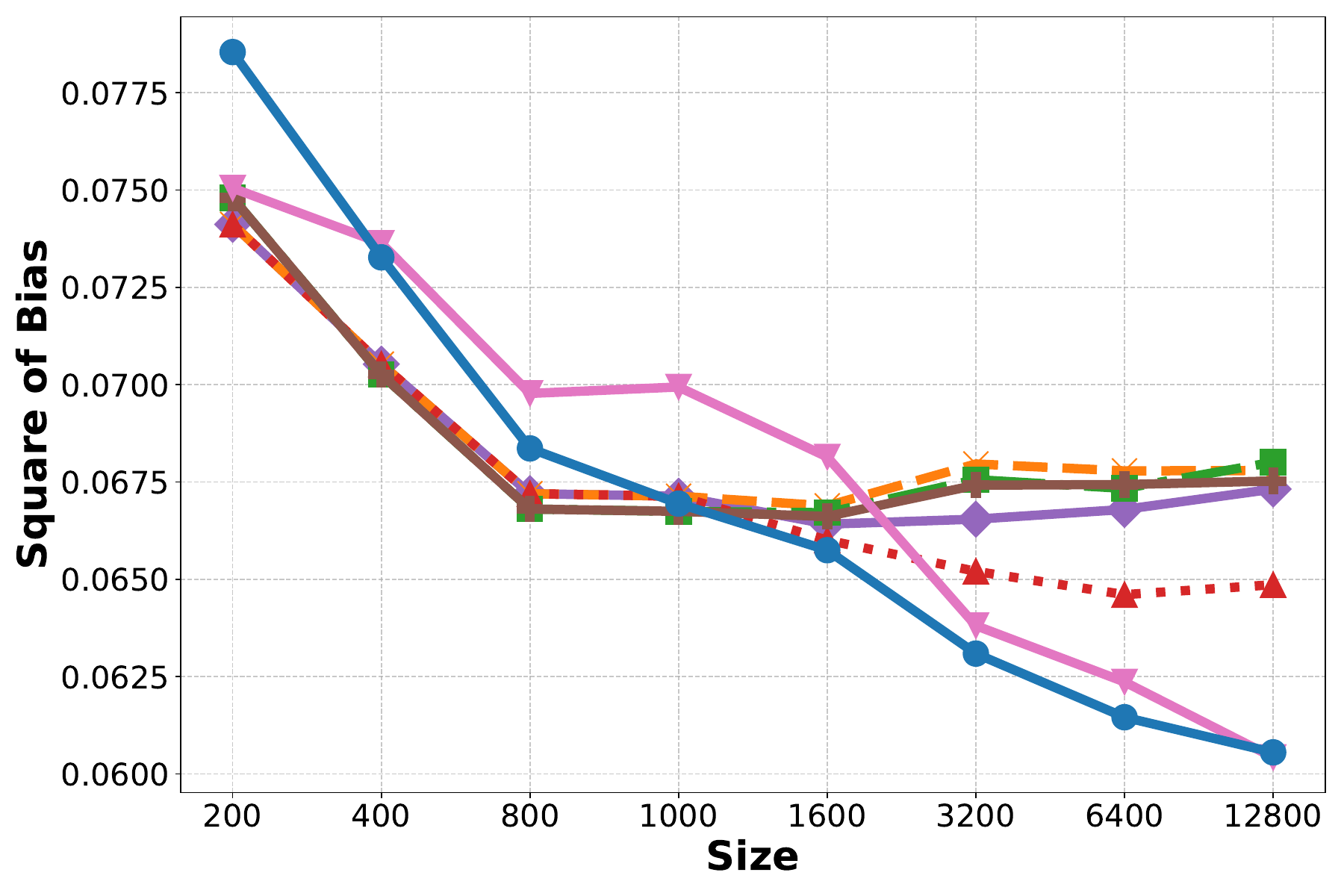}
    {\scriptsize \mbox{(e) {\textit{Bias} on the Bank Dataset.}}}
    \label{fig:bank_bias}
\end{minipage}%
\begin{minipage}{0.32\textwidth}
    \centering
    \includegraphics[width=\linewidth]{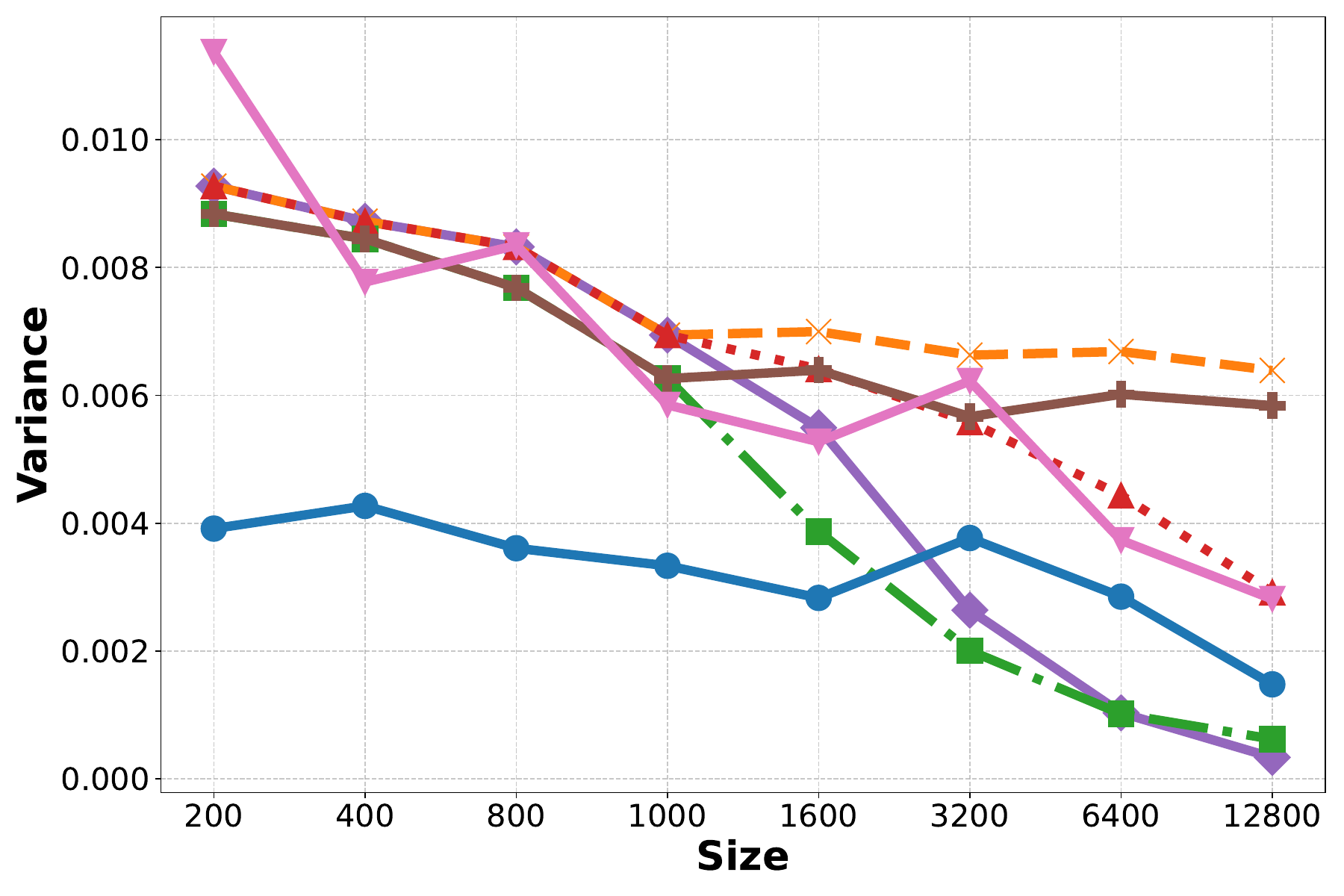}
    {\scriptsize \mbox{(f) {\textit{Variance} on the Bank Dataset.}}}
    \label{fig:bank_variance}
\end{minipage}
\caption{Generalization Error, Bias, and Variance for different methods on the Adult and Bank datasets. The methods shown include TabPFN-1000 (subsample size = 1000), TabPFN, TabPFN-en-16, TabPFN-Bagging, KNN-PFN, TabPFN-finetune, and Ours (\name). The legend is located in the top-right plot for clarity.}
\label{fig:generalization_bias_variance_plots}
\vspace{-5mm}
\end{figure*}
\textbf{Context Selection for Scaling TabPFN.}  The transformer architecture in TabPFN inherently leads to quadratic growth in memory usage as the context length increases.
As a result, the size of the support set used as context in each prediction is limited. Meanwhile, a well-chosen \(D_{\text{train}}\)
provides more relevant support examples, improving generalization. A simple approach is random subsampling~\cite{McElfreshKVCRGW23when}, but this can degrade performance, especially on large datasets~\cite{ICDPFN}. More structured methods include sketching techniques, such as CoreSet, K-Means, and Data Distillation~\cite{ICDPFN}, which aim to compress large datasets into representative subsets while preserving essential information~\cite{Scaling_TabPFN}. 
TuneTables~\cite{TuneTables}, on the other hand, attempts to encode the dataset into a compact learned representation, reducing memory overhead. Another class of methods selects sample-specific contexts instead of a fixed support set. For instance, MixturePFN~\cite{MixturePFN} partitions the training set into multiple subsets and assigns the most relevant one to each test sample. Similarity, LocalPFN~\cite{LocalPFN} and TabDPT~\cite{MaTabDPT}, instead use nearest neighbors to dynamically construct the support set \(D_{\text{train}}\) for each query (KNN-based selection). While these approaches improve TabPFN’s performance by refining \(D_{\text{train}}\), they disrupt PFN-style batching, significantly reducing inference efficiency.

\textbf{Fine-tuning Strategies for Enhancing TabPFN Performance.}
The second approach to improving TabPFN’s performance focuses on fine-tuning the pre-trained model to better adapt to downstream datasets. As shown in Equation~\ref{eq:tabpfn_post_pred}, the predictive distribution 
\( p_\theta(\boldsymbol{y}_{\text{q}} \mid \boldsymbol{X}_{\text{q}}, D_{\text{train}}) \) 
depends on both the training set \( D_{\text{train}} \) and the model parameters \( \theta \). While optimizing \( D_{\text{train}} \) improves the quality of support examples, fine-tuning enhances \( q_\theta \), enabling better alignment with downstream tasks.
A straightforward approach is to fine-tune all model parameters, as seen in TabForestPFN~\cite{TabForestPFN} and LocalPFN~\cite{LocalPFN}, which improves performance by adapting TabPFN’s learned prior to specific datasets. However, full fine-tuning incurs substantial computational costs due to the large number of model parameters. To mitigate this, TuneTables~\cite{TuneTables} offers a more efficient alternative by either fine-tuning the entire model or applying prompt-tuning~\cite{DBLP:conf/emnlp/LesterAC21}, which adjusts only a small set of parameters, reducing resource consumption.
Another efficient approach is adapter-based fine-tuning, as employed in MixturePFN~\cite{MixturePFN}, which fine-tunes additional adapter layers~\cite{Serial_Adapter} rather than modifying the entire model. This strategy provides a balance between computational efficiency and performance improvement, allowing the model to adapt while preserving the efficiency of the original pre-trained TabPFN.  A
detailed related work is presented in~\autoref{sec:related}.\looseness=-1

\subsection{Generalization Analysis of TabPFN Variants}
\label{sec:generalization_analysis_of_tabpfn}
\begin{table*}[h]
\caption{Comparison of TabPFN and related methods in terms of their impact on bias and variance, as well as their effectiveness in handling large datasets, high-dimensional features, adaptability to multiclass classification, and inference efficiency. The compared methods include TabPFN~\cite{Hollmann2022TabPFN}, TuneTables~\cite{TuneTables}, TabForestPFN~\cite{TabForestPFN}, LocalPFN~\cite{LocalPFN}, and MixturePFN~\cite{MixturePFN}. Our proposed method (\name) is distinguished for its ability to balance bias and variance while maintaining efficient scaling, adaptability to multiclass classification, and lightweight fine-tuning. 
}
\vspace{1mm}
\centering
\setlength\tabcolsep{2.5pt}
\label{tab:limitations_comparison}
\resizebox{16cm}{!}{
\begin{tabular}{lcccccccc}
    \toprule
    & \name~(Ours) & TabPFN & TuneTables & TabForestPFN & LocaLPFN &MixturePFN\\
    \midrule
     Reduces Bias  & \greencheck & \redx & \greencheck & \greencheck & \greencheck &  \greencheck \\
     Reduces Variance & \greencheck & \redx & \redx & \redx & \redx & \redx \\
     Scales to Large Datasets & \greencheck & \redx & \greencheck &  \redx & \greencheck & \greencheck\\
     Handles High-Dimensional Data & \greencheck & \redx & \redx & \redx & \redx  & \redx\\
     Adapts to More Than 10 Classes & \greencheck & \redx & \greencheck & \redx & \redx  &\redx \\
     No Additional Inference Cost & \greencheck &\greencheck & \greencheck & \greencheck & \redx & \redx \\
       Fine-Tuning &  lightweight encoder & \redx & prompt~\&  backbone  & backbone & backbone &adapter\\
    \bottomrule
\end{tabular}
}
\end{table*}
While various TabPFN variants have been proposed to enhance performance, their underlying mechanisms remain insufficiently understood. In this section, we leverage the \textbf{bias-variance decomposition framework}, as described in~\autoref{eq:bias-var}, which is introduced by~\citet{SF_PFN}, to analyze the generalization error of TabPFN and its variants on two real-world datasets: \textit{Adult}~\cite{adult_2} and \textit{Bank}~\cite{bank_marketing_222}. 
To assess the impact of different strategies, and based on our previous analysis, we evaluate the original TabPFN (without context length restrictions) alongside several representative variants. These include TabPFN-1000 (random subsampling of 1000 samples), TabPFN-en-16 (randomly sample once and perform feature shuffling 16 times to form an ensemble of 16 predictions), TabPFN-KNN (KNN-based context selection), TabPFN-finetune (full model fine-tuning), and TabPFN-Bagging (bootstrapped sampling with 16 varying contexts).
TabPFN-Bagging, unlike ensemble methods that require training multiple independent models, employs bootstrapped sampling to construct diverse support sets within a single inference process. 
A detailed description of its implementation is provided in Appendix~\ref{appendix:pfn-bagging}.

To further contextualize these findings, we compare all evaluated methods against our proposed approach,~\name. The experimental results, presented in~\autoref{fig:generalization_bias_variance_plots}, reveal several key trends in the bias-variance tradeoff across different TabPFN variants.  
As shown in \autoref{fig:generalization_bias_variance_plots} (b,c,e,f), increasing context length reduces variance while bias plateaus, consistent with prior findings~\cite{SF_PFN}. Similarly,  KNN-based context selection reduces bias but increases variance compared to using the full dataset, as it relies on localized subsets. Since these trends align with previous work, we focus on additional findings.  

\textbf{1) Fine-Tuning and its Impact:}  
\autoref{fig:generalization_bias_variance_plots} (b,e) shows that fine-tuning (TabPFN-finetune) effectively reduces bias by aligning the model’s prior with the characteristics of the dataset. However, it also increases variance, especially when the training set is small, where overfitting amplifies prediction variability, as shown in~\autoref{fig:generalization_bias_variance_plots} (c,f). These results suggest that fine-tuning requires careful regularization to balance bias and variance.  

\textbf{2) Ensemble Strategy and Bias-Variance Tradeoff:}  
\autoref{fig:generalization_bias_variance_plots} (b,c) indicates that ensemble-based methods (TabPFN-en-16) reduce variance but may increase bias due to unsupervised feature transformations. This observation highlights that ensemble methods, while helpful in reducing variance, may require additional bias-reducing strategies to optimize overall model performance. 

\textbf{3) Effectiveness of Bagging:}  
As shown in \autoref{fig:generalization_bias_variance_plots} (c,f), Bagging significantly reduces variance while maintaining stable bias. By introducing diversity through bootstrapped sampling, it achieves variance reduction comparable to ensemble methods but at a lower computational cost. These results highlight Bagging as a simple yet effective approach for improving TabPFN’s performance.

The results above, based on additional findings not covered in~\cite{SF_PFN}, indicate that TabPFN variants typically impact either bias or variance, but rarely both simultaneously. For instance, KNN-based context selection reduces bias but increases variance, while the original ensemble strategy in~\citet{Hollmann2022TabPFN} (TabPFN-en-16) lowers variance but may increase bias. These observations highlight the need for a method that jointly optimizes both aspects, motivating the development of~\name.  

Existing approaches to improving TabPFN focus on context selection or fine-tuning, but often introduce trade-offs in computational efficiency, scalability, and adaptability. A comparison in~\autoref{tab:limitations_comparison} highlights how different strategies affect bias, variance, and their ability to handle large datasets and high-dimensional features efficiently.

\section{\name}
\label{sec:method}
\begin{figure*}[h]
    \centering
    \vspace{-2mm}
    \includegraphics[width=\textwidth]{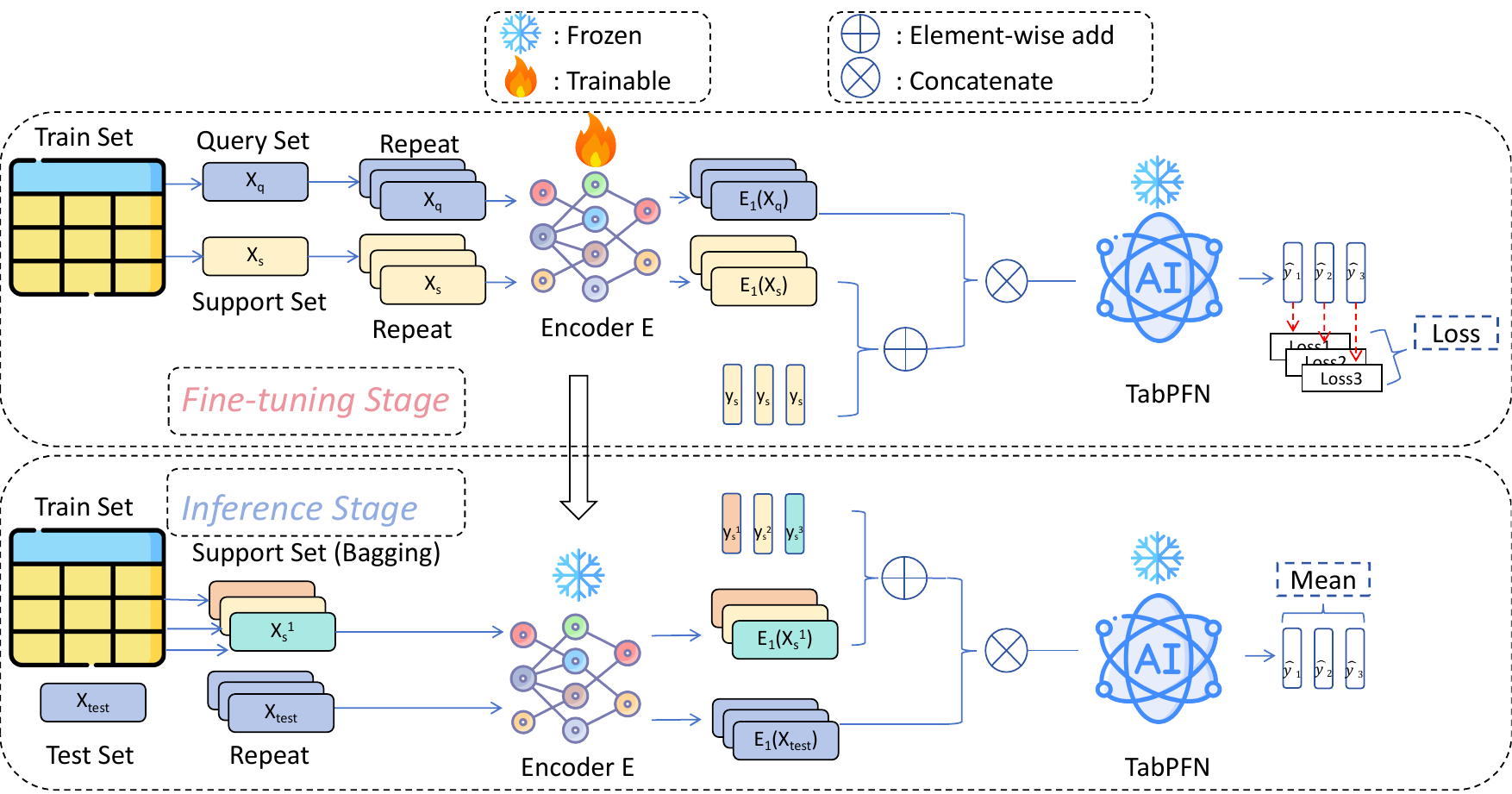}
    \caption{Overview of the proposed method,~\name, which consists of the \textit{fine-tuning stage} and the \textit{inference stage}.}
    \label{fig:overview}
    \vspace{-5mm}
\end{figure*}

To address both bias and variance issues of TabPFN observed during the previous experiments, we propose a unified strategy for improving the performance of TabPFN. In addition to the inference stage, we introduce a fine-tuning stage. Our approach improves performance in both fine-tuning and inference, as shown in~\autoref{fig:overview}. During fine-tuning, we refine input representations to better align the downstream data distribution with the pre-trained TabPFN, reducing both bias and variance. In the inference stage, we incorporate bootstrapped sampling to further reduce variance without additional computational overhead.

\textbf{Minimizing Bias with Encoder-Based {\color{red}\textit{Fine-Tuning}}.}  
Our analysis in~\autoref{sec:generalization_analysis_of_tabpfn} highlights the importance of minimizing bias for improved generalization. To achieve this, we introduce a lightweight encoder while keeping the pre-trained TabPFN parameters frozen. This encoder transforms raw input features into a latent space that better aligns the model’s prior with the downstream data distribution.  

Let \( \boldsymbol{x}_i \in \mathbb{R}^d \) denote the raw feature vector, and let \( \boldsymbol{E}_\Phi \) represent the encoder with parameters \( \Phi \). To enhance expressiveness and incorporate nonlinearity, these layers are constructed using a sequence of operations as described by~\citet{GorishniyRKB21Revisiting}:  
\begin{equation}
\boldsymbol{E}_{\Phi}(\boldsymbol{x}) = \text{Linear}\left( \text{Dropout}\left( \text{ReLU}\left( \text{Linear}\left(\boldsymbol{x} \right) \right) \right)\right).     
\label{eq:encoder}
\end{equation}  
This transformation can be extended by stacking multiple such blocks, allowing the encoder to learn hierarchical representations suited for complex downstream tasks.  

Given a query set \( \boldsymbol{X}_q \) and a support set \( \left( \boldsymbol{X}_{s}, \boldsymbol{y}_{s} \right) \), we define their respective latent representations as:  
\begin{equation}
\boldsymbol{Z}_{q} = \boldsymbol{E}_{\Phi}(\boldsymbol{X}_q), \quad  
\boldsymbol{Z}_{s} = \boldsymbol{E}_{\Phi}(\boldsymbol{X}_s).
\end{equation}  
The posterior predictive distribution over the query labels is then expressed as:  
\begin{equation}
q_\theta\left(\boldsymbol{y}_{q} \mid \boldsymbol{Z}_q, \left(\boldsymbol{Z}_{s},\boldsymbol{y}_{s}\right)\right).
\end{equation}  
This formulation ensures that the encoder effectively maps raw features into a structured space, enabling TabPFN to better capture relevant patterns while mitigating bias.

This fine-tuning method provides a lightweight and effective approach to reduce bias by better aligning the model’s prior with the downstream tasks.  It enables end-to-end fine-tuning, facilitating a more  efficient adaptation of the pre-trained TabPFN to the downstream tasks.

\textbf{Mitigating Variance with Multiple Encoders.}  
To further mitigate variance, we introduce multiple encoders, each learning a distinct transformation of the input data. By jointly training these encoders while keeping the pre-trained TabPFN parameters \(\theta\) frozen, the model captures diverse feature representations, reducing variance.\looseness=-1

For each encoder\( k \), the support and query set representations are encoded as follows:
\[
\boldsymbol{Z}_{\text{s}}^{(k)} = \left[ \boldsymbol{E}^{(k)}_{\Phi}(\boldsymbol{x}_s^{(i)}) \right]_{i=1}^{N_{s}}, \quad \boldsymbol{Z}_{q}^{(k)} = \left[ \boldsymbol{E}^{(k)}_{\Phi}(\boldsymbol{x}_{q}^{(i)}) \right]_{i=1}^{N_{q}},
\]
where $N_s$ and $N_q$
denote the number of samples in the support set and the query set, respectively.
The multiple encoders are trained jointly by minimizing the sum of the individual losses across all encoders. The optimization objective for the fine-tuning phase is:
\begin{equation}
\label{eq:loss}
\min_{\Phi}\mathcal{L}_{\text{total}} = -\sum_{k=1}^{K} \log\left(q_{\theta}\left(\boldsymbol{y}_{q} \mid \boldsymbol{Z}_{\text{q}}^{(k)}, \left(\boldsymbol{Z}_{\text{s}}^{(k)}, \boldsymbol{y}_{s}\right)\right)\right).
\end{equation}
Minimizing \( \mathcal{L}_{\text{total}} \) in~\autoref{eq:loss} ensures that the model jointly trains all encoders to generate diverse yet TabPFN-compatible latent representations, thus reducing variance and providing more stable predictions.

\textbf{Enhancing Performance with Batch Ensemble for Computational Efficiency.}
To further enhance performance without introducing additional computational cost, we integrate the Batch Ensemble technique into the encoder. Specifically, we replace the linear layers in \( \boldsymbol{E}_{\Phi} \) with Batch Ensemble versions, allowing the model to maintain diversity while avoiding the overhead of training multiple independent encoders. This technique introduces shared weight matrices and member-specific scaling factors, reducing the number of trainable parameters while preserving the benefits of ensembling.  

The output of the \( k \)-th base model for encoder layer \( l \) is given by:
\begin{equation}
l_k(\boldsymbol{x}) = s_k \odot \left( W (r_k \odot \boldsymbol{x}) \right) + b_k
\end{equation}
where \( W \) is shared across all base models, and \( r_k, s_k, b_k \) are specific to each base model~\cite{WenTB20BatchEnsemble}. \looseness=-1
\textbf{Bootstrapped Sampling for Variance Reduction in \textit{{\color{blue} Inference} Stage}.}  
To further reduce variance during inference, we apply bootstrapped sampling, generating random subsets of the training set as support sets. These are used to compute predictions across multiple encoders, and their aggregation stabilizes the final output without additional computational overhead.
For each encoder \( k \), the bootstrapped support set is
\(
D_{\text{bootstrap}}^{(k)} = (\boldsymbol{X}_{\text{bootstrap}}^{(k)}, \boldsymbol{Y}_{\text{bootstrap}}^{(k)}).
\)
The final prediction is obtained by averaging over all encoders:
\begin{equation}
\hat{y} = \frac{1}{K} \sum_{k=1}^{K} q_{\theta}\left(\boldsymbol{E}_\Phi(\boldsymbol{x}_{\text{test}}), \left( \boldsymbol{E}_\Phi^{(k)}(\boldsymbol{X}_{\text{bootstrap}}^{(k)}), \boldsymbol{Y}_{\text{bootstrap}}^{(k)} \right) \right).
\end{equation}  
This approach reduces variance while maintaining computational efficiency, ensuring scalability and robustness.

\textbf{Expanding to MultiClass Tasks Beyond 10 Classes.}  
To address TabPFN’s limitation in handling tasks with more than 10 classes, we integrate an Error-Correcting Output Code (ECOC)~\cite{DietterichB95ECOC} strategy. Instead of training separate classifiers, distinct encoders within a single model handle individual binary classification tasks, enabling efficient multiclass classification without additional computational overhead.

\textbf{Summary of Our Approach.}  
Our method reduces bias using a lightweight encoder to align input representations with the pre-trained TabPFN, which is effective for high-dimensional data. Variance is reduced through Bagging with bootstrapped sampling, enabling better generalization on large datasets. For multiclass tasks, we integrate Error-Correcting Output Codes (ECOC) to efficiently handle more than two classes. Additionally, we preserve PFN-style batching, ensuring inference efficiency and scalability for large-scale and high-dimensional datasets.\looseness=-1

\section{Experiments}
\label{sec:exp}
\subsection{Experiment Setup}
\textbf{Datasets.} In our experiments, we evaluate~\name~on one of the largest publicly available tabular benchmark TALENT~\cite{YeACloser}, which includes 120 binary classification datasets and 80 multi-class classification datasets. These datasets are collected from various sources such as UCI, OpenML, Kaggle, and others. To ensure fairness, we remove two datasets, ``PizzaCutter3'' and ``PieChart3,'' as they overlap with TabPFN’s validation set.
In addition, to validate the ability of~\name~to handle high-dimensional feature datasets, we also include 20 high-dimensional datasets sourced from the \href{https://jundongl.github.io/scikit-feature/datasets}{scikit-feature repository}. 

\textbf{Evaluation.}
For the TALENT datasets, we follow the evaluation protocol from~\cite{GorishniyRKB21Revisiting}. Each dataset is randomly split into training, validation, and test sets with proportions of 64\%, 16\%, and 20\%, respectively. For each dataset, we train each model using 15 different random seeds and calculate the average performance on the test set. We report accuracy as the evaluation metric, where higher accuracy indicates better performance.
For a more detailed comparison with other TabPFN variants,  we separately evaluate datasets with fewer than 10 classes, those with more than 10 classes, and high-dimensional datasets. Further details on the experimental setup, are provided in~\autoref{appendix:datasets}.

\textbf{Methods Compared.} We compare~\name~against five categories of methods to evaluate its effectiveness comprehensively:
(1) \textbf{Classical Machine Learning Algorithms:} This category includes widely used classical approaches such as Support Vector Machines (SVM), K-Nearest Neighbors (KNN), and tree-based methods like Random Forest (RForest)~\cite{Breiman01RandomForest}, XGBoost (XGB)~\cite{chen2016xgboost}, CatBoost (CatB)~\cite{Prokhorenkova2018Catboost}, and LightGBM (LightG)~\cite{ke2017lightgbm}.
(2) \textbf{Tabular Deep Learning Models:} We consider state-of-the-art deep learning models for tabular data, including MLP, ResNet, FT-Transformer (FT-T)~\cite{GorishniyRKB21Revisiting}, MLP-PLR~\cite{Gorishniy2022On}, DCNv2~\cite{WangSCJLHC21DCNv2}, AutoInt~\cite{SongS0DX0T19AutoInt}, SNN~\cite{KlambauerUMH17SNN}, ExcelFormer (ExcelF)~\cite{Chen2023Excel}, DANets~\cite{ChenLWCW22DAN}, TabTransformer (TabT)~\cite{Huang2020TabTransformer}, and TabNet~\cite{ArikP21TabNet}.
(3) \textbf{Neighborhood-Based Methods:} To explore neighbor-based strategies, we evaluate TabR~\cite{gorishniy2023tabr} and ModernNCA (MNCA)~\cite{Ye2024ModernNCA}.
(4) \textbf{Ensemble-Based Methods:} We include methods that leverage ensemble strategies, such as TabM~\cite{Yury2024TabM}, NODE, and GrowNet.
(5) \textbf{TabPFN and Its Variants:} Lastly, we compare with TabPFN and its recent variants, including knnPFN~\cite{LocalPFN}, LocalPFN~\cite{LocalPFN}, MixturePFN~\cite{MixturePFN}, and TuneTables~\cite{TuneTables}. \looseness=-1

\textbf{Implementation Details.} All datasets are pre-processed following the methodology outlined in~\citet{GorishniyRKB21Revisiting}. For deep learning-based methods, we set the batch size to 1024. Hyper-parameters for the compared methods are tuned using Optuna~\cite{akiba2019optuna}, performing over 100 trials. The search ranges for hyper-parameters are determined based on~\citet{GorishniyRKB21Revisiting,TALENT} and the official implementations of each method. Once the optimal hyper-parameters are identified, they are fixed for the final evaluation using 15 random seeds. 
To ensure a fair comparison, \textbf{all TabPFN variants, including~\name}, are evaluated using their \textbf{default hyper-parameters without additional tuning}. More details can be found in~\autoref{appendix:datasets}. \looseness=-1

\subsection{\name: State-of-the-Art Performance}
We conducted pairwise significance testing using the Wilcoxon-Holm test~\cite{Demsar06Statistical} among~\name~and all the compared methods. To ensure a fair comparison, we selected 186 datasets from TALENT with fewer than 10 classes, as TabPFN and its variants are not capable of handling datasets with more than 10 classes.
\begin{figure}[t]
    \centering
    \includegraphics[width=0.5\textwidth]{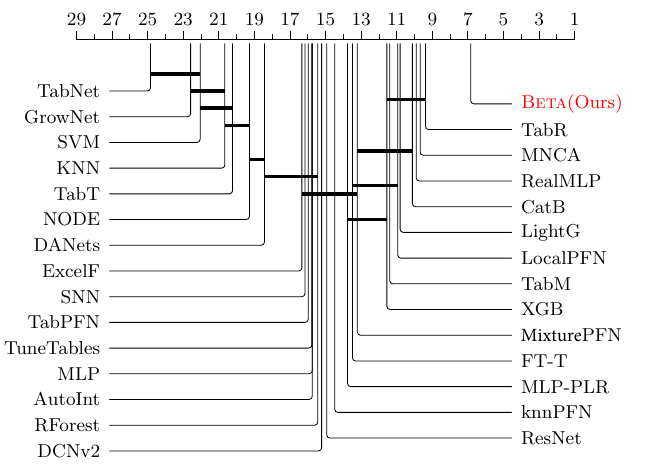}
    \vspace{-5mm}
    \caption{The critical difference diagrams based on the Wilcoxon-Holm test with a significance level
of 0.05 to detect pairwise significance for TALENT datasets with fewer than 10 classes.}
    \label{fig:Wilcoxon-Holm}
    \vspace{-5mm}
\end{figure}
From~\autoref{fig:Wilcoxon-Holm}, it is evident that~\name~outperforms other methods, \textbf{even without hyper-parameter tuning}. This includes methods based on nearest neighbors such as TabR and ModernNCA, ensemble-based approaches like TabM, traditional tree models, and other TabPFN variants. These results underscore the significant potential of pre-trained models for tabular data and demonstrate the effectiveness of our proposed method in adapting to downstream datasets. 

\begin{figure}[t]
    \centering
    \vspace{-3mm}
    \includegraphics[width=0.5\textwidth]{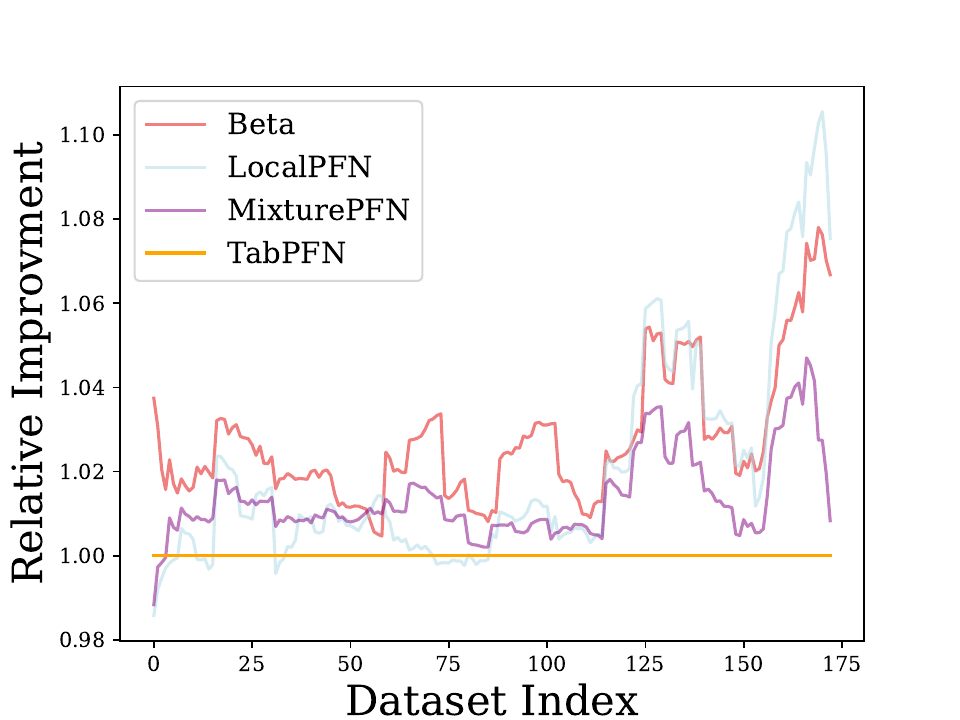}
    \vspace{-5mm}
    \caption{The relative improvement of LocalPFN~\cite{LocalPFN}, MixturePFN~\cite{MixturePFN}, and ~\name~(Ours) over TabPFN across 173 tabular datasets sorted by dataset size (number of rows). The curves represent smoothed results to better illustrate trends in performance improvement as dataset size increases.}
    \label{fig:relative_improvement}
    \vspace{-5mm}
\end{figure}

\textbf{Scale to Large Datasets.} We evaluated~\name~and the variants of TabPFN on 173 tabular datasets, excluding those with more than 100 features, which TabPFN cannot handle. These datasets were sorted by the number of rows in ascending order, and the relative improvement of these methods over TabPFN was shown in~\autoref{fig:relative_improvement}. The results clearly show that the performance improvement by~\name~becomes more pronounced as the dataset size increases, highlighting the scalability and effectiveness of our approach in handling larger datasets. Notably, on the largest datasets in the benchmark, LocalPFN slightly outperforms~\name~in terms of accuracy. However, on more other datasets, \name~demonstrates superior performance. \looseness=-1


\begin{figure}
    \centering
    \includegraphics[width=0.5\textwidth]{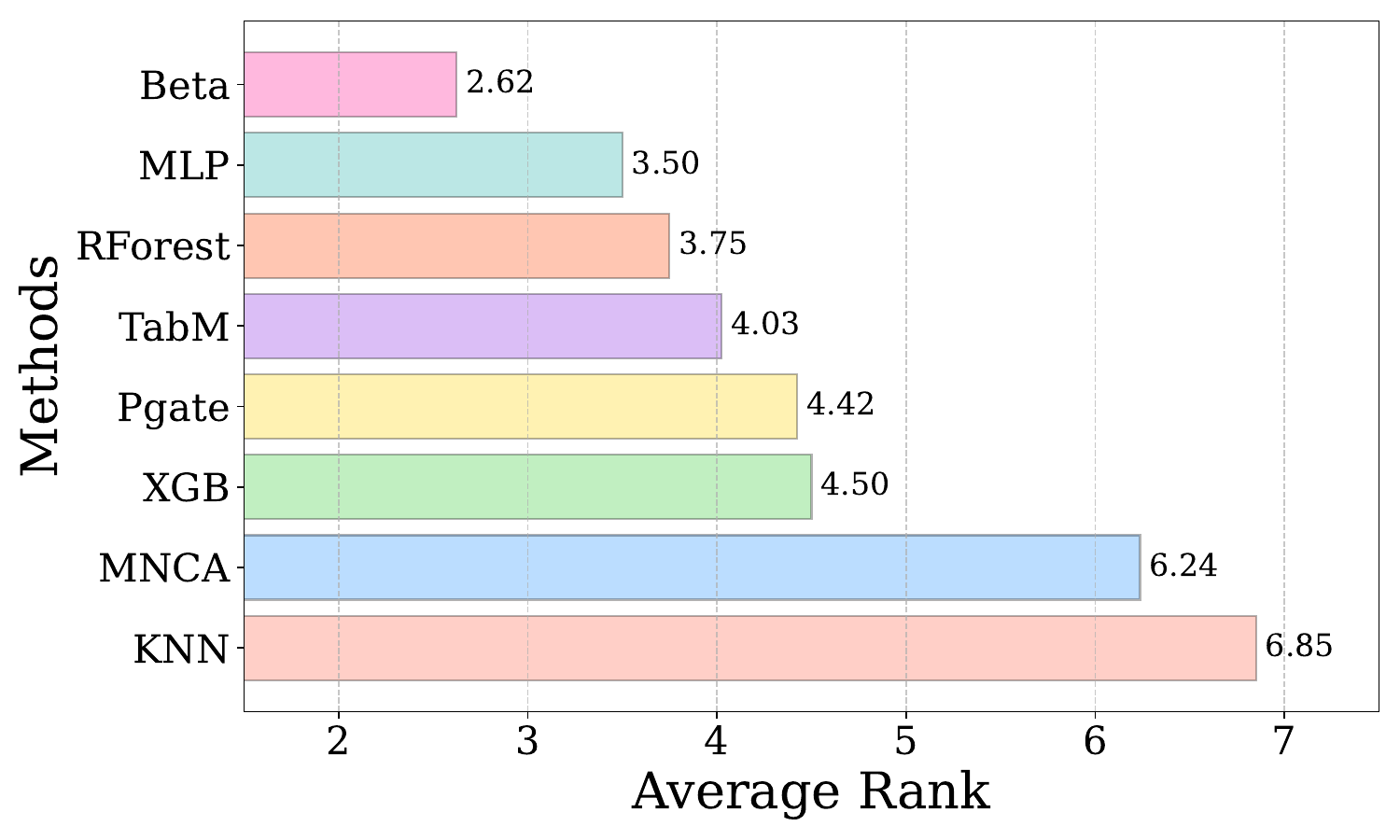}
    \vspace{-5mm}
    \caption{Average ranks of methods on 17 high-dimensional datasets. We compare~\name~with TabM, KNN, MLP, XGBoost (XGB), RandomForest (RForest), ProtoGate (Pgate)~\cite{Jiang2024ProtoGate}, and ModernNCA (MNCA). Lower ranks indicate better performance.}
    \vspace{-5mm}
    \label{fig:high_dim_results}
\end{figure}
\textbf{Handling High-Dimensional Datasets.}  
To assess the effectiveness of~\name~on high-dimensional datasets, we conducted experiments on 20 datasets with extremely high feature dimensions, as detailed in~\autoref{tab:high_dataset_info}. These datasets were selected to evaluate the scalability and adaptability of different methods in complex feature spaces. The average ranks of the compared methods are summarized in~\autoref{fig:high_dim_results}.
The results show that~\name~achieves the best performance, attaining the lowest average rank and outperforming all other methods. TabM and MLP also demonstrate competitive results, ranking second and third, respectively. In contrast, traditional models such as RandomForest and XGBoost, as well as deep learning-based ModernNCA, exhibit lower ranks, highlighting their limitations in high-dimensional settings. Due to memory constraints, we were unable to compare with methods such as FT-T and TabR in this setting.\looseness=-1

\textbf{Performance on MultiClass Classification Tasks with More Than 10 Classes.} In addition, we further investigated the performance of~\name~on classification tasks with more than 10 categories. We selected 12 classification datasets with more than 10 classes from TALENT~\cite{YeACloser}~and compared~\name~with other methods. We report the mean accuracy of each method across the 12 datasets in~\autoref{fig:multiclass_res}~and find that~\name~outperforms the compared methods, demonstrating its superiority on multiclass classification tasks with a larger number of categories.

\subsection{Bias-Variance Analysis, Ablation Study, and Efficiency Comparison}

To evaluate the effectiveness of~\name~in addressing bias and variance, we conduct comprehensive experiments across diverse dataset sizes. Our findings reveal that~\name~consistently achieves lower generalization error than existing methods, demonstrating robust performance regardless of dataset scale. Bias is effectively reduced through encoder-based fine-tuning, aligning TabPFN with downstream data distributions, while variance reduction is achieved via Bagging and the introduction of multiple encoders. Importantly, our method remains effective even on small datasets, where excessive fine-tuning may otherwise increase bias.
In addition, we perform an ablation study to examine the contributions of key components, including the number of encoder paths, periodic activation, Batch Ensemble, and partial parameter tuning. Our results indicate that each of these design elements plays a crucial role in enhancing~\name’s performance. 
Lastly, we compare the inference time and parameter count of~\name~against other methods to highlight its efficiency. A comprehensive analysis of bias-variance trends, ablation experiments, and efficiency comparisons, including inference time and parameter count, is provided in Appendix~\ref{appendix:bias-variance}, Appendix~\ref{appendix:ablation_study}, and Appendix~\ref{appendix:efficiency}.

\section{Conclusion}
\label{sec:conclusion}
In this paper, we propose~\name, a novel approach that enhances the performance of TabPFN by simultaneously addressing both bias and variance. Through a combination of lightweight encoder-based fine-tuning and bootstrapped sampling,~\name~significantly improves TabPFN’s adaptability to high-dimensional, large-scale, and multiclass classification tasks. Our method efficiently reduces bias by aligning downstream data distributions with the pre-trained TabPFN and reduces variance through diverse latent representations and robust inference techniques. Experimental results on over 200 benchmark datasets demonstrate that~\name~consistently outperforms or matches state-of-the-art methods, highlighting its potential to handle complex tabular data tasks with enhanced scalability and robustness.  These contributions provide a scalable and effective solution for leveraging TabPFN in real-world applications, ensuring its success across a broad range of tabular data challenges.


\newpage
\section*{Impact Statement}
Our work advances the field of tabular learning and in-context learning (ICL) by introducing a method that significantly improves TabPFN’s scalability, adaptability, and performance. Given that tabular data is widely used across industries such as finance, healthcare, and e-commerce, achieving state-of-the-art classification accuracy has far-reaching benefits. Our approach enhances the usability of tabular foundation models (TFMs), providing a more effective and scalable alternative to conventional methods while preserving inference efficiency.
We hope that our results inspire further research into ICL-driven approaches for tabular data and their applications in broader real-world tasks. As our method is built on pre-trained TabPFN, its deployment in high-stakes applications should consider potential limitations, such as reliance on the model's prior and sensitivity to dataset characteristics. We do not foresee significant negative societal impacts arising from our work.

\nocite{langley00}

\bibliography{main}
\bibliographystyle{ICML/icml2025}

\newpage
\appendix
\onecolumn

The Appendix consists of four sections:
\begin{enumerate}
\item {\autoref{sec:related}}: Related work.
    \item {\autoref{appendix:datasets}}: Datasets and implementation details.
    \item{\autoref{appendix:limit}}: Hardware and limitations.
    \item{\autoref{appendix:results}}: Additional experimental results
\end{enumerate}

\section{Related Work}
\label{sec:related}

\subsection{Decision-tree-based Models}
Tabular data is one of the most widely used dataset types in machine learning. Gradient-boosted decision trees (GBDTs)~\cite{chen2016xgboost,Prokhorenkova2018Catboost,ke2017lightgbm},  remain a strong baseline for tabular tasks due to their efficiency and high performance. As an ensemble-based method, GBDTs construct multiple decision trees to iteratively minimize the residual loss. 

\subsection{Tabular Deep Learning}
With the advancement of deep learning, an increasing number of studies have explored using deep learning methods for tabular data prediction. These approaches include MLP variants~\cite{KlambauerUMH17SNN,GorishniyRKB21Revisiting,Gorishniy2022On,David2024RealMLP}, neural networks specifically designed for tabular structures~\cite{WangFFW17DCN,WangSCJLHC21DCNv2,Chen2023TabCaps}, attention-based models~\cite{SongS0DX0T19AutoInt,Huang2020TabTransformer,GorishniyRKB21Revisiting,Chen2023Excel}, methods incorporating regularization~\cite{PTARL,jeffares2023tangos,Wu2024SwitchTab}, tree-mimic methods~\cite{ArikP21TabNet,PopovMB20Neural,Badirli2020GrowNet}, and context-based methods~\cite{gorishniy2023tabr,Ye2024ModernNCA}. Despite these advancements, recent benchmarks~\cite{Grinsztajn2022Why,McElfreshKVCRGW23when,YeACloser} have consistently demonstrated that gradient-boosted decision trees (GBDTs) outperform deep learning in tabular prediction tasks.
The superior performance of GBDTs can be attributed to two main factors: (1) their ability to handle heterogeneous tabular datasets, which often describe high-frequency target functions~\cite{DBLP:conf/icml/BasriGGJKK20,Grinsztajn2022Why},  and (2) the ensemble nature of GBDTs. Prior attempts to introduce ensemble-like mechanisms into tabular deep learning~\cite{Badirli2020GrowNet,PopovMB20Neural,DBLP:conf/icml/Chen2023Trompt}, have not been widely successful~\cite{Grinsztajn2022Why,YeACloser}. However, recent works like TabM~\cite{Yury2024TabM} integrate Batch Ensemble~\cite{WenTB20BatchEnsemble} techniques into the tabular domain, showing how efficient ensembling can be achieved with deep learning models. \looseness=-1

\subsection{Tabular Foundation Models}
While tabular foundation models (TFMs) are not as developed as foundation models in other domains, such as computer vision~\cite{SahariaCSLWDGLA22Diffusion} and natural language processing~\cite{BrownGPT3}, recent efforts have introduced various architectures to bridge this gap~\cite{WhyTFM}. Some approaches aim to explore model components that can be shared across datasets~\cite{LiuDEN,zhu2023xtab}, while others focus on utilizing the semantic information inherent in tabular datasets~\cite{Wang2022TransTab,YanTpBerta,Tabular_8B}.
As a Transformer-based model, TabPFN~\cite{Hollmann2022TabPFN,hollmann2025tabpfn} stands out for its exceptional performance and efficiency on small datasets. By leveraging the in-context learning capability of transformer~\cite{BrownGPT3}, it can make predictions for unseen instances without parameter updates. However, TabPFN faces limitations related to dataset size and feature dimensionality. To address these challenges, some studies have explored improvements in its architecture and training paradigm~\cite{TabForestPFN,MaTabDPT}, while others focus on adaptation techniques that expand TabPFN's applicability to a broader range of downstream tabular datasets~\cite{TuneTables,LocalPFN}.
Our work falls into the latter category, offering a straightforward, efficient, and effective approach to align TabPFN with downstream datasets.\looseness=-1

\subsection{Parameter-efficient Fine-Tuning} 
Some approaches for adapting Tabular Foundation Models (TFMs) to downstream tasks simply fine-tune the entire model~\cite{LocalPFN,TabForestPFN}, which leads to performance improvements but incurs high computational and storage costs. Parameter-efficient fine-tuning (PEFT) offers a solution to the challenge by enabling adaptation with a minimal number of trainable parameters~\cite{Guide_PEFT}. 
Based on their operational mechanisms, PEFT methods can be broadly categorized into four paradigms~\cite{PEFT_Survey}:
\begin{itemize}[noitemsep,topsep=0pt,leftmargin=*]
\item \textbf{Additive Methods}. These methods incorporate additional lightweight modules into the model architecture, such as adapters~\cite{Serial_Adapter,SideTuning,MixturePFN} or soft prompts~\cite{Prefix_Tuning,TuneTables}. 
\item \textbf{Selective Methods}. Instead of introducing new parameters, selective methods strategically identify and update the most relevant parameters while freezing the rest~\cite{FuYSLBC23SAM,He0ZTZ23SPT}.
\item \textbf{Reparameterized Methods}. These methods employ low-rank decomposition or equivalent transformations to reduce the parameter space during fine-tuning~\cite{AghajanyanGZ20SAID,HuSWALWWC22LoRA,ValipourRKG23DyLoRA}. 
\item \textbf{Hybrid Methods}. By combining the strengths of multiple PEFT strategies, hybrid methods create a unified framework that enhances fine-tuning performance while maintaining efficiency~\cite{MaoMHAM0YK22UniPELT,ChenZS0SY23S4}.
\end{itemize}

Our proposed method adopts parameter-efficient tuning focused on input feature adaptation. This design is motivated by the unique characteristics of tabular data: Tabular datasets are inherently heterogeneous, with varying structures and feature distributions across different datasets~\cite{Zhou2023TabToken,TabPTM,BorisovLSHPK24TabularSurvey}. Adapting TabPFN through input feature alignment effectively mitigates the constraints on input dimensionality, enhancing its applicability across a wider range of tasks.  By employing parameter-efficient fine-tuning, we align TabPFN with downstream datasets, addressing its existing limitations~\cite{Hollmann2022TabPFN}. 

\section{Datasets and implementation details}
\label{appendix:datasets}
In this section, we outline the descriptions of the datasets used in the experiments and the preprocessing steps applied to them before training. Additionally, we will describe the implementation details of~\name~and the comparison methods.

\subsection{Datasets information}
For both the main experiments and the classification tasks with more than 10 categories, we use the current largest tabular benchmark~\cite{YeACloser}, which includes \textbf{120 binary classification datasets and 80 multiclass classification} datasets, spanning diverse domains such as healthcare, biology, finance, education, and physics. For more detailed information about these datasets, please refer to~\citet{YeACloser}.

For each dataset, we randomly select 20\% of the instances to form the test set. The remaining 80\% is further split, with 20\% reserved as a validation set. This validation set is used for hyperparameter tuning (for the comparison methods) and early stopping. The hyperparameters that yield the best performance on the validation set are selected for final evaluation on the test set.\looseness=-1

\textbf{Remark:} \name~are based on pre-trained models TabPFN~\cite{Hollmann2022TabPFN}. To ensure a fair comparison, we do not perform hyperparameter search, and instead use the default hyperparameters for all methods.\looseness=-1

For high-dimensional datasets, we obtained the data from~\href{https://jundongl.github.io/scikit-feature/datasets}{scikit-feature repository}, and excluded those with more than 10 classes. This resulted in a final set of 20 datasets, as shown in~\autoref{tab:high_dataset_info}. The dataset splitting follows the same approach as in the main experiments, but due to the smaller number of instances in these datasets, we use the default hyperparameters for the experiments and do not perform hyperparameter search. The validation set is only used for early stopping. For each dataset, we perform five random splits and run all methods three times on each split using three different seeds (0, 1, 2), resulting in a total of 15 runs per dataset. The final results are reported as the mean of these 15 runs. \looseness=-1

\begin{table}[ht]
\centering
\caption{Dataset Information for High-Dimensional Data Experiments: A collection of 20 datasets with varying numbers of instances, features, and classes used in our high-dimensional experiments.}
\begin{tabular}{lccc|lccc}
\toprule
\textbf{Dataset} & \textbf{\#Instances} & \textbf{\#Features} & \textbf{\#Classes} &\textbf{Dataset} & \textbf{\#Instances} & \textbf{\#Features} & \textbf{\#Classes} \\
\midrule
BASEHOCK&	1993&	4862&	2  & lung\_discrete&	73&	325&	7 \\
PCMAC	&1943	&3289	&2  & warpPIE10P&	210&	2420&	10\\
RELATHE	&1427	&4322	&2  &orlraws10P &	100&	10304&	10 \\
ALLAML	& 72	&7129	&2  & Prostate\_GE	&102&	5966&	2 \\
CLL\_SUB\_111&	111&	11340&	3 & SMK\_CAN\_187&	187&	19993&	2 \\
colon&	62&	2000&	2  & warpAR10P	&130	&2400	&10 \\
GLI\_85&	85	&22283	&2  & arcene&	200&	10000&	2\\
GLIOMA&	50	&4434&	4  & gisette	&7000&	5000&	2 \\
leukemia&	72&	7070&	2  & madelon&	2600&	500&	2 \\
lung&	203&	3312&	5 & TOX\_171&	171&	5748&	4\\
\bottomrule
\end{tabular}
\label{tab:high_dataset_info}
\end{table}

\subsection{Dataset Pre-processing}
Unless otherwise specified,  we adopt the data preprocessing pipeline outlined by~\citet{GorishniyRKB21Revisiting}. For numerical features, we standardize by subtracting the mean and scaling the values to unit variance. Categorical features are converted into a model-compatible format using one-hot encoding.

\subsection{Implementations}
\textbf{\name.} For all experiments except for the ablation study, we set the context size to 1000 and fixed the number of bootstrap sampling iterations to 16. For the feature transformation encoder, the main structure is a two-layer MLP, with both the hidden and output dimensions set to 100, using the ReLU activation function. In addition, to enhance the expressive power of the encoder, we also apply periodic activation~\cite{Gorishniy2022On}. The initialization of Batch Ensemble is inspired by~\citet{Yury2024TabM}. During fine-tuning, we use the pre-trained model checkpoint\footnote{\href{https://github.com/automl/TabPFN/blob/tabpfn_v1/tabpfn/models_diff/prior_diff_real_checkpoint_n_0_epoch_42.cpkt}{https://github.com/automl/TabPFN/blob/tabpfn\_v1/tabpfn/models\_diff/prior\_diff\_real\_checkpoint\_n\_0\_epoch\_42.cpkt}}, and fine-tuning is performed using the AdamW~\cite{LoshchilovH19AdamW} optimizer with a learning rate of 0.003, weight decay of 1e-5, and a batch size of 1024. In the fine-tuning phase, only the encoder parameters are updated, while the pre-trained TabPFN parameters are frozen.  
For experiments on high-dimensional datasets, due to the large number of features and the limitations of device memory, we do not use periodic activation. In the classification experiments with more than 10 classes, we use a code length of 32, which means there are 32 paths in the encoder. Additionally, we observed that for certain datasets, simply adjusting the output dimension of the pre-trained model and fine-tuning it can yield strong results~\cite{TuneTables}. Based on validation set performance, we choose between  ECOC  and fine-tuning the output layer MLP approaches.\looseness=-1

\textbf{TabPFN and its variants.} 
For TabPFN, we use the implementation from TALENT~\cite{TALENT}; For TuneTables, we adopt the official code\footnote{\href{https://github.com/penfever/TuneTables}{https://github.com/penfever/TuneTables}} and use the prompt tuning mode without performing full model fine-tuning; For LocalPFN, since the code has not been released, we replicate the hyperparameters provided in the original paper. Specifically, we set the number of neighbors \(k = 1000\). During our experiments, we found that fine-tuning LocalPFN with a smaller learning rate yields better performance, so we set the learning rate to 1e-5; For MixturePFN, we reproduce the model using the hyperparameters recommended in the original paper. The number of training samples in the prompt is set to \(B = 3000\), and the number of experts is set to the training set size divided by \(B\), rounded up to the nearest integer.

\textbf{Other Methods from TALENT.}
For the other methods in TALENT, we either use the results provided~\cite{YeACloser} or perform a hyperparameter search using the hyperparameters provided by TALENT. In the case of hyperparameter tuning, we conduct 100 trials of hyperparameter search for each method\footnote{\href{https://github.com/qile2000/LAMDA-TALENT/tree/main/LAMDA_TALENT/configs}{https://github.com/qile2000/LAMDA-TALENT/tree/main/LAMDA\_TALENT/configs}}. For each search, we run the experiments using 15 different seeds and report the average results across these runs.

\subsection{Details of TabPFN}
\label{appendix:details_of_pfn}

In the original TabPFN approach, the model processes the input data in two phases: pre-training and inference. The pre-training phase involves training the model on synthetic data, while the inference phase is used to make predictions on new, unseen data.

For the input, we consider a training dataset \( D_{\text{train}} = (X_{\text{train}}, y_{\text{train}}) \), where \( X_{\text{train}} \in \mathbb{R}^{N_{\text{train}} \times d} \) represents the feature matrix, and \( y_{\text{train}} \in \mathbb{R}^{N_{\text{train}}} \) represents the corresponding labels. Here, \( N_{\text{train}} \) denotes the number of training samples, and \( d \) is the number of features. For datasets with fewer features than the fixed input dimensionality \( d_{\max} \), zero-padding is applied to extend the feature vectors to the required size \( d_{\max} \). Specifically, each feature vector \( X_i \in \mathbb{R}^d \) is padded with zeros to form \( \tilde{X}_i \in \mathbb{R}^{d_{\max}} \), as follows:

\[
\tilde{X}_i = \left[ X_i, \mathbf{0} \right] \in \mathbb{R}^{d_{\max}}.
\]

The zero-padded feature matrix \( \tilde{X}_{\text{train}} \) is then used during inference for making predictions.

For inference, we consider a support set \( S \) containing the training examples and a query set \( Q \) containing the new, unseen instances for which we wish to make predictions. The query set \( Q \) consists of \( N_{\text{test}} \) test samples, and the goal is to predict the corresponding labels \( y_{\text{test}} \).

TabPFN uses \textbf{PFN-style batching} for inference, where both the support set \( S \) and query set \( Q \) are batched together into a single prompt, which is then fed into the pre-trained model for prediction. This batching process allows the model to utilize the support set as context for making predictions on the query set.

During the processing, the features of both the support set and the query set are transformed into token representations. The support tokens \(L_{\text{support}}\) are obtained as:

\[
L_{\text{support}} = \tilde{X}_{\text{support}} W_x + y_{\text{support}} w_y^T,
\]

where \( \tilde{X}_{\text{support}} \in \mathbb{R}^{|S| \times d_{\max}} \) is the zero-padded feature matrix of the support set, \( W_x \in \mathbb{R}^{d_{\max} \times d_{\text{token}}} \) is the embedding matrix for the features, and \( w_y \in \mathbb{R}^{d_{\text{token}}} \) is the embedding matrix for the labels. The query tokens \(L_{\text{query}}\) are similarly transformed as:

\[
L_{\text{query}} = X_{\text{query}} W_x,
\]

where \( X_{\text{query}} \in \mathbb{R}^{|Q| \times d_{\max}} \) represents the query set of test samples.

These token representations are then passed through a standard transformer model, which includes a special attention mechanism. The attention mask ensures that the support tokens can only attend to other support tokens, and query tokens can attend to both the support tokens and themselves. Query tokens do not attend to other query tokens, preventing any information leakage between query samples during inference. Finally, the output tokens corresponding to the test instances are extracted and mapped to 10-class logits for classification.

\subsection{Implementation of TabPFN-Bagging}
\label{appendix:pfn-bagging}

TabPFN-Bagging is a variance reduction technique that leverages \textbf{bootstrapped sampling} to create diverse support sets for model inference. Unlike standard ensemble approaches, which require training multiple independent models, TabPFN-Bagging generates multiple resampled versions of \( D_{\text{train}} \) within a single inference process, reducing variance without additional computational overhead.

\textbf{Bootstrapped Sampling in TabPFN.} 
Given a training dataset \( D_{\text{train}} = \{ (\boldsymbol{x}_{i}, \boldsymbol{y}_{i}) \}_{i=1}^{N_{\text{train}}} \), we generate \( K \) bootstrapped support sets \( D_{\text{train}}^{(k)} \), where each set is constructed by randomly sampling \( N_{\text{sub}} \) instances with replacement:
\begin{equation}
D_{\text{train}}^{(k)} = \left\{ \left(\boldsymbol{x}_{i}^{(k)}, \boldsymbol{y}_{i}^{(k)}\right) \right\}_{i=1}^{N_{\text{sub}}}, \quad k = 1, \dots, K.
\end{equation}
Each \( D_{\text{train}}^{(k)} \) serves as an alternative context set for model inference, introducing diversity into the prediction process.

\textbf{Prediction Aggregation.} 
For each test instance \( \boldsymbol{x}_{\text{q}} \), we compute the posterior predictive distribution using the bootstrapped support sets. Given the pre-trained TabPFN model parameterized by \( \theta \), the predictive probability for label \( \boldsymbol{y}_{\text{q}} \) is computed as:
\begin{equation}
p_{\theta}^{(k)}(\boldsymbol{y}_{\text{q}} \mid \boldsymbol{x}_{\text{q}}, D_{\text{train}}^{(k)}) = 
\frac{\exp(q_{\theta}(\boldsymbol{x}_{\text{q}}, D_{\text{train}}^{(k)})_{[\boldsymbol{y}_{\text{q}}]})}
{\sum_{c=1}^{C} \exp(q_{\theta}(\boldsymbol{x}_{\text{q}}, D_{\text{train}}^{(k)})_{[c]})}.
\end{equation}
where \( q_{\theta} \) denotes the logits produced by TabPFN for the given support set \( D_{\text{train}}^{(k)} \).

To obtain the final prediction, we aggregate the results across all bootstrapped support sets using either uniform averaging:
\begin{equation}
p_{\theta}(\boldsymbol{y}_{\text{q}} \mid \boldsymbol{x}_{\text{q}}, D_{\text{train}}) = 
\frac{1}{K} \sum_{k=1}^{K} p_{\theta}^{(k)}(\boldsymbol{y}_{\text{q}} \mid \boldsymbol{x}_{\text{q}}, D_{\text{train}}^{(k)}),
\end{equation}
or a weighted aggregation method, where weights \( w_k \) are assigned based on the confidence of each individual model:
\begin{equation}
p_{\theta}(\boldsymbol{y}_{\text{q}} \mid \boldsymbol{x}_{\text{q}}, D_{\text{train}}) = 
\sum_{k=1}^{K} w_k p_{\theta}^{(k)}(\boldsymbol{y}_{\text{q}} \mid \boldsymbol{x}_{\text{q}}, D_{\text{train}}^{(k)}), \quad \sum_{k=1}^{K} w_k = 1.
\end{equation}

\textbf{Computational Efficiency.} Since bootstrapped sampling only modifies the selection of \( D_{\text{train}} \) without altering the model architecture, TabPFN-Bagging incurs no additional computational cost beyond standard inference with ensemble. Unlike ensemble-based methods that require multiple independent model evaluations, all computations occur within a single forward pass of the pre-trained TabPFN. This makes TabPFN-Bagging an efficient and scalable variance reduction strategy.

\section{Hardware and limitations}
\label{appendix:limit}
\subsection{Hardware}
Most of the experiments were performed  with four NVIDIA 4090 GPUs and four NVIDIA A6000 GPUs.\looseness=-1

\subsection{Limitations}
While our approach significantly improves the scalability and adaptability of TabPFN, it has certain limitations that present opportunities for future research. Due to the current design of TabPFN, our method has been evaluated solely on classification tasks, and extending it to regression tasks remains an open challenge. As tabular foundation models continue to evolve, we anticipate that our approach can be adapted to support regression, broadening its applicability to a wider range of real-world problems.
Additionally, our experiments assume that training and test instances are drawn from the same underlying distribution. However, in practical applications, distribution shifts such as covariate shift or concept drift may occur, which could impact model performance. Addressing such shifts requires robust adaptation strategies or specialized training techniques, which are beyond the scope of this work. Importantly, we emphasize that handling non-IID scenarios is not a trivial extension of existing methods but rather a distinct challenge that necessitates dedicated research efforts. Future work should explore how TabPFN-based models can be adapted to regression tasks and how their robustness to distributional shifts can be enhanced to ensure broader applicability in real-world settings.

\section{Additional Experiments}
\label{appendix:results}

\subsection{Performance on MultiClass Classification Tasks with
More Than 10 Classes}
\begin{figure}
    \centering
    \includegraphics[width=\linewidth]{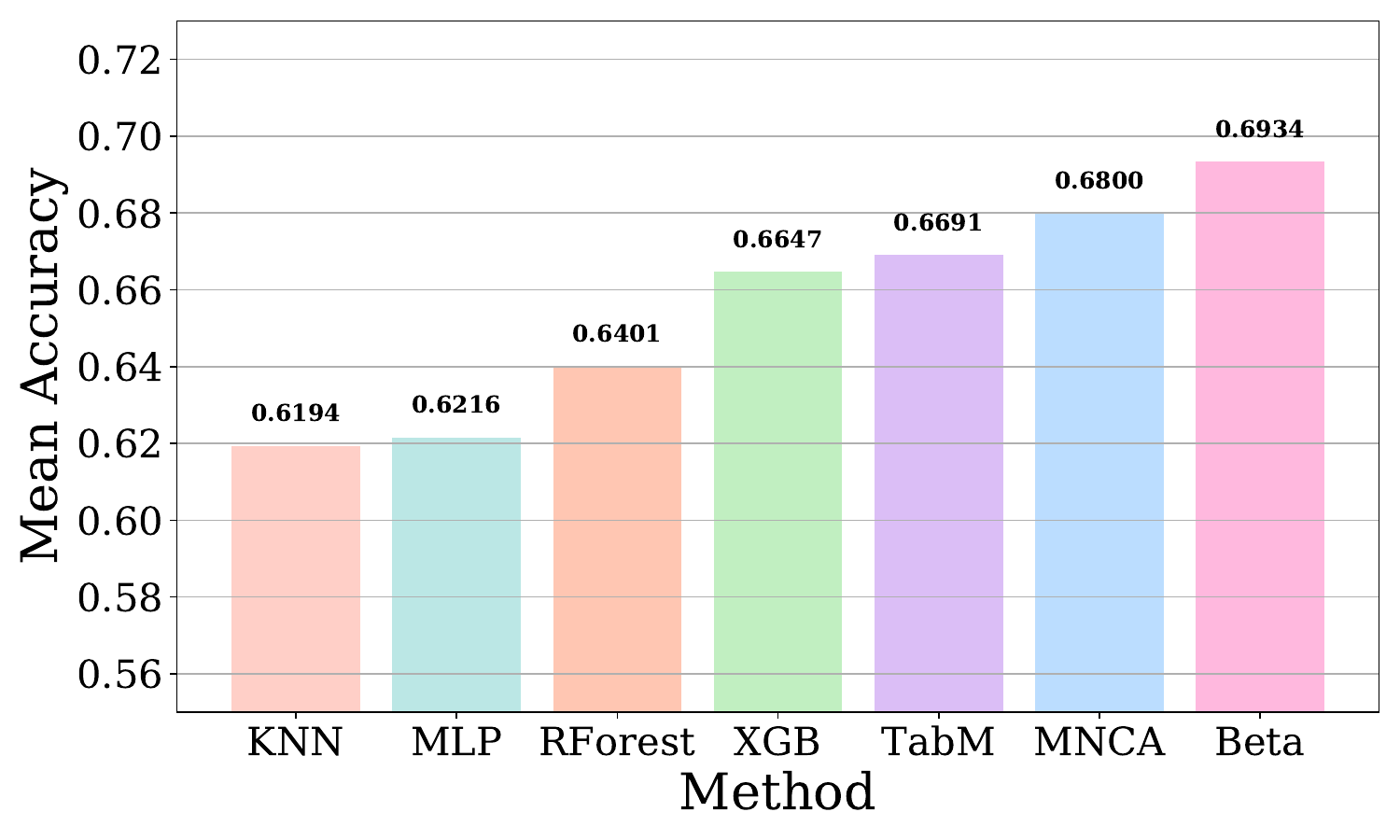}
    \caption{Comparison of the mean accuracy across 12 classification datasets with more than 10 classes. \textbf{Since the original TabPFN cannot process classification tasks with more than ten categories}, we compare~\name~against alternative methods, including TabM, KNN, MLP, XGBoost (XGB), RandomForest (RForest), and ModernNCA (MNCA). }
    \label{fig:multiclass_res}
\end{figure}

To evaluate the effectiveness of~\name~on multiclass classification tasks, we conducted experiments on 12 datasets containing more than 10 classes, sourced from TALENT\cite{YeACloser}. These datasets were selected to assess the scalability and adaptability of different methods in handling more complex classification problems beyond the 10-class limitation of TabPFN.~\autoref{fig:multiclass_res}~presents the mean accuracy of each method across these datasets. We compare~\name~against several strong baselines, including KNN, XGBoost~\cite{chen2016xgboost}, RandomForest~\cite{Breiman01RandomForest}, MLP~\cite{GorishniyRKB21Revisiting}, ModernNCA~\cite{Ye2024ModernNCA}, and TabM~\cite{Yury2024TabM}.
 All methods, including~\name, were evaluated using their default hyperparameters without any tuning, ensuring a fair comparison of out-of-the-box performance.
From the results,~\name~achieves the highest mean accuracy across all datasets, demonstrating superior generalization ability in high-category classification settings. This advantage stems from our integration of the Error-Correcting Output Codes (ECOC) framework, which effectively decomposes multiclass problems into multiple binary subproblems. This enables TabPFN to handle classification tasks with more than 10 classes efficiently, without requiring significant modifications or additional computational overhead.
Notably, tree-based methods like XGBoost and RandomForest exhibit a decline in performance as the number of categories increases.





\subsection{Performance on Real Datasets: Bias and Variance}
\label{appendix:bias-variance}

The generalization error of TabPFN can be attributed to two primary components: \textbf{bias} and \textbf{variance}. These components are influenced by the model’s architecture, pretraining assumptions, and dataset characteristics. Mathematically, the generalization error can be decomposed by analyzing the discrepancy between the model’s predictions, \( q_\theta(y \mid x, D_n) \), and the true conditional distribution, \( p_0(y \mid x) \). Specifically, it can be expressed as:
\begin{equation}    
\begin{aligned} 
    q_{\boldsymbol{\theta}}\left(y \mid \boldsymbol{x}, \mathcal{D}_{n}\right) - p_{0}(y \mid \boldsymbol{x}) 
    =  \underbrace{q_{\boldsymbol{\theta}}\left(y \mid \boldsymbol{x}, \mathcal{D}_{n}\right) - \mathbb{E}_{\mathcal{D}_{n} \sim p_{0}^{n}}\left[q_{\boldsymbol{\theta}}\left(y \mid \boldsymbol{x}, \mathcal{D}_{n}\right)\right]}_{\text{Variance}} 
     + \underbrace{\mathbb{E}_{\mathcal{D}_{n} \sim p_{0}^{n}}\left[q_{\boldsymbol{\theta}}\left(y \mid \boldsymbol{x}, \mathcal{D}_{n}\right)\right] - p_{0}(y \mid \boldsymbol{x})}_{\text{Bias}}.
     \label{eq:bias-var}
\end{aligned}
\end{equation}
Here, \( \mathbb{E}[\cdot] \) denotes the expectation over training datasets \( D_n \) sampled from the true distribution \( p_0 \). Understanding these error components is crucial for addressing the limitations of TabPFN and improving its performance across diverse tabular datasets.

Revisiting the results shown in Figure 2, we observe several key phenomena that highlight the effectiveness of~\name~in reducing generalization error across varying dataset sizes:

\textbf{Lowest Generalization Error Across All Dataset Sizes:}  
Our method consistently achieves the smallest generalization error, demonstrating robust performance regardless of dataset size.

\textbf{Impact of Fine-Tuning on Small Datasets:}  
For small datasets (e.g., with only a few hundred samples), fine-tuning the encoder may slightly increase bias. This occurs because the limited data volume can lead to overfitting during fine-tuning.

\textbf{Mechanisms of Bias and Variance Reduction:}  
The bias reduction is primarily attributed to the fine-tuned encoder, which aligns TabPFN with the downstream dataset distribution, improving compatibility between the pre-trained model and the target task. Variance reduction, on the other hand, is not solely due to Bagging. The learning of multiple encoders also contributes significantly. When dataset sizes are small (e.g., fewer than 1000 samples), where each bootstrap sample covers the entire training set, training multiple encoders still introduces diversity in representations, further reducing variance.

These findings highlight the effectiveness of~\name~in addressing the bias-variance tradeoff, making it a robust and scalable solution for real-world tabular learning.


\subsection{Ablation Study}
\label{appendix:ablation_study}
\textbf{Influence of Encoder Path Count on Performance Improvement over TabPFN.}~\autoref{fig:relative_improvement} (a) presents the results of an ablation experiment designed to investigate the impact of different encoder path counts on the performance of the model. The primary goal is to explore how the number of encoder paths influences the model’s relative improvement over TabPFN. We display the relative improvement for various methods with different encoder path configurations. From the plot, we observe that as the number of encoder paths (denoted by $K$) increases, the relative performance improvement of~\name~over TabPFN becomes more pronounced. This trend indicates that a higher number of encoder paths contributes positively to the model's ability to outperform TabPFN, suggesting that more encoder paths enhance the model’s expressive power and its ability to capture more complex patterns in the data.

\textbf{Impact of Periodic Activation, Batch Ensemble, and Partial Parameter Tuning on Model Performance.}~\autoref{fig:relative_improvement}~(b) visualizes the effect of three key components—periodic activation, Batch Ensemble, and partial parameter tuning—on the relative performance improvement of~\name~over TabPFN. In this experiment, we compare the original~\name~with three variations: one without periodic activation, one without Batch Ensemble, and one with full parameter tuning. The plot reveals that each of these components contributes positively to the model’s performance. Specifically, removing any of these elements leads to a noticeable decline in relative improvement over TabPFN, highlighting that all parts are essential for achieving the best performance. Notably, the variation without partial parameter tuning (PPT) shows the highest upper-bound improvement, but it also incurs significantly higher computational costs. Moreover, the median relative improvement decreases compared to~\name~, further emphasizing the superior efficiency of our approach. This suggests that periodic activation, Batch Ensemble, and partial parameter tuning each play a crucial role in enhancing the model's ability to outperform TabPFN, and the absence of any of these components compromises the model's ability to leverage its full potential while also increasing computational resource usage. \looseness=-1
\begin{figure}[H]
\begin{minipage}{0.48\linewidth}
    \includegraphics[width=\textwidth]{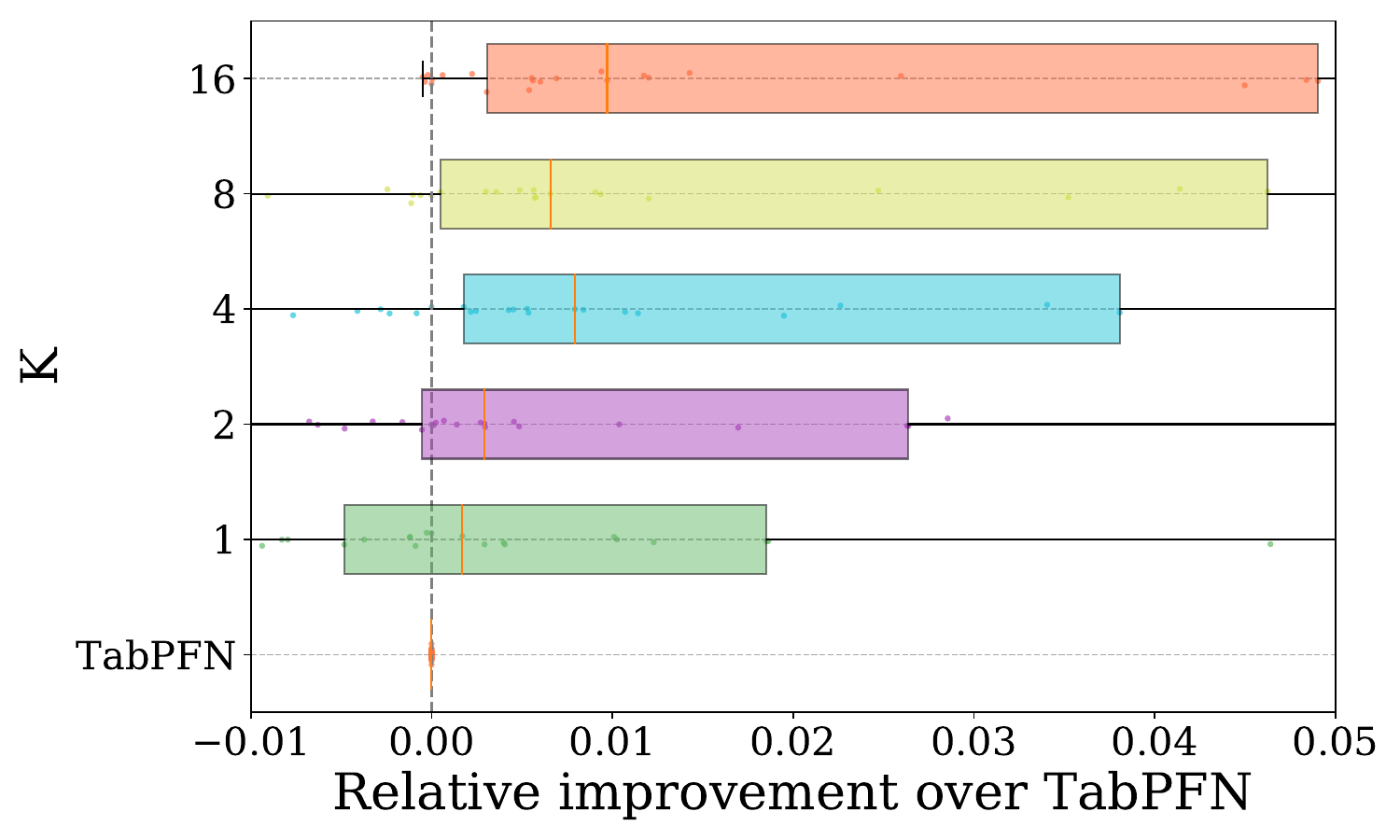}
    \centering
    {\small \mbox{(a) {Infulence of K}}}
    \end{minipage}
    \begin{minipage}{0.48\linewidth}
    \includegraphics[width=\textwidth]{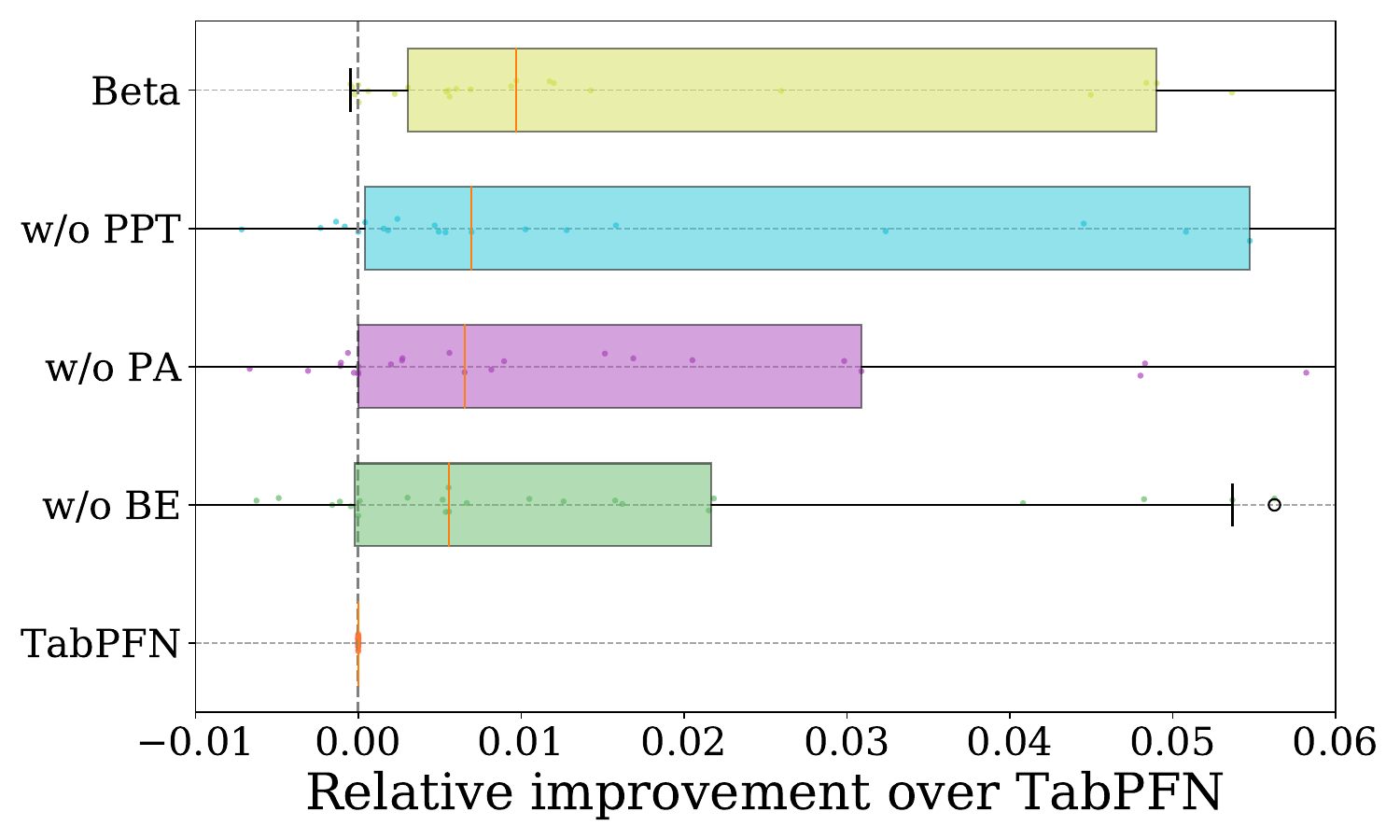}
    \centering
    {\small \mbox{(b) {Influence of some important components}}}
    \end{minipage}

    \caption{(a) The box plot demonstrates the effect of varying the number of encoder paths, $K$, on the relative improvement over TabPFN. The horizontal axis represents the relative improvement in performance, while the vertical axis lists numbers of different encoder path settings; (b) This box plot illustrates the effect of different components—periodic activation (PA), Batch Ensemble (BE), and partial parameter tuning (PPT)—on the relative improvement of~\name~ over TabPFN. The plot compares the original \name model with three variations: (1)~\name~with partial parameter tuning (w/o PPT), (2)~\name~without periodic activation (w/o PA), and (3)~\name~without Batch Ensemble (w/o BE). }
\end{figure}

\subsection{Comparison of inference time and the number of learnable parameters.}
\label{appendix:efficiency}
We compared the inference time (in seconds) of~\name~with other methods, as well as the average rank in the main experiment. Additionally, the number of learnable parameters was roughly estimated by the size of the stored checkpoint. The results confirm that~\name~not only delivers state-of-the-art classification accuracy but also maintains low inference time and reduced memory overhead, making it highly scalable and efficient for various tabular classification tasks.
\begin{table}[ht]
\centering
\caption{Comparison of Inference Time, Average Rank, and Number of Learnable Parameters. All metrics are computed as the mean values across multiple datasets.}
\begin{tabular}{lcccccc}
\hline
\textbf{Metric} & \textbf{\name} &\textbf{TabPFN} & \textbf{LocalPFN}  & \textbf{MixturePFN} & \textbf{FT-T}& \textbf{TabR}\\
\hline
\textbf{Inference Time (s)} & 0.91 & 0.78 & 12.51 &3.24 & \textbf{0.36} & 0.38 \\
\textbf{Average Rank} & \textbf{6.76} & 15.63 & 12.15 & 12.86 & 13.27 & 9.26\\
\textbf{Checkpoint Size (MB)} &  \textbf{0.60} & 0 & 98.65  &21.42 & 7.15  & 14.07  \\
\hline
\end{tabular}
\label{tab:comparison_transposed}
\end{table}

\subsection{Visialzation results}

\begin{figure*}[t]
    \begin{minipage}{0.24\linewidth}
    \includegraphics[width=\textwidth]{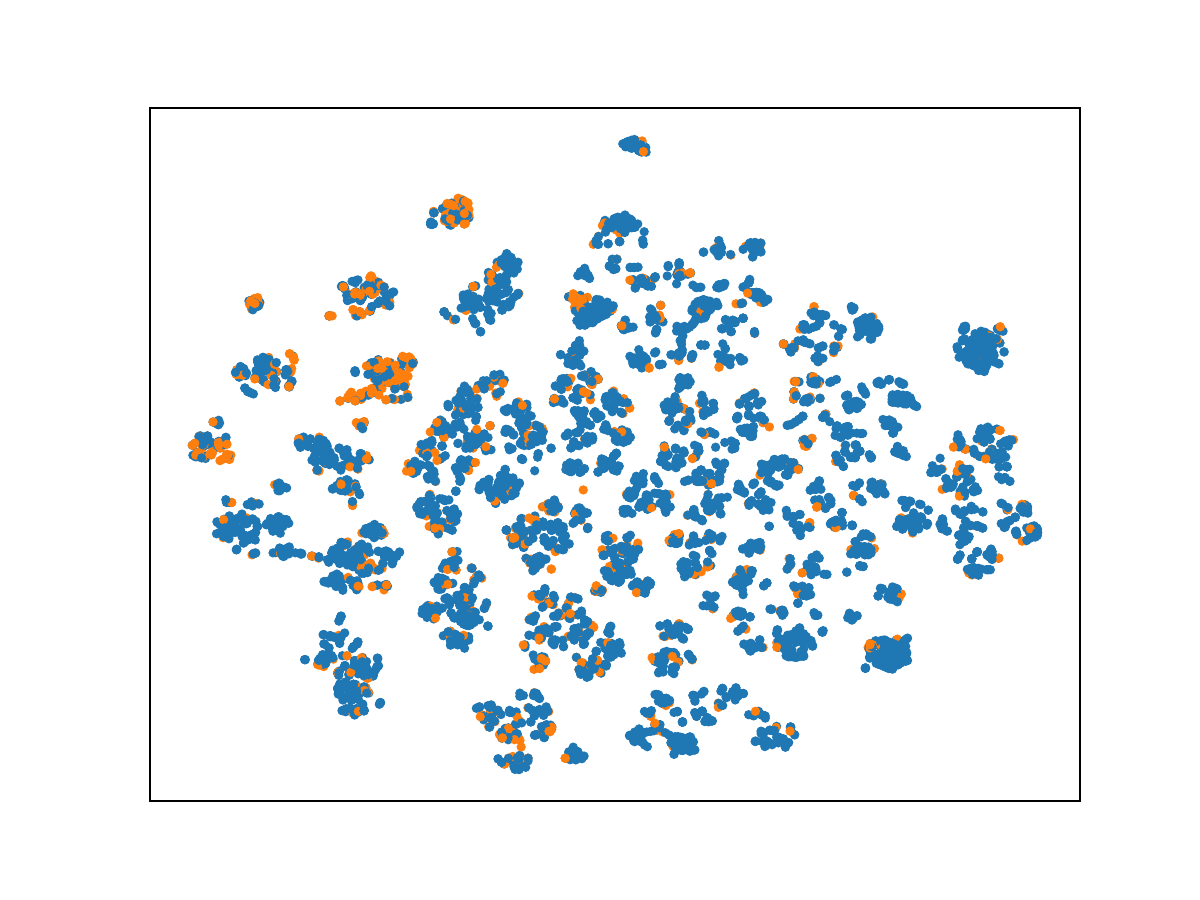}
    \centering
    {\small \mbox{(a) {Bank  Raw Feature}}}
    \end{minipage}
    \begin{minipage}{0.24\linewidth}
    \includegraphics[width=\textwidth]{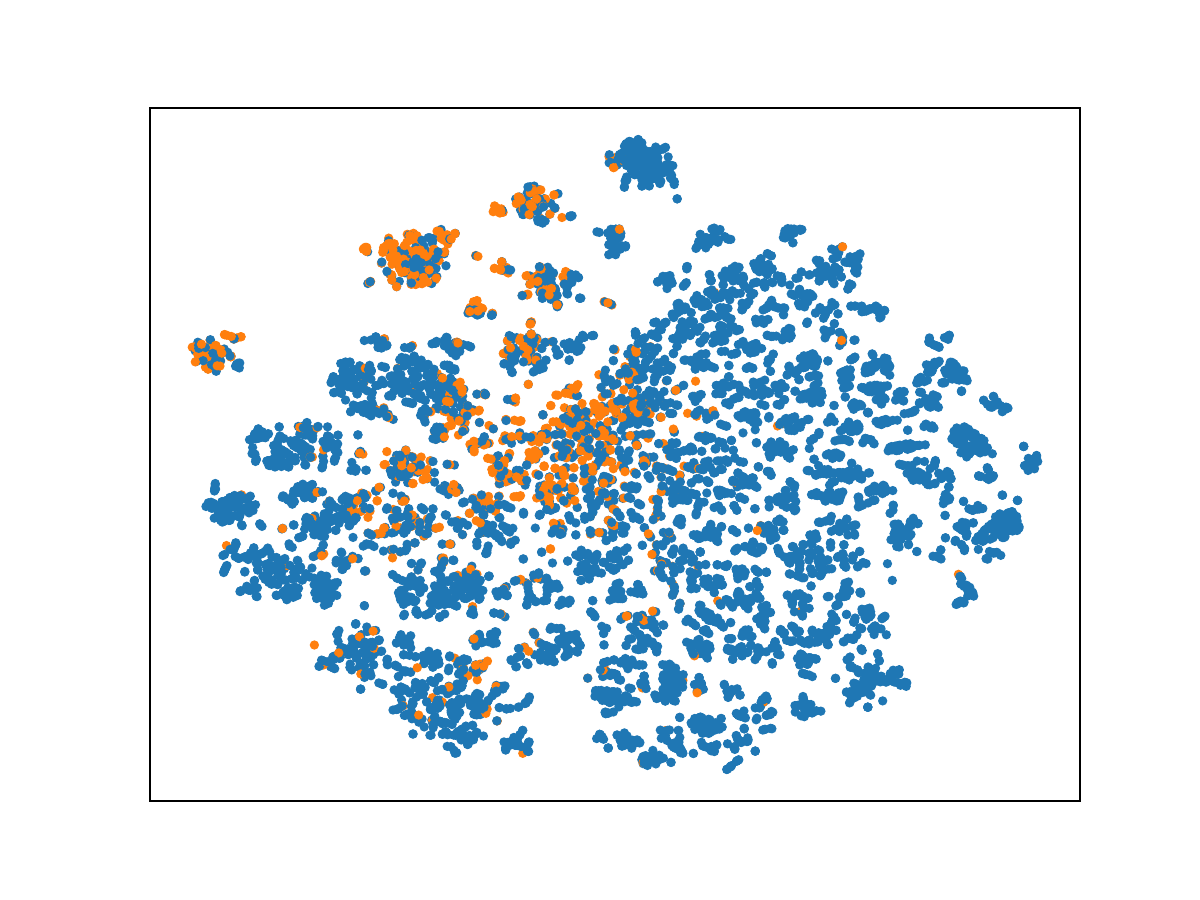}
    \centering
    {\small \mbox{(b) {Bank  MLP}}}
    \end{minipage}
    \begin{minipage}{0.24\linewidth}
    \includegraphics[width=\textwidth]{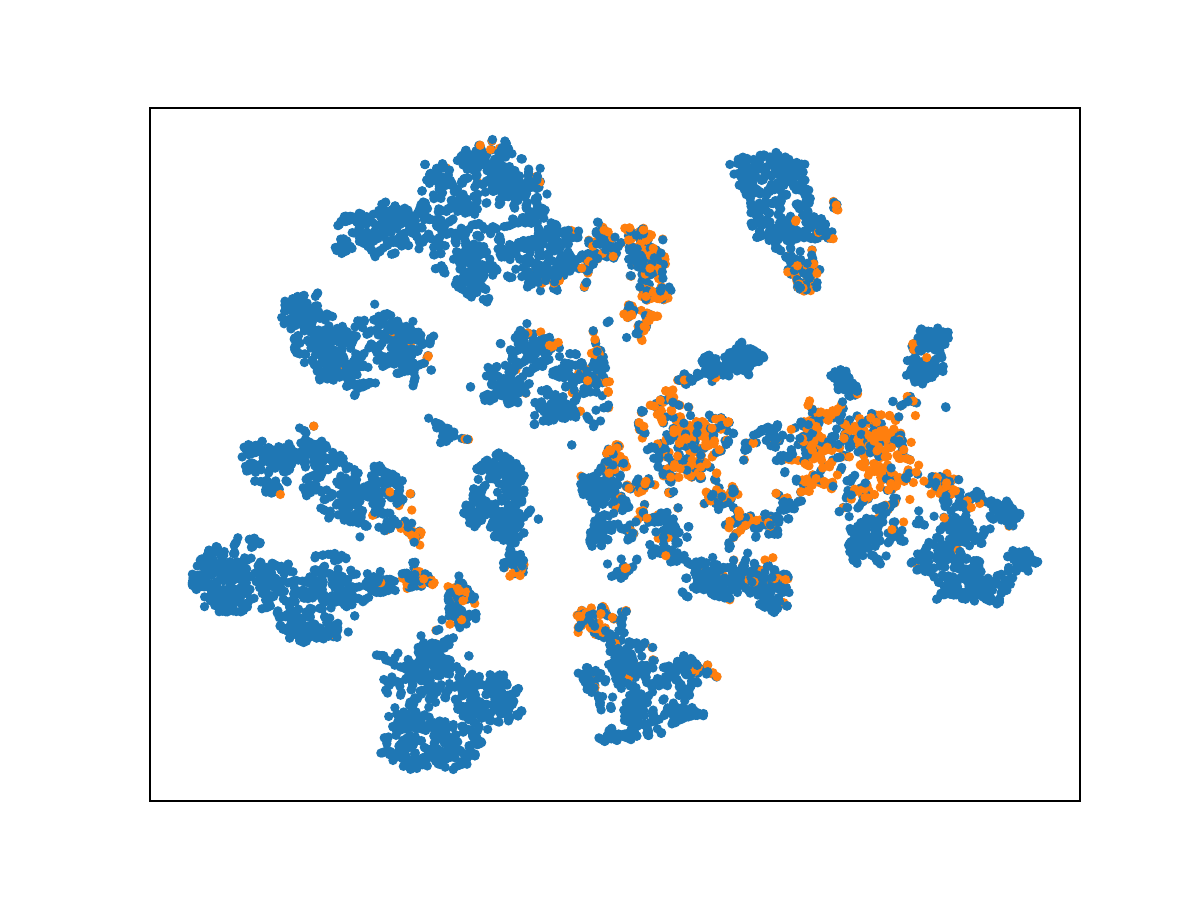}
    \centering
    {\small \mbox{(c) {Bank  ModernNCA}}}
    \end{minipage}
    \begin{minipage}{0.24\linewidth}
    \includegraphics[width=\textwidth]{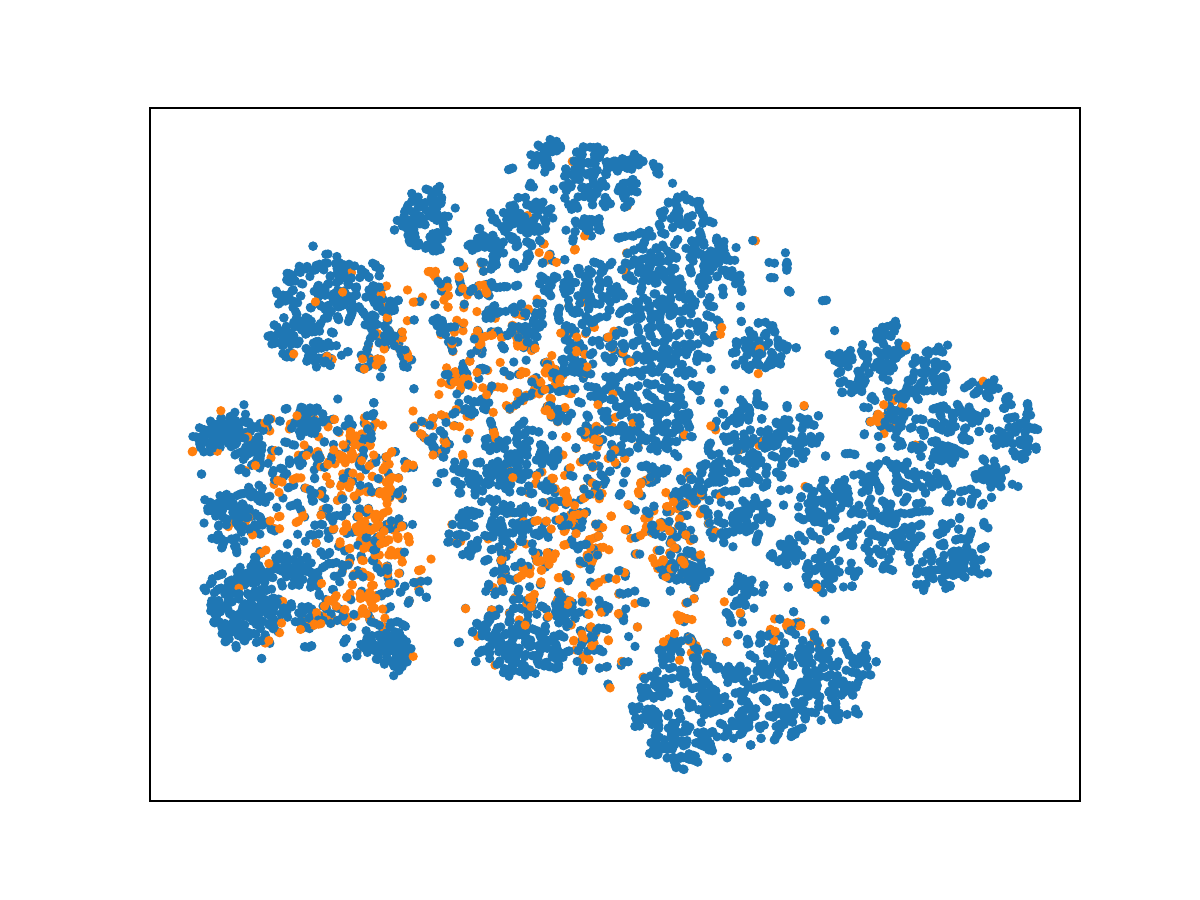}
    \centering
    {\small \mbox{(d) {Bank  TabR}}}
    \end{minipage} 

    \begin{minipage}{0.24\linewidth}
    \includegraphics[width=\textwidth]{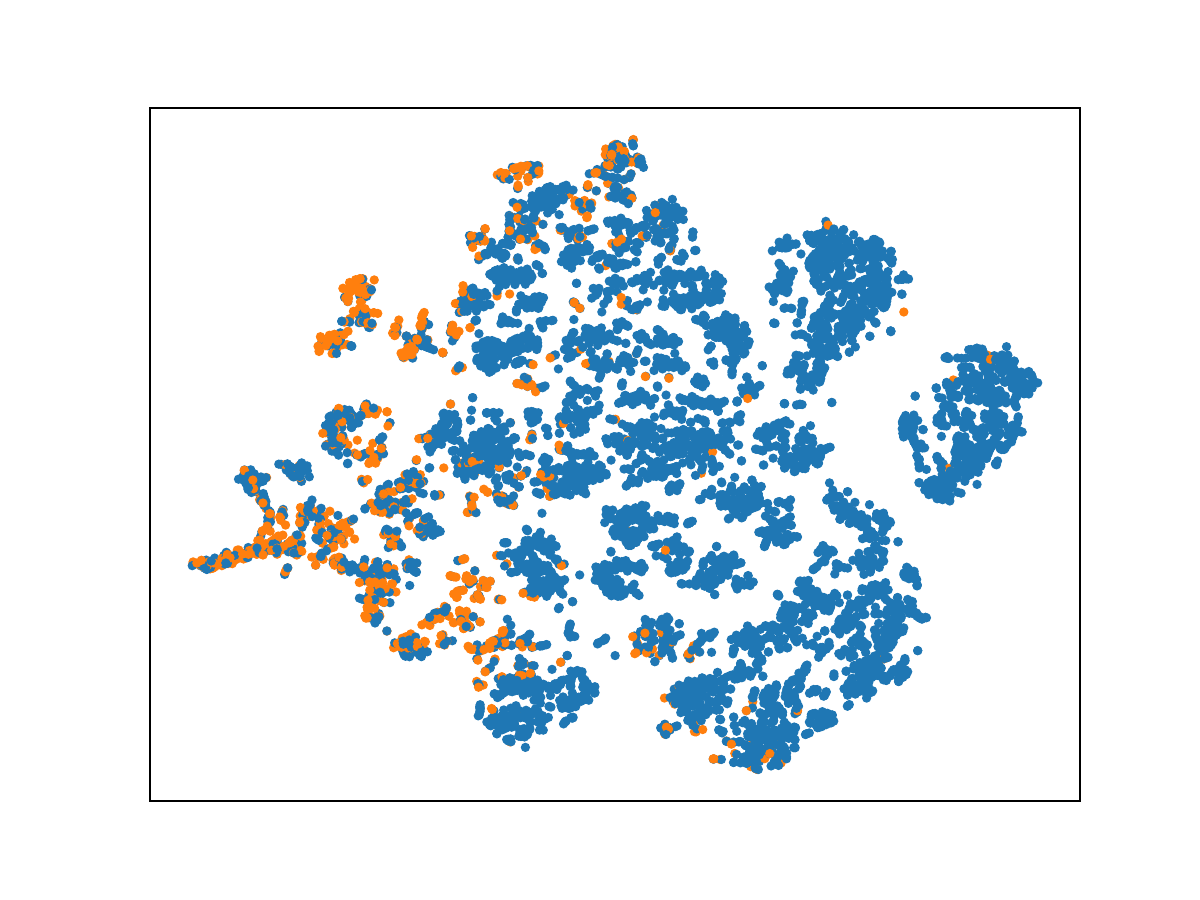}
    \centering
    {\small \mbox{(e) {Bank  \name~1}}}
    \end{minipage}
    \begin{minipage}{0.24\linewidth}
    \includegraphics[width=\textwidth]{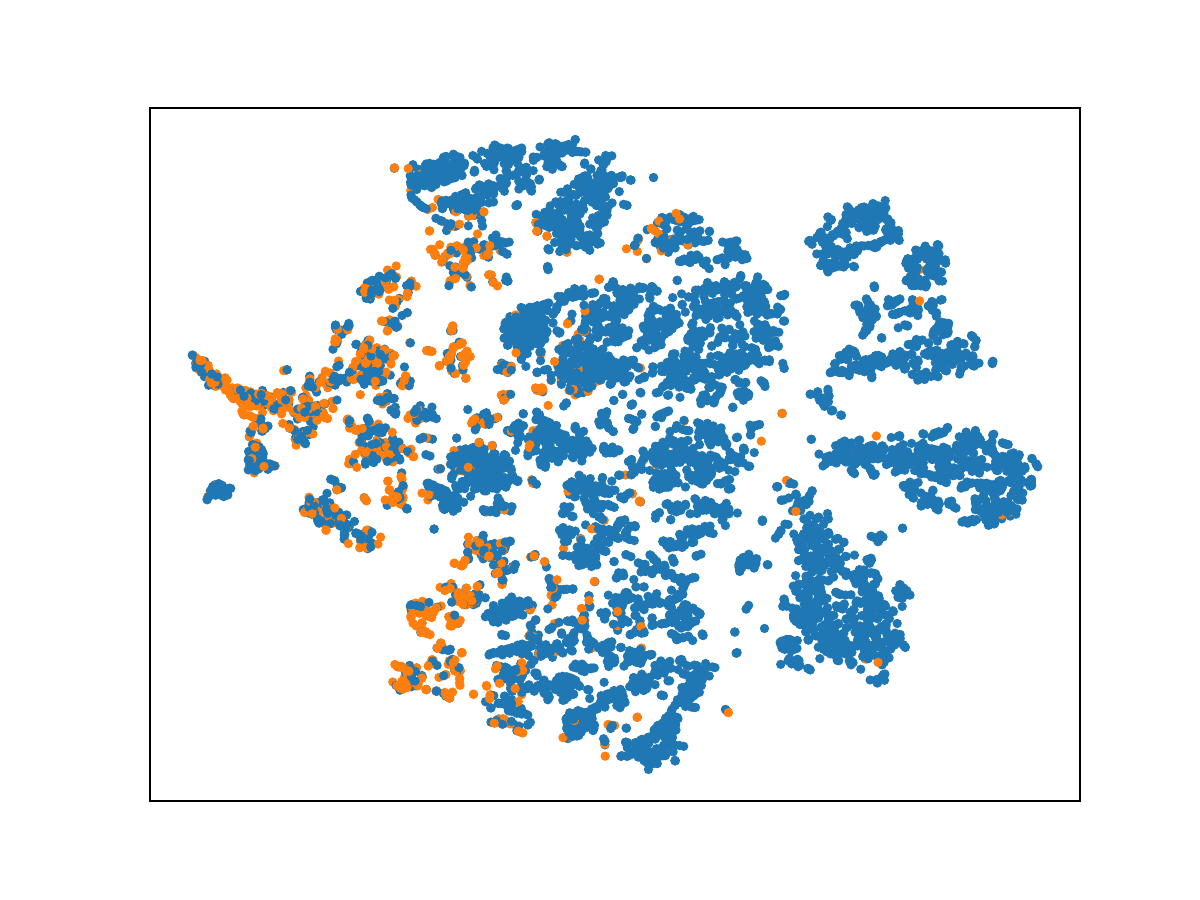}
    \centering
    {\small \mbox{(f) {Bank  \name~2}}}
    \end{minipage}
    \begin{minipage}{0.24\linewidth}
    \includegraphics[width=\textwidth]{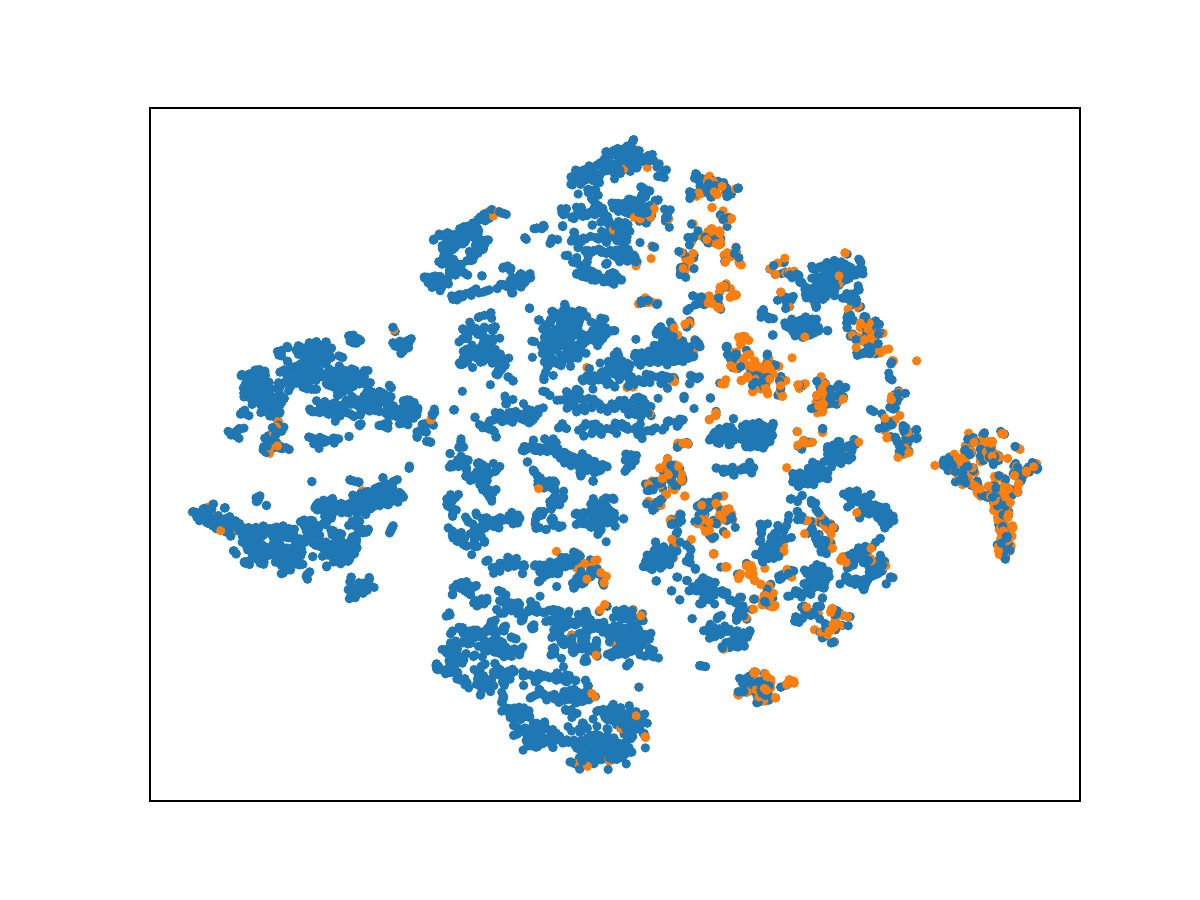}
    \centering
    {\small \mbox{(g) {Bank  \name~3}}}
    \end{minipage}
    \begin{minipage}{0.24\linewidth}
    \includegraphics[width=\textwidth]{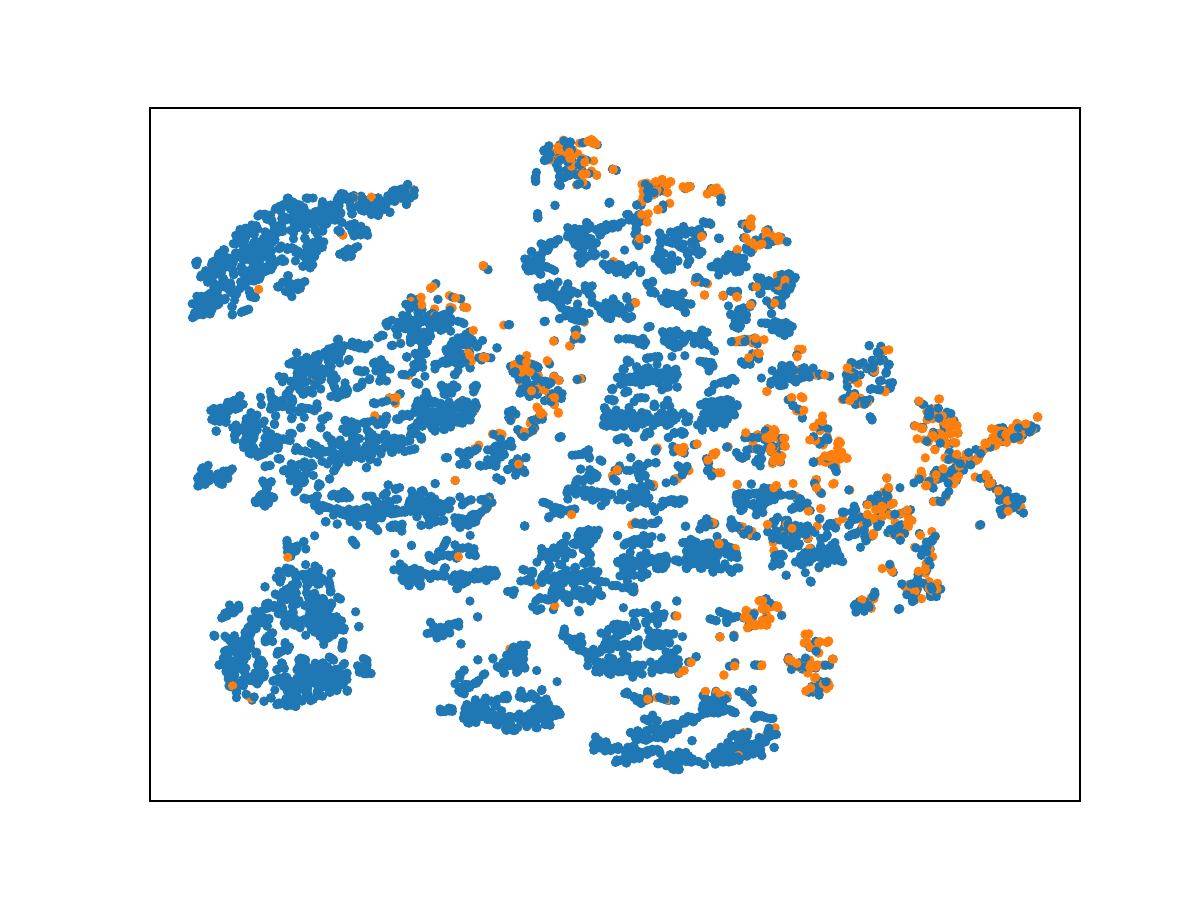}
    \centering
    {\small \mbox{(h) {Bank  \name~4}}}
    \end{minipage}

    \begin{minipage}{0.24\linewidth}
    \includegraphics[width=\textwidth]{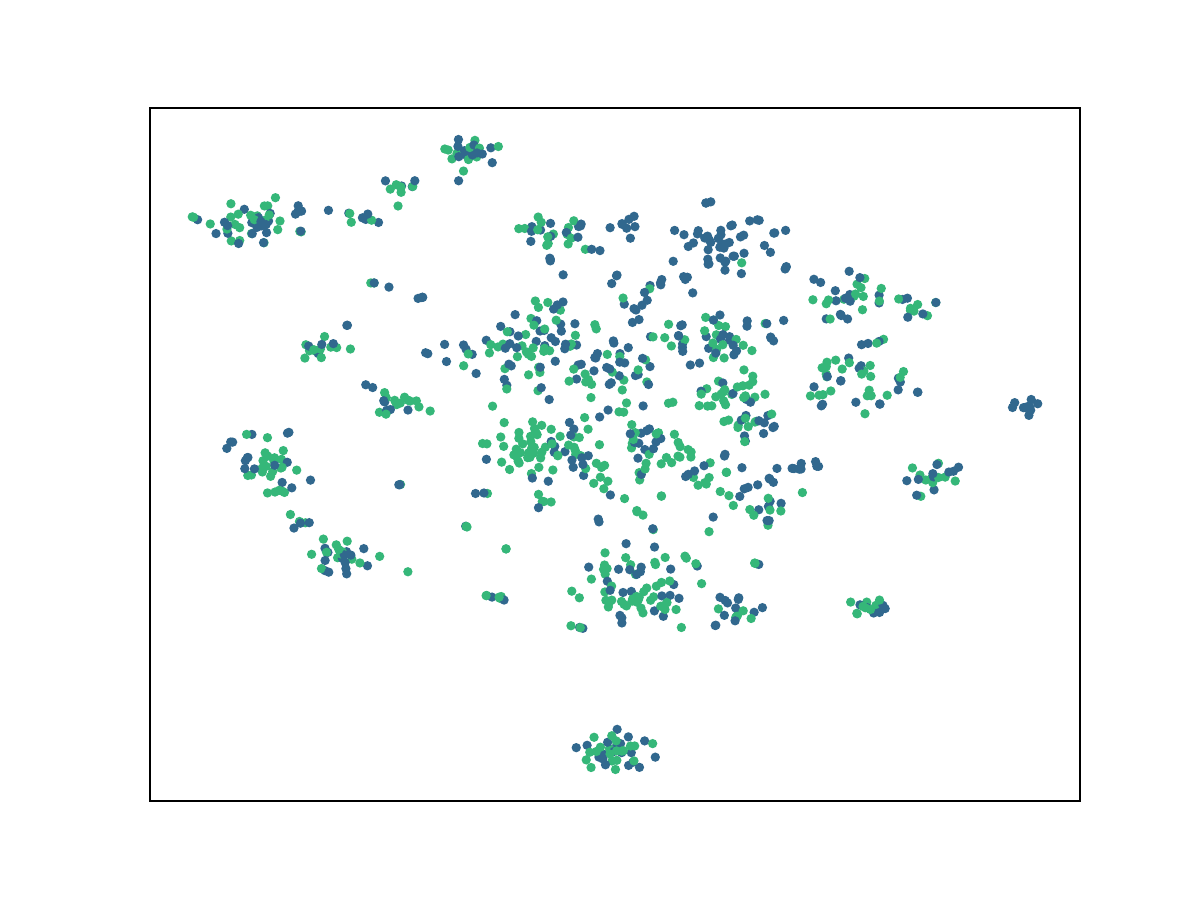}
    \centering
    {\small \mbox{(a) {KDD  Raw Feature}}}
    \end{minipage}
    \begin{minipage}{0.24\linewidth}
    \includegraphics[width=\textwidth]{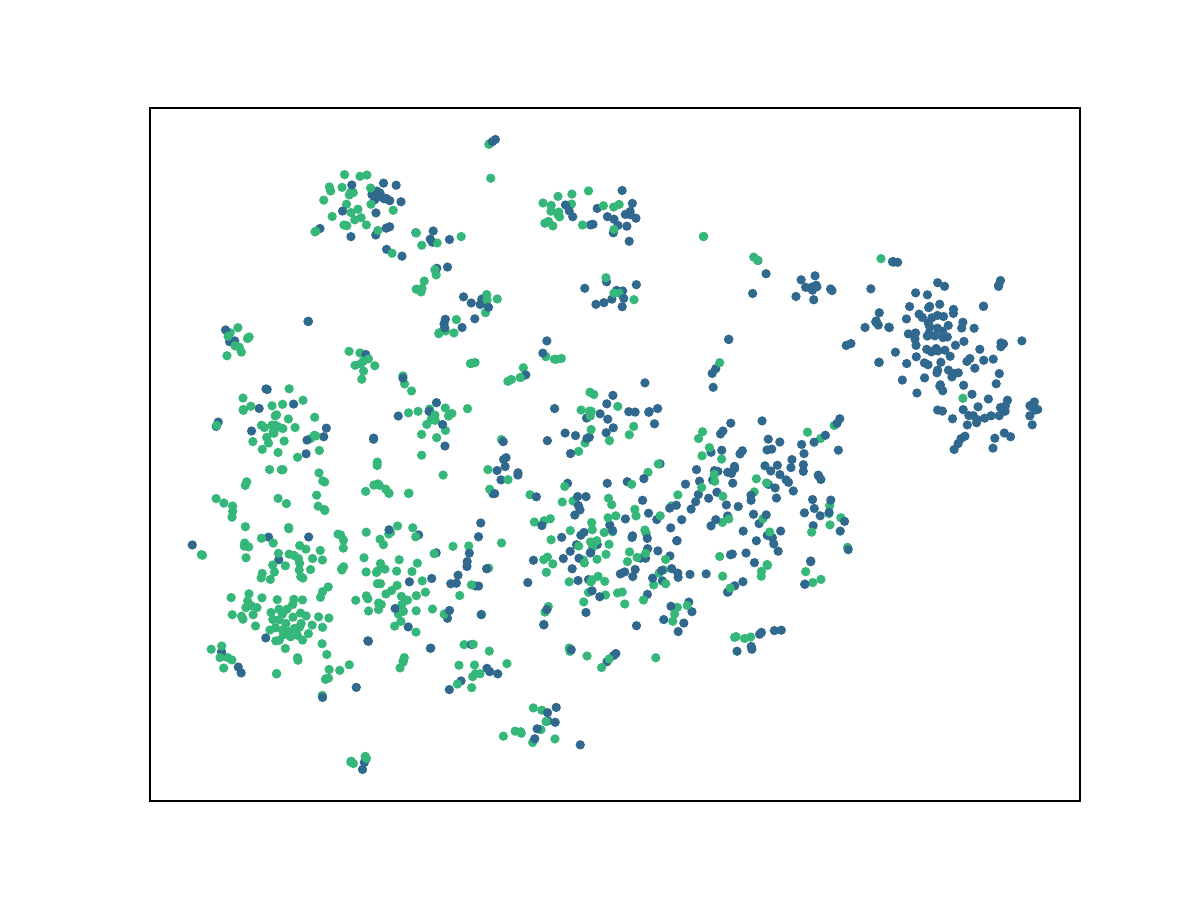}
    \centering
    {\small \mbox{(b) {KDD  MLP}}}
    \end{minipage}
    \begin{minipage}{0.24\linewidth}
    \includegraphics[width=\textwidth]{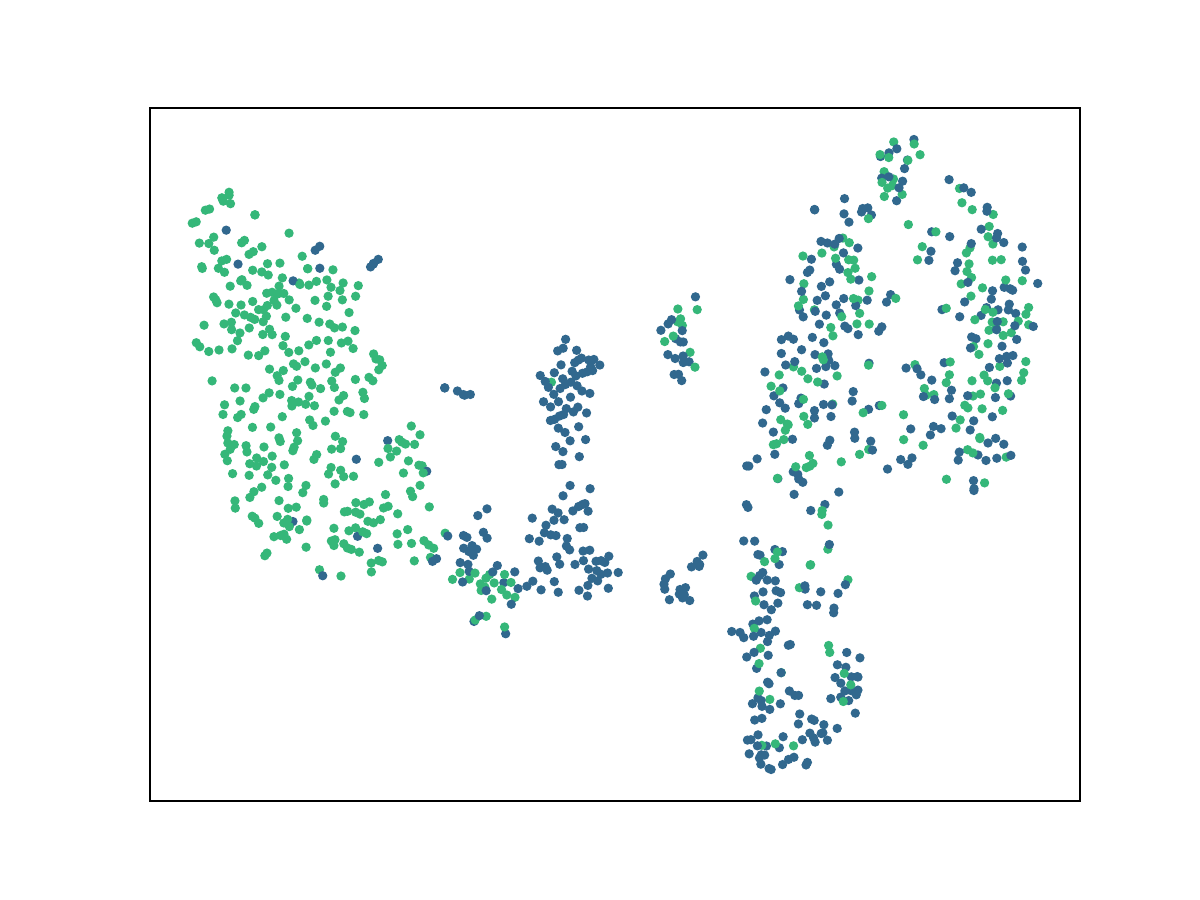}
    \centering
    {\small \mbox{(c) {KDD  ModernNCA}}}
    \end{minipage}
    \begin{minipage}{0.24\linewidth}
    \includegraphics[width=\textwidth]{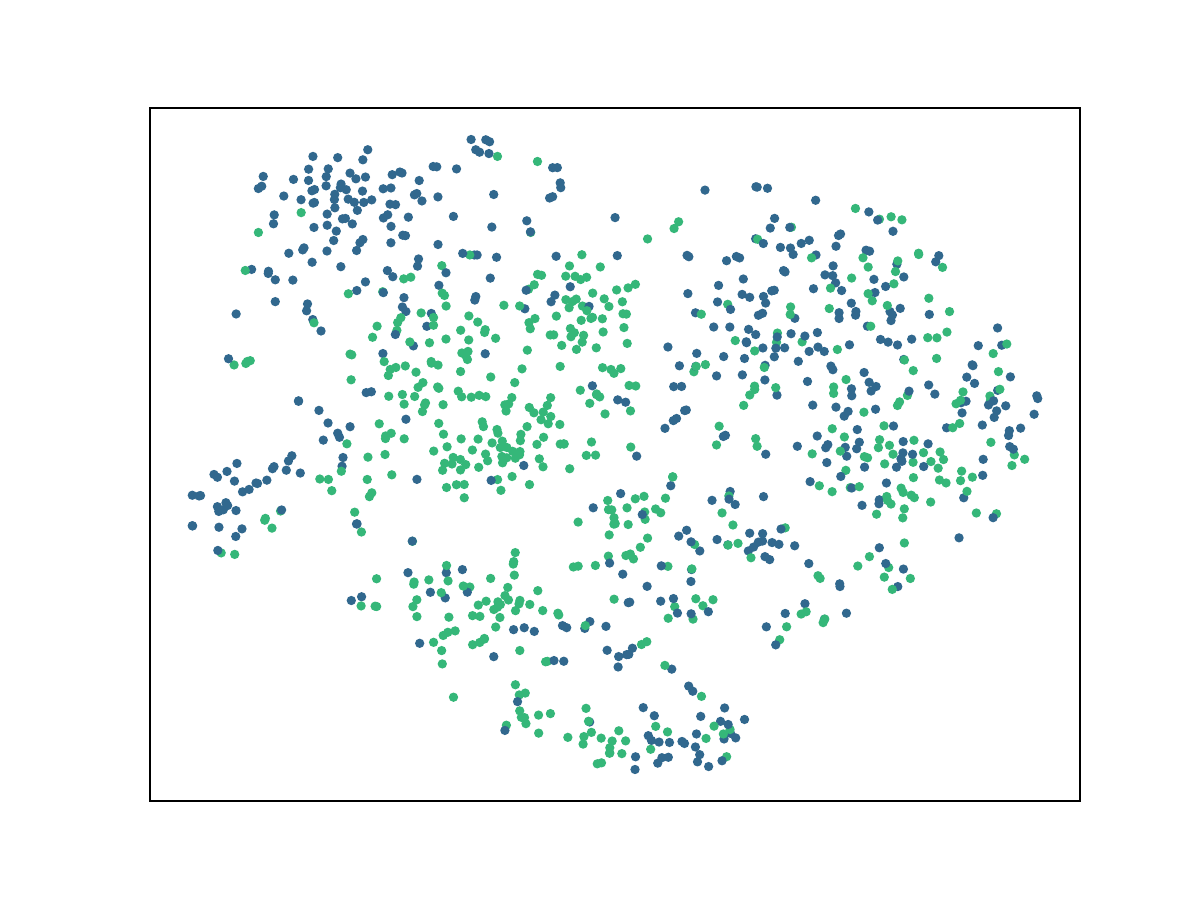}
    \centering
    {\small \mbox{(d) {KDD  TabR}}}
    \end{minipage} 

    \begin{minipage}{0.24\linewidth}
    \includegraphics[width=\textwidth]{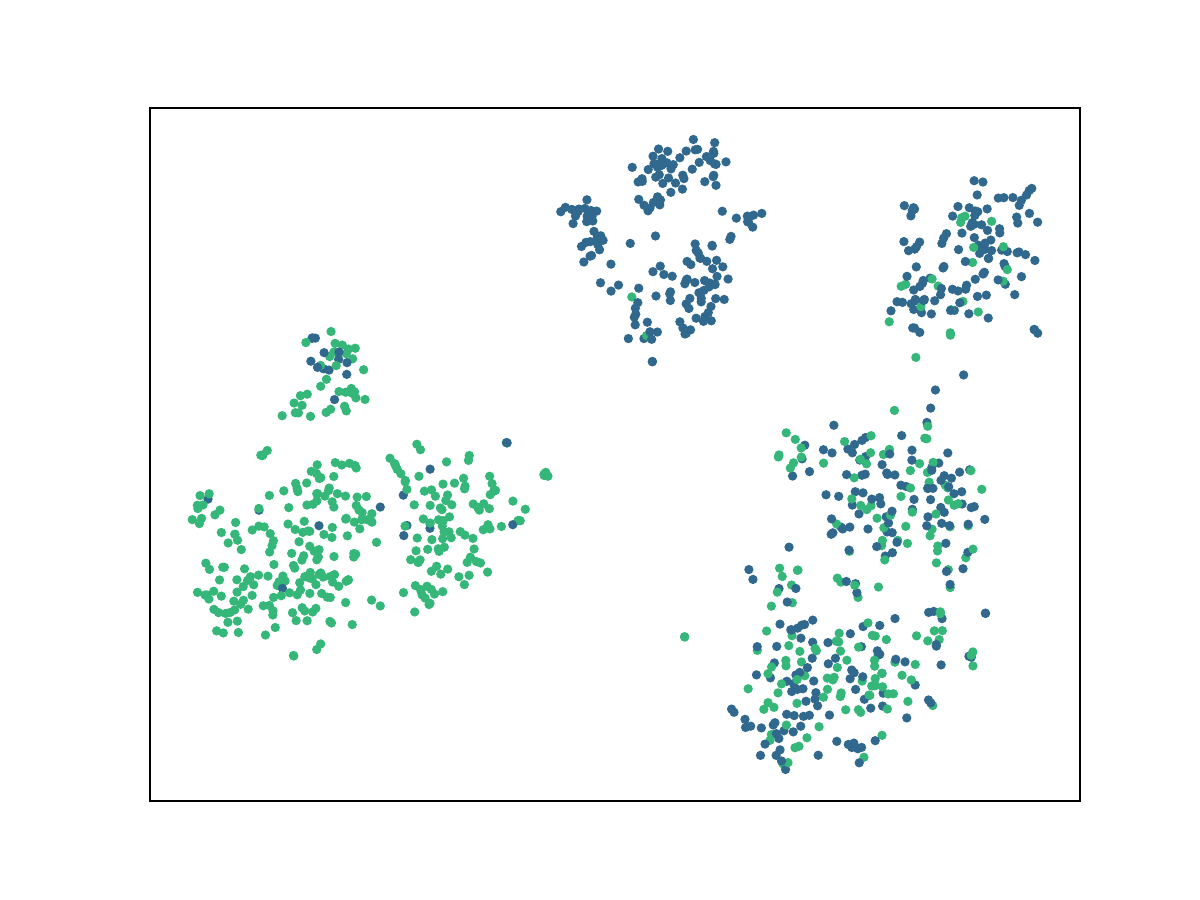}
    \centering
    {\small \mbox{(e) {KDD  \name~1}}}
    \end{minipage}
    \begin{minipage}{0.24\linewidth}
    \includegraphics[width=\textwidth]{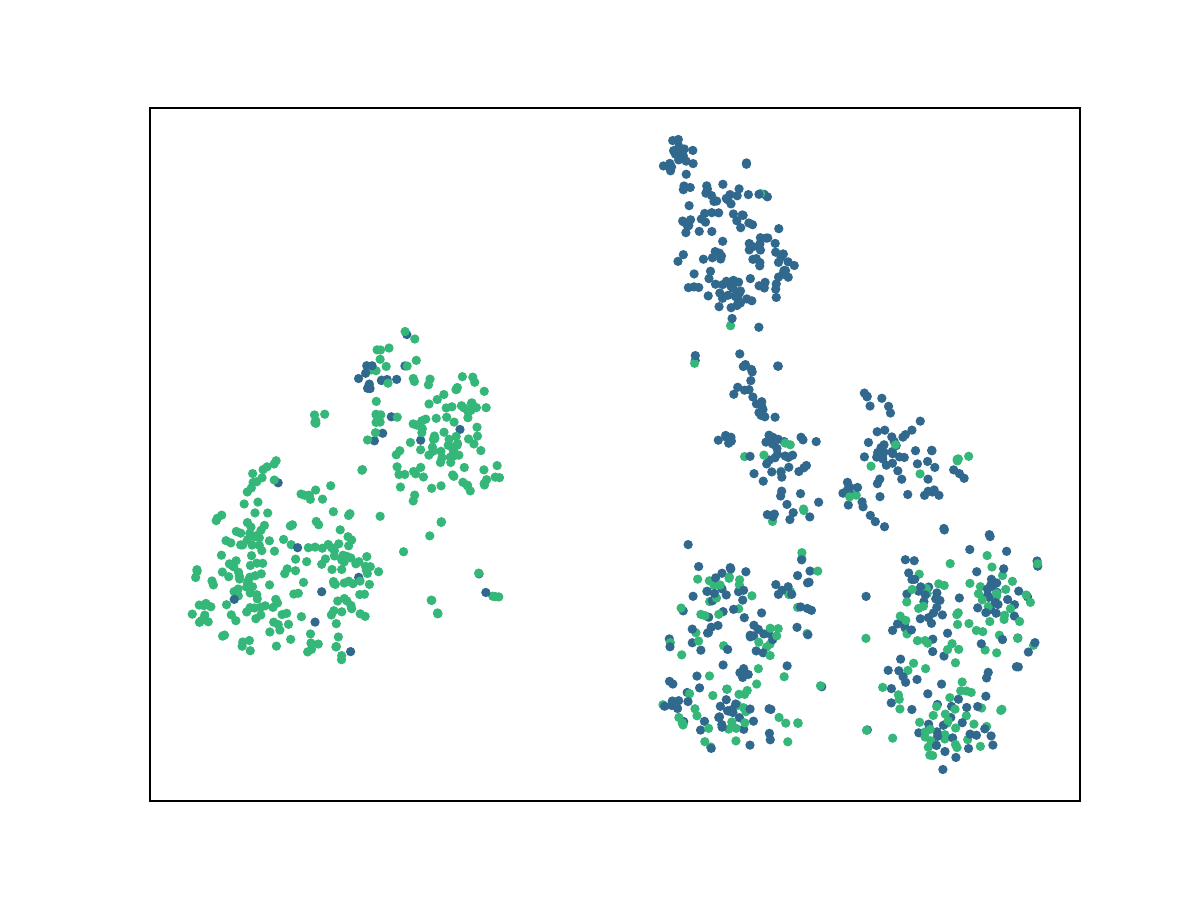}
    \centering
    {\small \mbox{(f) {KDD  \name~2}}}
    \end{minipage}
    \begin{minipage}{0.24\linewidth}
    \includegraphics[width=\textwidth]{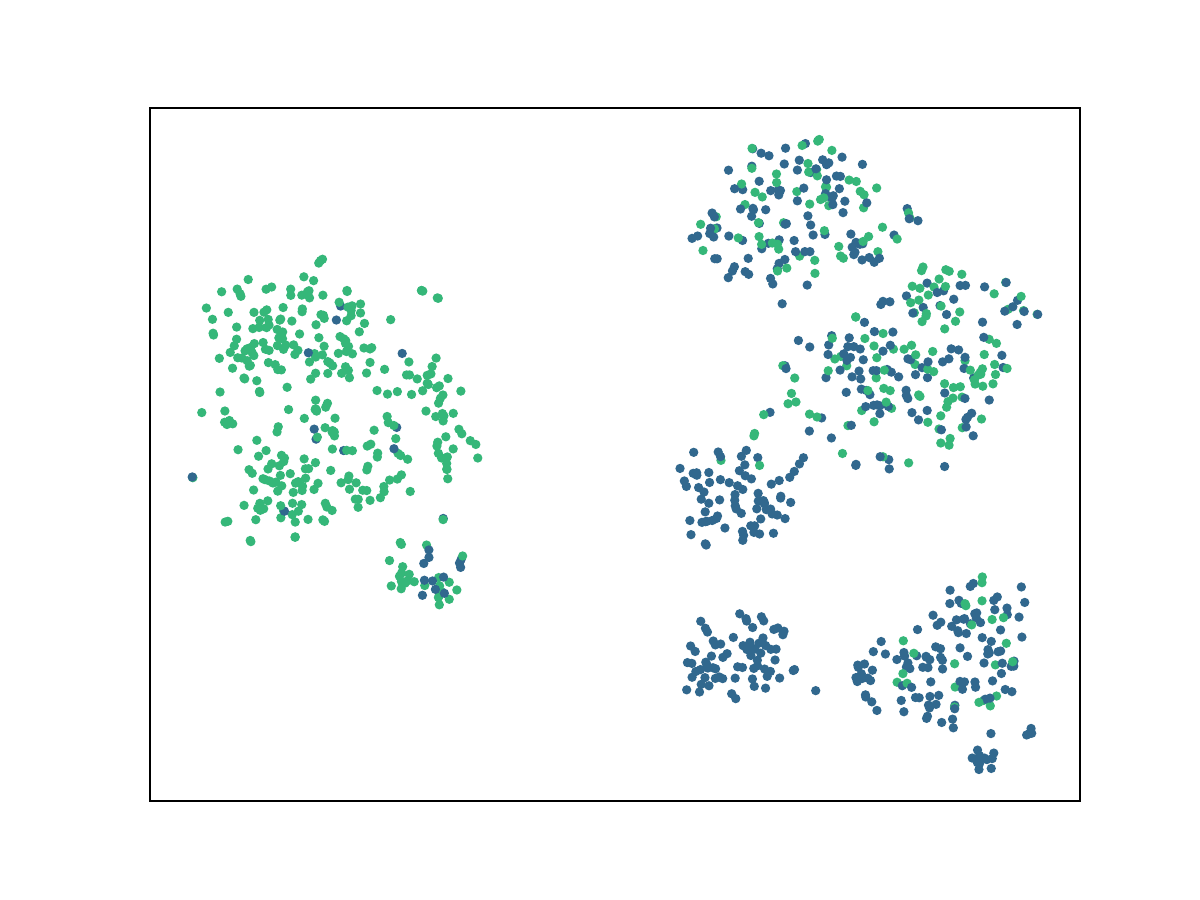}
    \centering
    {\small \mbox{(g) {KDD  \name~3}}}
    \end{minipage}
    \begin{minipage}{0.24\linewidth}
    \includegraphics[width=\textwidth]{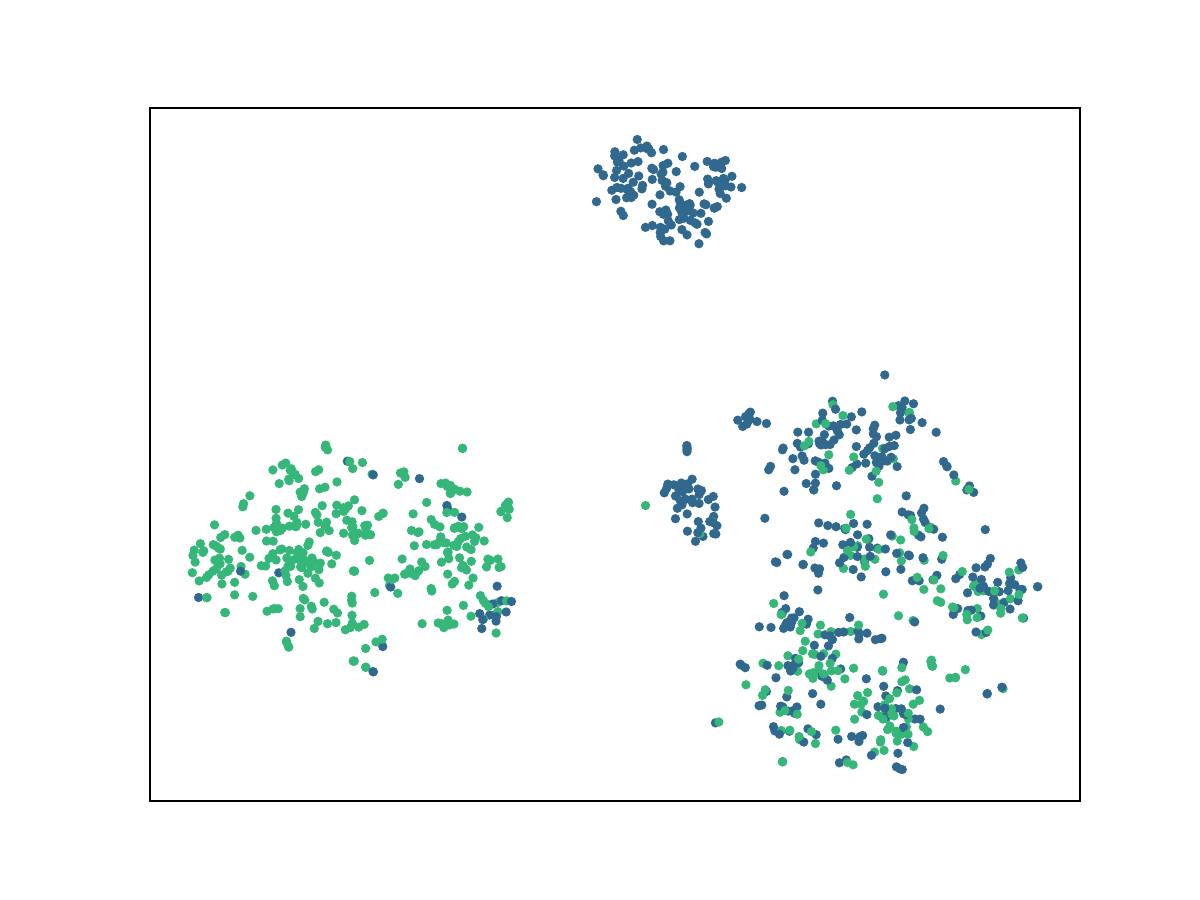}
    \centering
    {\small \mbox{(h) {KDD  \name~4}}}
    \end{minipage}

    \begin{minipage}{0.24\linewidth}
    \includegraphics[width=\textwidth]{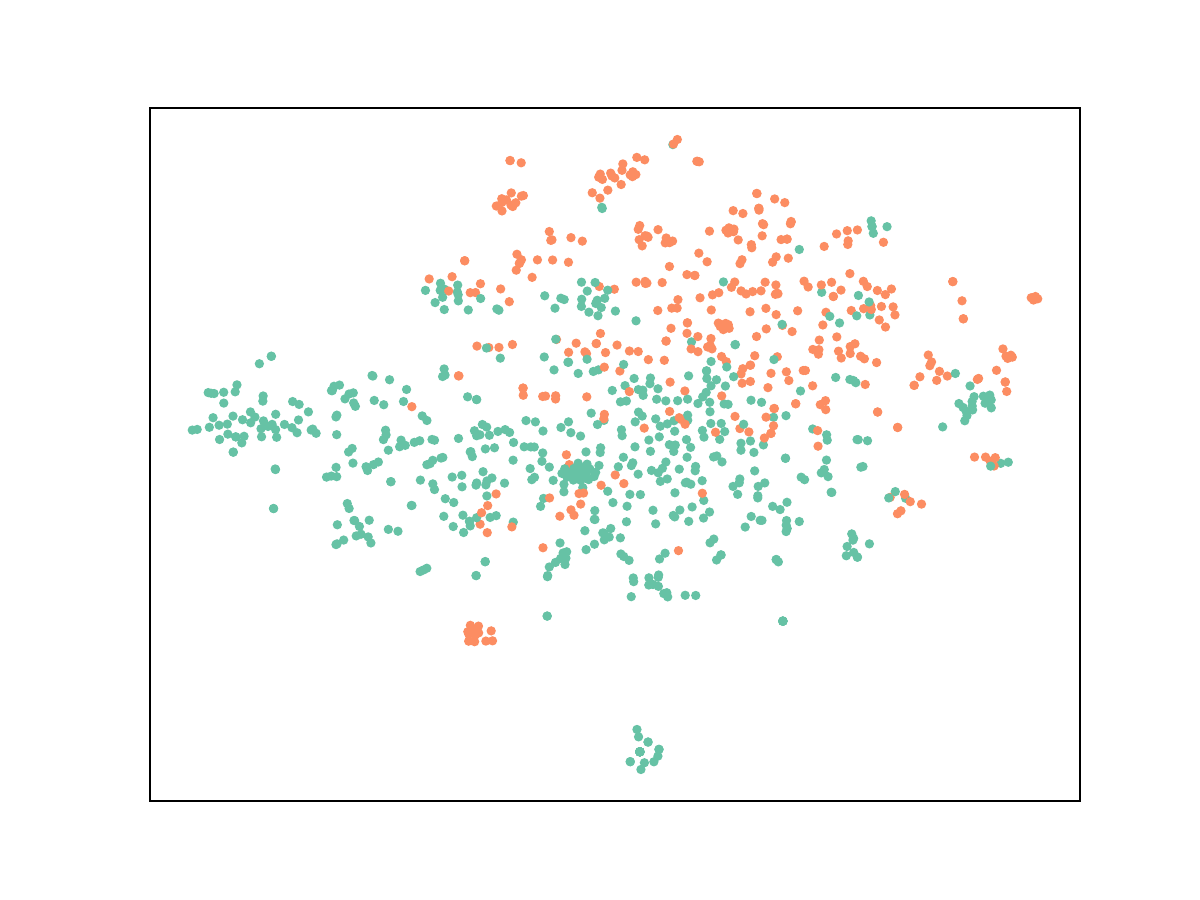}
    \centering
    {\small \mbox{(a) {Spambase Raw Feature}}}
    \end{minipage}
    \begin{minipage}{0.24\linewidth}
    \includegraphics[width=\textwidth]{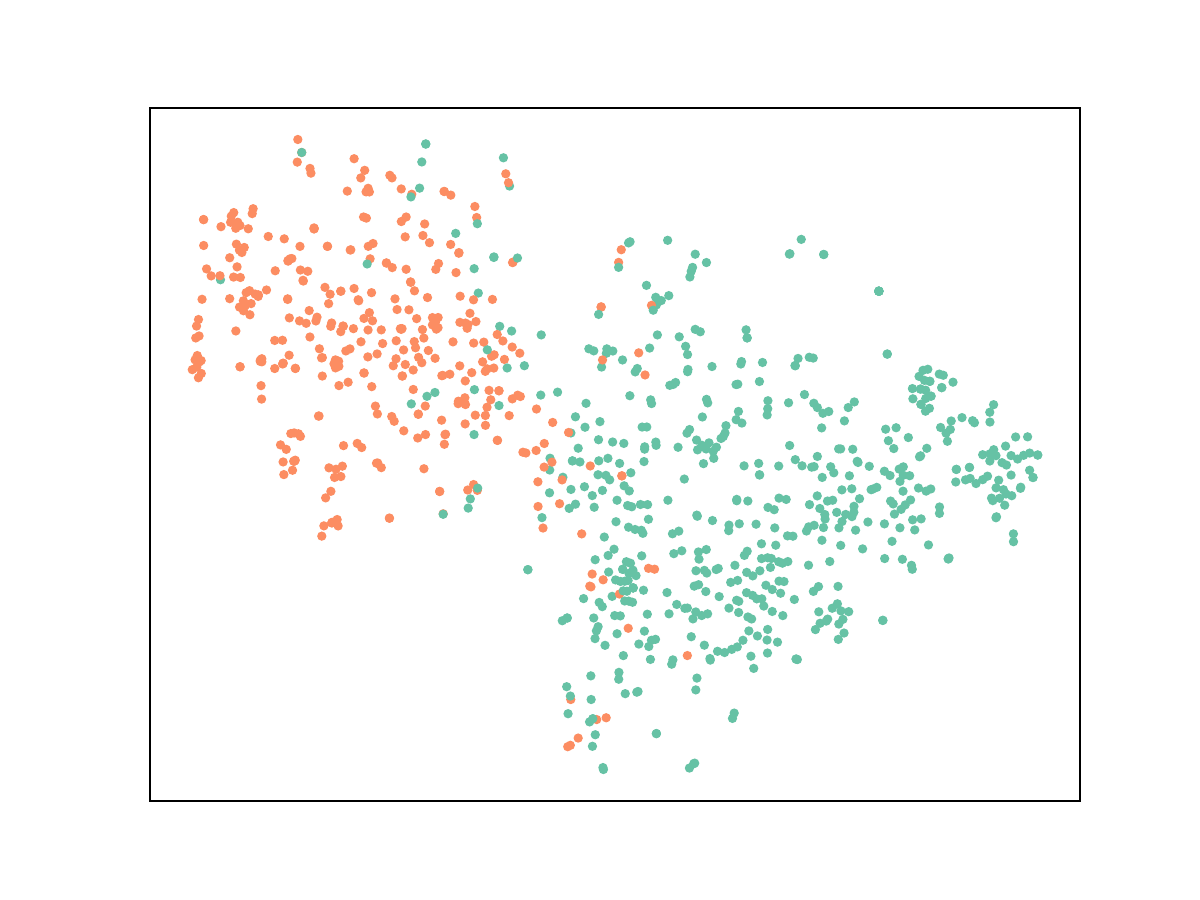}
    \centering
    {\small \mbox{(b) {Spambase MLP}}}
    \end{minipage}
    \begin{minipage}{0.24\linewidth}
    \includegraphics[width=\textwidth]{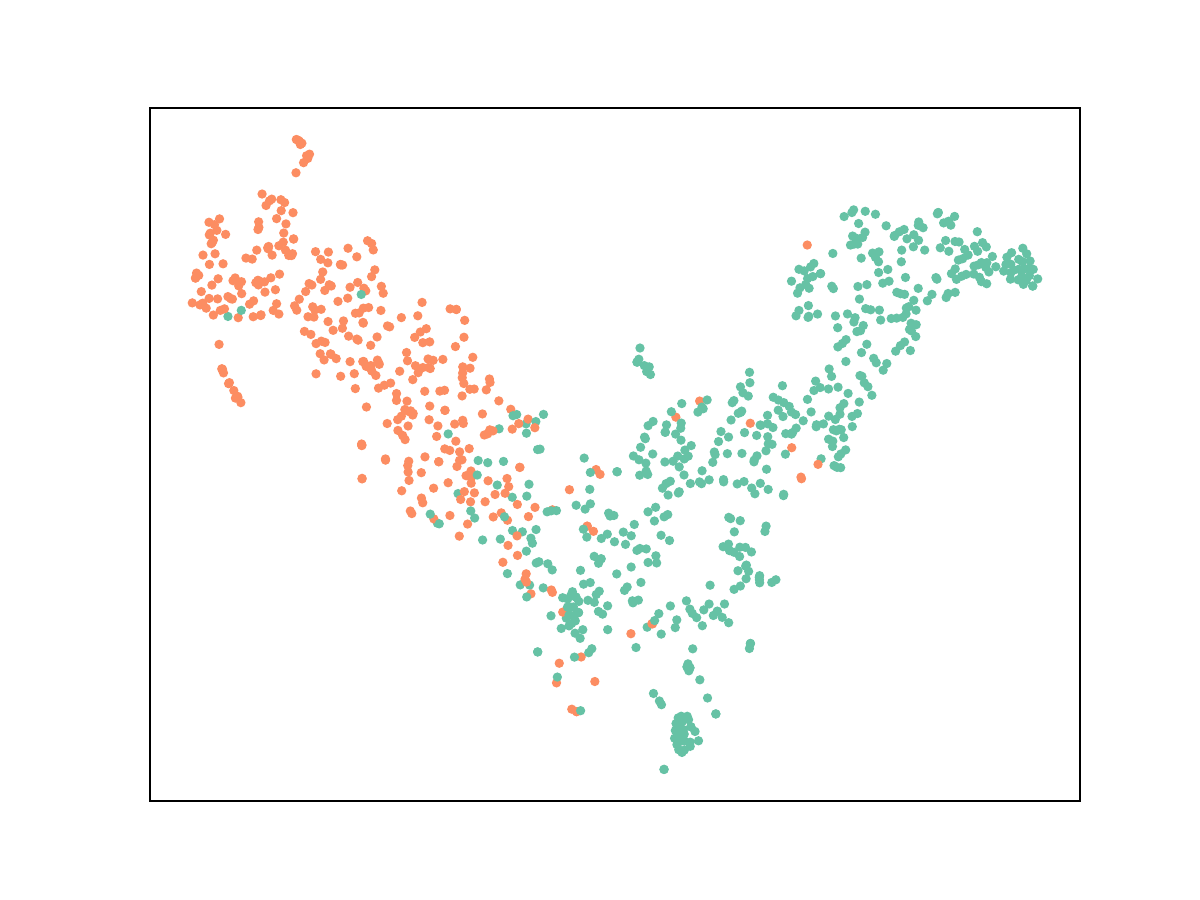}
    \centering
    {\small \mbox{(c) {Spambase ModernNCA}}}
    \end{minipage}
    \begin{minipage}{0.24\linewidth}
    \includegraphics[width=\textwidth]{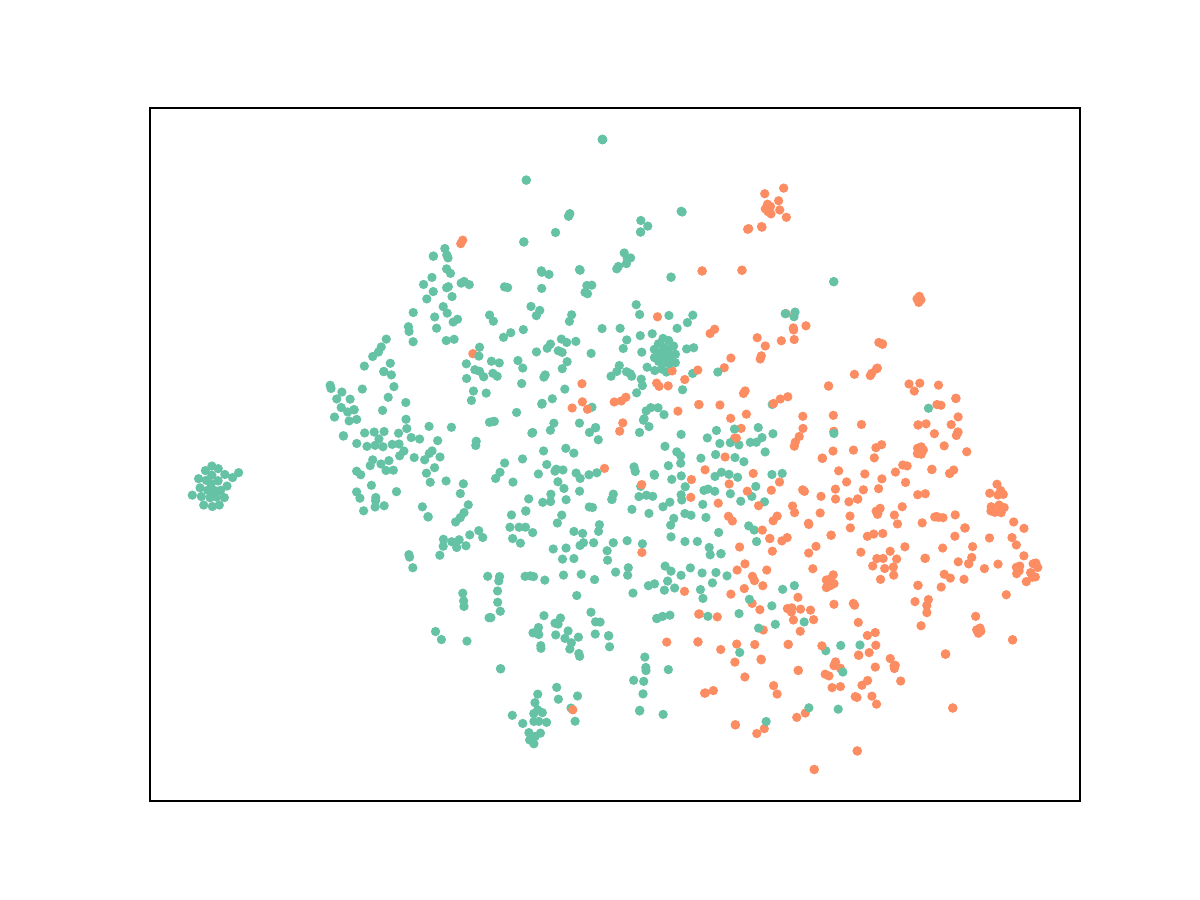}
    \centering
    {\small \mbox{(d) {Spambase TabR}}}
    \end{minipage} 

    \begin{minipage}{0.24\linewidth}
    \includegraphics[width=\textwidth]{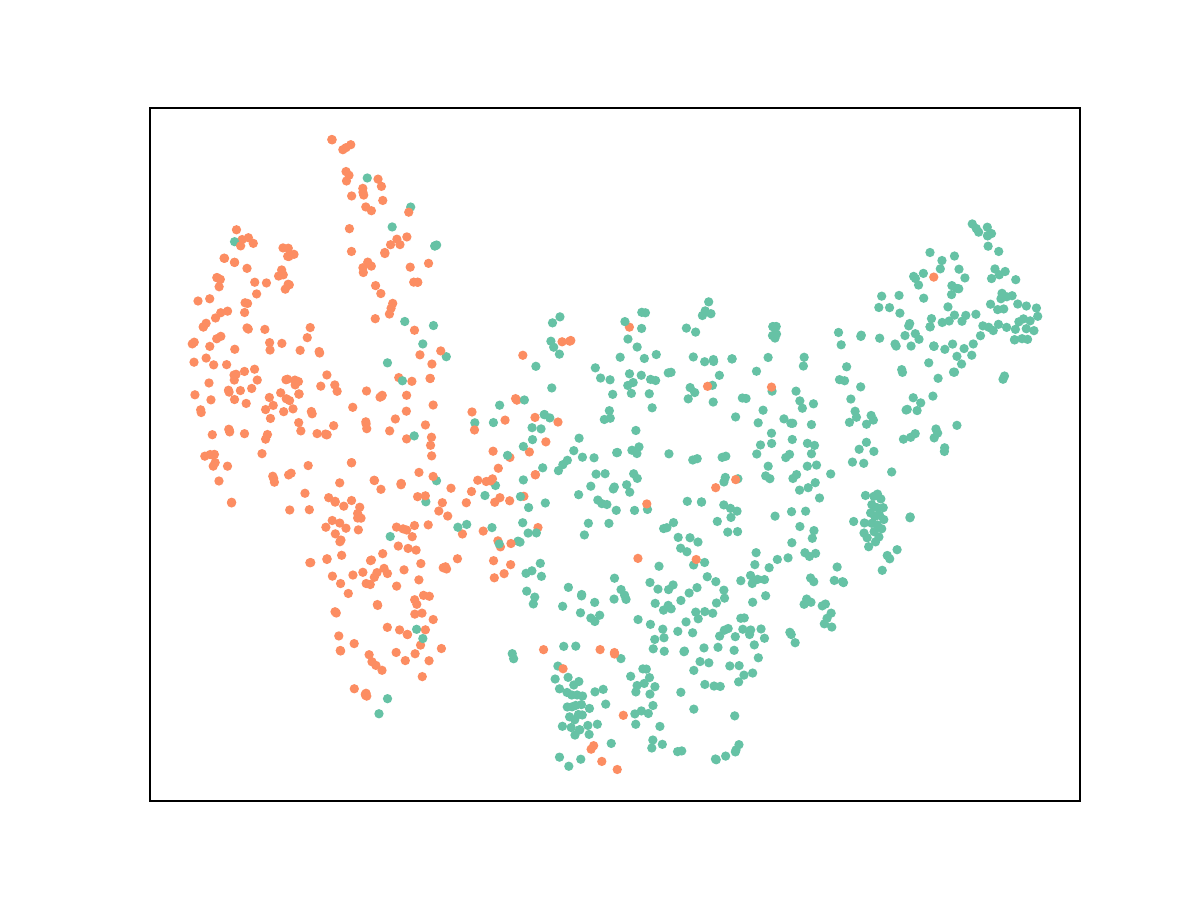}
    \centering
    {\small \mbox{(e) {Spambase \name~1}}}
    \end{minipage}
    \begin{minipage}{0.24\linewidth}
    \includegraphics[width=\textwidth]{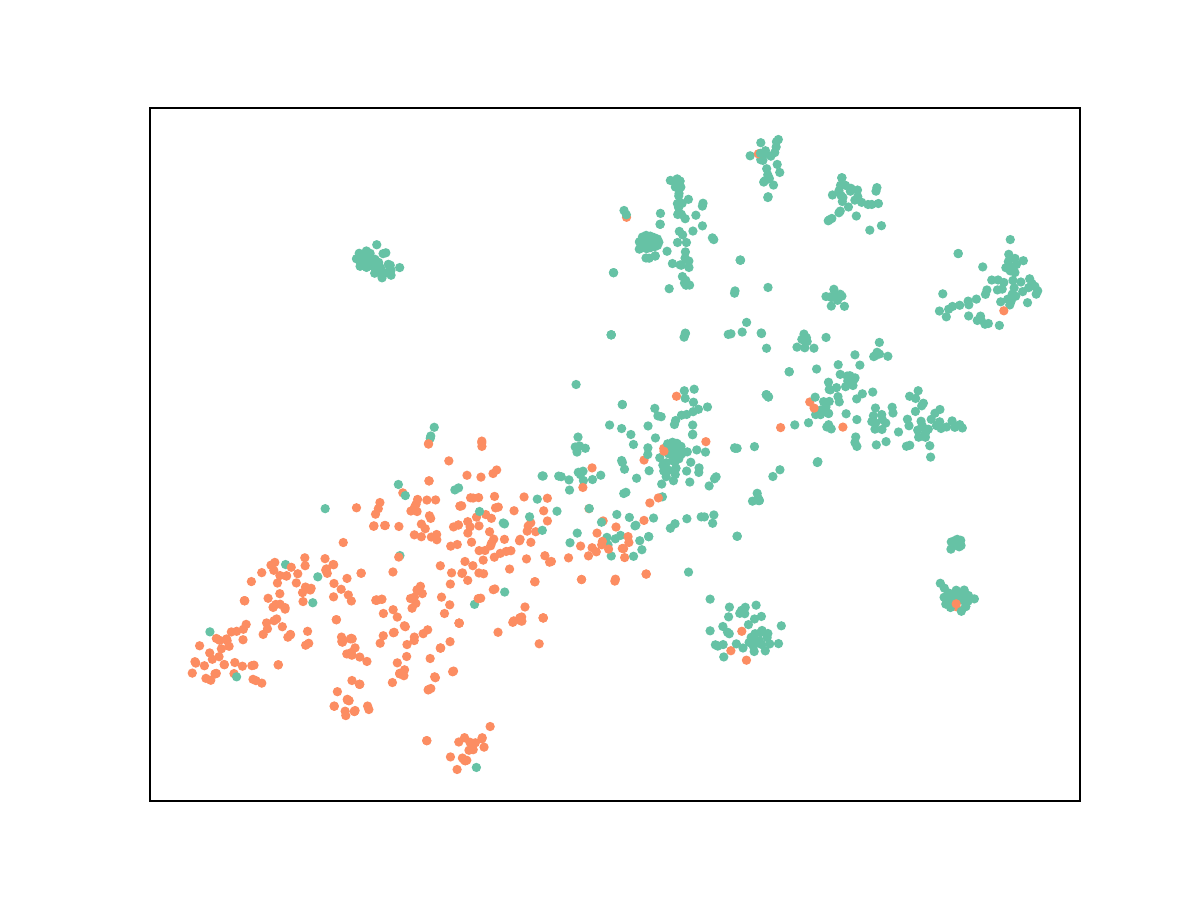}
    \centering
    {\small \mbox{(f) {Spambase \name~2}}}
    \end{minipage}
    \begin{minipage}{0.24\linewidth}
    \includegraphics[width=\textwidth]{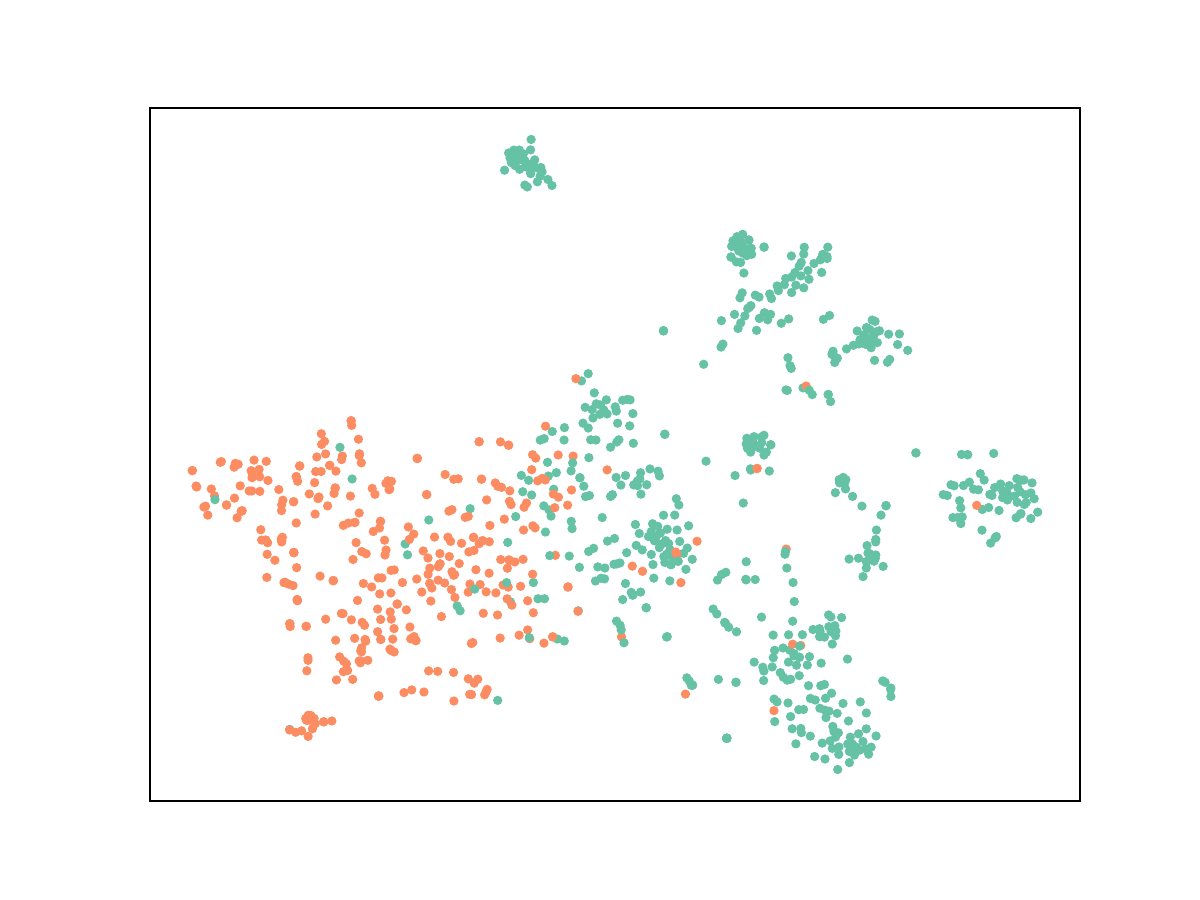}
    \centering
    {\small \mbox{(g) {Spambase \name~3}}}
    \end{minipage}
    \begin{minipage}{0.24\linewidth}
    \includegraphics[width=\textwidth]{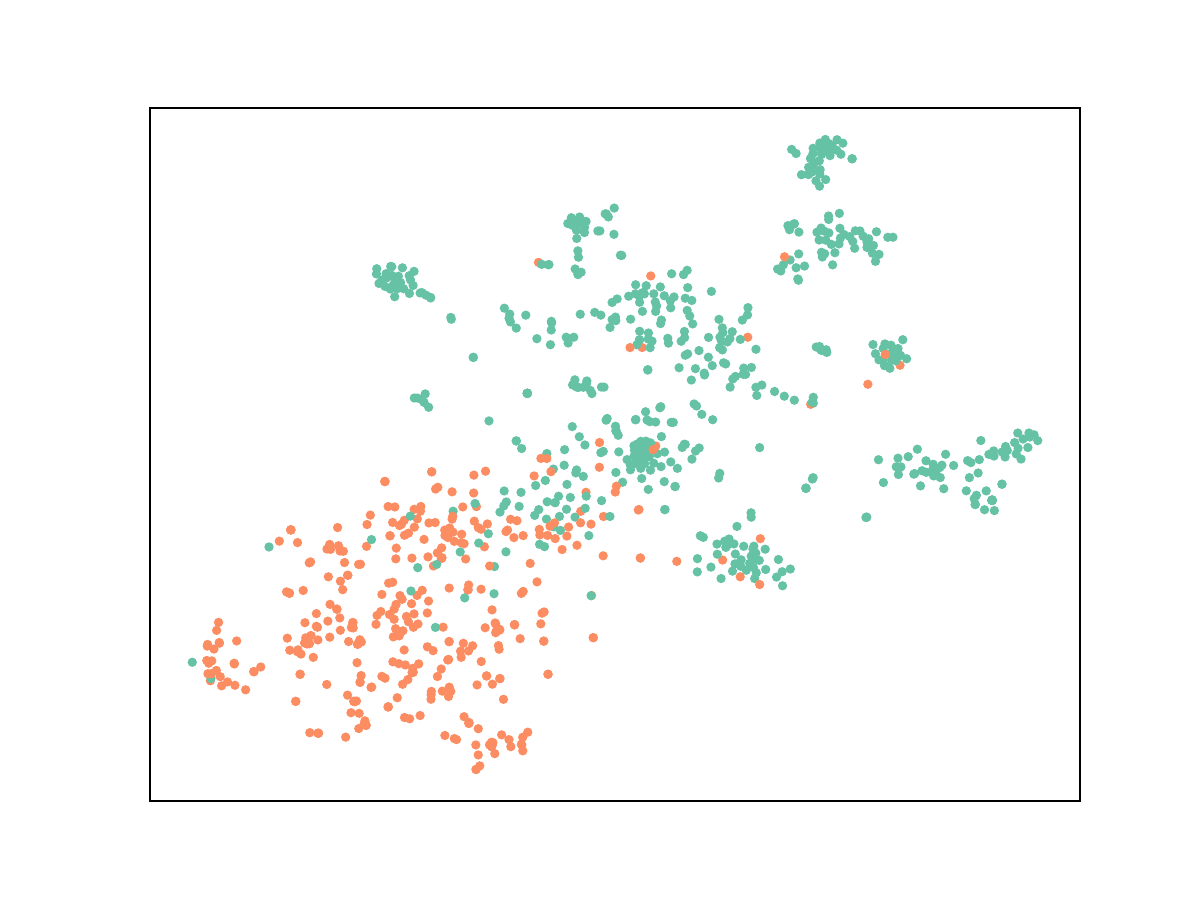}
    \centering
    {\small \mbox{(h) {Spambase \name~4}}}
    \end{minipage}   
    
    \caption{Visualization of the embedding space of different methods.}
    \label{fig:visualize}
\end{figure*}

To gain deeper insights into the properties of~\name, we visualize the learned embeddings \( \boldsymbol{E}_{\Phi}(\boldsymbol{x}) \) for~\name, MLP, ModernNCA, and TabR using T-SNE~\cite{van2008t-SNE}. As shown in~\autoref{fig:visualize}, we present the embeddings of three datasets (\textit{Bank}, \textit{KDD}, and \textit{Spambase}), comparing different methods as well as the representations learned by different encoder paths within~\name.  

All deep tabular methods effectively transform the feature space, making it more conducive for classification compared to raw input features. MLP produces well-separated clusters, grouping same-class samples into distinct, compact regions. In contrast, TabR and ModernNCA form multiple clusters for the same class, positioning similar samples closer to each other while maintaining local class separability. Unlike these methods,~\name~does not enforce strict clustering of same-class samples into single compact groups. This behavior aligns with TabPFN’s pretraining paradigm, where the model does not require fully separable embeddings to make accurate predictions.  

Moreover, we observe that different encoder paths in~\name~map raw features into diverse representation spaces, contributing to variance reduction and improved robustness. The variation among encoder paths enhances the diversity of the learned representations, which is a key factor in~\name's superior generalization. The visualization further confirms that~\name~preserves the flexibility of the feature space while effectively adapting to downstream classification tasks.\looseness=-1

\subsection{Detail Results}

{\tiny{
\tabcolsep 1pt
\begin{longtable}{lccccccccccccccccccccccccccccc}
\caption{Average accuracy of different methods on 186 datasets. Each value represents the mean accuracy over 15 random seeds. The dataset ID corresponds to the numbering in~\citet{YeACloser}. ``XGB'', ``CatB'', ``RF'', ``LightG'', ``TNet'',  ``FTT'',  ``DCN2'',  ``TabT'', ``GNet'',  ``MNCA'', ``M-PLR'', ``RMLP'',``ExcelF'',``PFN'', ``LPFN'', ``kPFN'', ``MPFN'', and ``TTab'' denote ``XGBoost'', ``Catboost'', ``RandomForest'', ``LightGBM'', ``TabNet'', ``FT-Transformer'', ``DCNv2'', ``TabTransformer'', ``GrowNet'', ``ModernNCA'', ``MLP-PLR'', ``RealMLP'', ``ExcelFormer'', ``TabPFN'', ``LocalFPN'', ``knnPFN'', ``MixturePFN'',  and ``TuneTable'', respectively. In each cell, if the experiment fails to perform due to being out of memory, we denote it by ``-''. The original TabPFN cannot handle datasets with more than 10 classes or more than 100 features.}
\label{tab:dataset_additional_result_bin} \\
\toprule
ID & KNN & SVM & XGB & CatB & RF & LightG & MLP & ResNet & NODE & TNet & DaNet & FTT & AutoInt & DCN2 & SNN & TabT & GNet & TabR & MNCA & M-PLR & RMLP & ExcelF & PFN & TabM & \textbf{\name} & LPFN & kPFN & MPFN & TTab \\
\midrule
8 & 94.25 & 94.28 & 94.48 & 94.67 & 94.22 & 94.56 & 94.27 & 94.20 & 94.22 & 94.18 & 94.23 & 94.20 & 94.31 & 94.21 & 94.31 & 94.50 & 93.78 & 94.39 & 94.29 & 94.29 & 94.36 & 94.37 & 94.22 & 94.29 & 94.22 & 94.26 & 94.22 & 94.22 & 94.22 \\
10 & 75.96 & 73.66 & 74.55 & 74.45 & 74.22 & 73.99 & 73.44 & 73.55 & 72.97 & 73.41 & 73.11 & 73.39 & 73.32 & 73.43 & 73.56 & 63.16 & 73.51 & 75.69 & 75.64 & 73.37 & 73.70 & 73.27 & 73.22 & 73.94 & 73.58 & 73.51 & 73.56 & 73.39 & 73.39 \\
11 & 98.37 & 97.81 & 98.76 & 98.78 & 98.52 & 98.73 & 98.46 & 98.46 & 98.56 & 98.34 & 98.40 & 98.60 & 98.48 & 98.41 & 98.48 & 95.94 & 98.39 & 98.68 & 98.59 & 98.52 & 98.62 & 98.56 & 98.27 & 98.62 & 98.80 & 98.47 & 98.44 & 98.31 & 98.31 \\
12 & 55.74 & 52.39 & 58.60 & 58.93 & 57.30 & 58.67 & 57.90 & 57.91 & 58.81 & 57.72 & 57.95 & 58.50 & 58.29 & 58.38 & 57.80 & 52.73 & 52.81 & 58.23 & 58.56 & 58.45 & 58.28 & 58.01 & 55.58 & 58.70 & 58.56 & 57.77 & 57.37 & 56.36 & 55.54 \\
17 & 78.31 & 72.80 & 81.21 & 81.33 & 80.14 & 81.55 & 81.53 & 81.18 & 81.17 & 79.74 & 81.27 & 81.31 & 81.00 & 81.53 & 81.50 & 81.39 & 81.12 & 81.14 & 81.08 & 81.42 & 81.36 & 80.80 & 75.99 & 81.45 & 81.37 & 80.21 & 79.96 & 78.20 & 78.20 \\
18 & 83.85 & 81.65 & 86.97 & 87.59 & 86.92 & 86.31 & 86.56 & 86.56 & 86.54 & 86.31 & 86.52 & 87.09 & 86.74 & 86.74 & 86.82 & 83.13 & 86.24 & 87.43 & 85.42 & 87.43 & 87.35 & 87.06 & 85.79 & 86.75 & 87.94 & 86.78 & 86.82 & 85.27 & 85.27 \\
19 & 66.79 & 70.82 & 68.58 & 67.61 & 68.91 & 68.53 & 69.75 & 70.35 & 65.42 & 65.77 & 70.75 & 70.15 & 70.02 & 70.12 & 69.58 & 69.28 & 66.94 & 70.50 & 68.11 & 69.30 & 70.50 & 69.70 & 70.42 & 68.03 & 70.27 & 70.17 & 70.35 & 70.82 & 70.35 \\
25 & 85.22 & 84.13 & 90.00 & 90.82 & 87.65 & 90.70 & 88.47 & 88.51 & 87.30 & 87.87 & 88.30 & 89.23 & 88.47 & 88.58 & 88.43 & 85.01 & 87.43 & 91.35 & 90.53 & 88.78 & 89.36 & 89.15 & 87.31 & 89.53 & 89.90 & 89.47 & 88.52 & 87.39 & 87.05 \\
26 & 67.37 & 68.78 & 73.43 & 73.40 & 73.23 & 73.37 & 73.35 & 73.07 & 73.20 & 71.47 & 73.15 & 73.22 & 73.26 & 73.21 & 73.28 & 67.99 & 66.77 & 73.08 & 73.20 & 73.25 & 73.29 & 73.17 & 72.54 & 73.15 & 73.24 & 73.27 & 72.62 & 72.62 & 72.62 \\
27 & 83.25 & 79.21 & 83.25 & 83.26 & 83.25 & 83.32 & 83.34 & 83.28 & 83.31 & 83.19 & 83.21 & 83.33 & 83.30 & 83.27 & 83.32 & 83.20 & 83.26 & 83.22 & 83.31 & 83.33 & 83.33 & 83.36 & 83.15 & 83.31 & 83.32 & 83.20 & 83.21 & 83.17 & 83.15 \\
28 & 87.92 & 89.17 & 91.35 & 90.57 & 87.97 & 91.57 & 92.67 & 94.07 & 93.28 & 83.24 & 93.19 & 91.51 & 93.72 & 93.71 & 92.47 & 92.94 & 91.60 & 92.72 & 94.94 & 92.08 & 93.33 & 90.03 & 94.44 & 90.97 & 93.60 & 94.99 & 94.28 & 95.19 & 93.60 \\
29 & 89.58 & 87.29 & 91.60 & 91.33 & 88.51 & 91.33 & 92.88 & 93.72 & 93.06 & 85.99 & 93.82 & 90.82 & 93.43 & 93.50 & 93.15 & 92.82 & 91.39 & 94.36 & 94.35 & 91.93 & 94.29 & 90.31 & 94.64 & 90.83 & 94.15 & 96.21 & 94.19 & 95.76 & 94.15 \\
30 & 86.46 & 87.29 & 90.11 & 89.86 & 88.78 & 89.99 & 93.04 & 93.22 & 92.61 & 84.94 & 92.15 & 90.06 & 91.53 & 92.57 & 92.22 & 91.74 & 90.68 & 93.93 & 93.42 & 92.17 & 92.62 & 89.21 & 93.51 & 90.67 & 92.64 & 94.35 & 93.28 & 95.07 & 92.64 \\
31 & 86.46 & 85.62 & 90.14 & 90.21 & 88.92 & 90.39 & 91.85 & 92.15 & 91.81 & 85.58 & 92.15 & 89.00 & 91.31 & 92.17 & 91.94 & 91.29 & 90.11 & 92.96 & 93.22 & 89.35 & 92.78 & 88.79 & 92.86 & 89.24 & 92.86 & 94.71 & 92.86 & 93.92 & 92.86 \\
32 & 86.04 & 87.50 & 88.19 & 87.51 & 87.15 & 88.10 & 91.29 & 92.64 & 91.54 & 82.06 & 91.72 & 88.96 & 91.25 & 92.85 & 91.60 & 91.61 & 88.99 & 89.49 & 93.81 & 89.47 & 92.22 & 86.68 & 92.75 & 86.68 & 92.64 & 95.28 & 93.19 & 93.94 & 92.64 \\
34 & 77.86 & 63.11 & 78.01 & 79.72 & 58.54 & 77.12 & 77.56 & 76.98 & 71.94 & 70.34 & 77.08 & 77.70 & 77.12 & 77.71 & 76.98 & 76.77 & 64.23 & 82.01 & 79.86 & 78.91 & 78.50 & 78.41 & 66.76 & 79.06 & 75.47 & 53.17 & 53.17 & 53.17 & 68.79 \\
35 & 86.61 & 88.90 & 88.41 & 89.05 & 87.71 & 89.29 & 88.74 & 89.06 & 85.80 & 85.30 & 88.07 & 89.27 & 89.08 & 89.17 & 88.26 & 88.24 & 86.92 & 89.51 & 89.14 & 88.75 & 89.40 & 89.14 & 89.32 & 89.43 & 89.51 & 89.24 & 89.32 & 89.36 & 88.56 \\
37 & 62.34 & 72.27 & 72.67 & 74.14 & 73.68 & 72.01 & 74.81 & 74.23 & 57.72 & 60.87 & 75.61 & 73.19 & 71.60 & 76.71 & 73.30 & 71.77 & 66.44 & 75.04 & 74.69 & 74.46 & 73.74 & 73.80 & 72.61 & 75.01 & 75.82 & 73.22 & 73.62 & 73.22 & 73.69 \\
39 & 62.55 & 63.64 & 67.45 & 66.72 & 66.60 & 67.54 & 66.32 & 66.46 & 66.34 & 66.28 & 65.33 & 67.28 & 66.50 & 66.27 & 65.84 & 62.92 & 64.76 & 67.00 & 67.97 & 67.41 & 66.80 & 66.94 & 65.02 & 66.88 & 67.28 & 66.43 & 65.96 & 65.80 & 64.96 \\
40 & 79.81 & 72.50 & 84.77 & 85.00 & 85.13 & 84.93 & 85.03 & 84.81 & 84.38 & 81.57 & 85.08 & 84.77 & 84.86 & 84.97 & 84.65 & 77.03 & 84.60 & 83.82 & 84.59 & 85.07 & 84.35 & 85.06 & 85.01 & 84.53 & 84.68 & 85.36 & 85.08 & 84.75 & 84.29 \\
41 & 71.90 & 72.71 & 73.96 & 74.27 & 73.65 & 74.72 & 73.25 & 73.40 & 74.51 & 73.10 & 73.43 & 74.10 & 73.93 & 73.74 & 73.42 & 73.53 & 72.75 & 73.63 & 74.28 & 73.99 & 74.21 & 74.11 & 73.82 & 73.85 & 74.04 & 74.06 & 74.22 & 73.64 & 73.48 \\
42 & 62.13 & 73.91 & 67.14 & 68.16 & 67.50 & 68.12 & 73.84 & 74.80 & 55.77 & 67.87 & 72.77 & 73.46 & 70.90 & 73.70 & 71.93 & 73.86 & 70.10 & 74.28 & 70.57 & 69.45 & 74.22 & 68.07 & 75.00 & 64.80 & 75.26 & 75.50 & 75.26 & 75.33 & 75.26 \\
43 & 57.65 & 70.34 & 63.81 & 66.30 & 58.49 & 66.42 & 71.21 & 71.21 & 59.00 & 71.06 & 71.12 & 69.01 & 70.89 & 71.08 & 71.14 & 70.70 & 70.10 & 71.16 & 71.15 & 70.08 & 71.42 & 70.52 & 69.89 & 69.78 & 71.13 & 71.36 & 70.42 & 70.42 & 70.71 \\
44 & 59.40 & 75.20 & 66.27 & 63.98 & 65.54 & 66.18 & 69.01 & 74.64 & 65.23 & 51.35 & 53.70 & 51.79 & 65.67 & 74.80 & 72.62 & 74.24 & 67.54 & 75.20 & 75.44 & 65.00 & 75.19 & 52.97 & 75.59 & 62.05 & 75.55 & 75.97 & 75.46 & 75.45 & 75.46 \\
45 & 59.13 & 76.64 & 64.87 & 65.58 & 64.16 & 64.60 & 73.04 & 76.02 & 55.06 & 56.71 & 67.16 & 67.03 & 68.83 & 74.50 & 74.02 & 73.13 & 69.41 & 76.55 & 75.26 & 67.74 & 77.09 & 64.21 & 75.89 & 63.05 & 76.17 & 76.06 & 76.24 & 76.13 & 76.87 \\
46 & 54.80 & 71.42 & 64.07 & 64.95 & 58.18 & 64.40 & 71.73 & 71.78 & 54.36 & 58.69 & 70.58 & 69.13 & 70.06 & 71.32 & 71.38 & 71.32 & 65.47 & 72.02 & 72.09 & 67.77 & 72.08 & 69.28 & 67.25 & 70.61 & 71.36 & 71.88 & 66.74 & 68.15 & 71.22 \\
47 & 52.54 & 70.41 & 60.23 & 61.58 & 56.53 & 62.28 & 69.85 & 70.92 & 52.25 & 64.13 & 65.62 & 61.84 & 68.84 & 69.09 & 68.99 & 66.99 & 63.36 & 70.80 & 71.09 & 64.53 & 70.56 & 61.46 & 58.13 & 65.59 & 66.28 & 70.73 & 56.87 & 65.31 & 69.45 \\
48 & 53.13 & 69.13 & 60.80 & 61.65 & 56.57 & 62.25 & 68.31 & 69.32 & 53.47 & 67.62 & 64.94 & 67.17 & 65.59 & 69.76 & 68.82 & 69.65 & 62.14 & 71.09 & 70.98 & 61.62 & 70.84 & 56.90 & 54.51 & 67.46 & 59.37 & 70.61 & 55.33 & 63.62 & 52.96 \\
49 & 52.04 & 71.39 & 60.25 & 58.89 & 57.13 & 59.13 & 72.04 & 70.05 & 70.63 & 56.71 & 70.14 & 66.65 & 67.67 & 71.52 & 59.26 & 71.50 & 65.45 & 71.24 & 70.41 & 61.36 & 73.32 & 62.65 & 73.08 & 63.65 & 72.84 & 72.74 & 72.84 & 72.74 & 72.84 \\
50 & 54.61 & 70.99 & 61.21 & 62.59 & 57.18 & 62.72 & 70.61 & 70.96 & 53.38 & 62.43 & 68.71 & 68.80 & 69.71 & 70.45 & 69.90 & 70.34 & 65.86 & 64.32 & 71.18 & 63.79 & 71.31 & 66.44 & 62.90 & 70.26 & 69.15 & 71.10 & 61.94 & 66.46 & 70.39 \\
52 & 80.37 & 77.62 & 85.71 & 85.97 & 80.72 & 85.76 & 86.74 & 87.11 & 81.42 & 83.01 & 85.79 & 87.11 & 86.48 & 86.39 & 86.40 & 86.36 & 86.20 & 87.34 & 87.19 & 86.60 & 87.20 & 85.30 & 83.86 & 87.15 & 87.08 & 87.05 & 84.81 & 86.37 & 83.39 \\
53 & 74.00 & 76.67 & 75.38 & 78.11 & 77.93 & 79.64 & 76.29 & 77.56 & 78.62 & 75.11 & 76.18 & 77.02 & 76.24 & 76.93 & 77.04 & 74.58 & 75.38 & 78.04 & 78.40 & 77.24 & 77.69 & 74.04 & 79.93 & 78.98 & 78.56 & 79.00 & 78.91 & 80.49 & 78.24 \\
55 & 55.94 & 48.77 & 60.35 & 68.08 & 58.96 & 61.02 & 59.40 & 60.23 & 56.96 & 49.38 & 55.85 & 49.31 & 54.12 & 58.90 & 63.65 & 64.33 & 57.75 & 63.77 & 59.90 & 55.40 & 65.31 & 50.98 & 58.94 & 48.85 & 65.69 & 63.50 & 56.69 & 61.12 & 56.69 \\
56 & 64.06 & 50.19 & 67.12 & 67.96 & 65.44 & 68.23 & 65.12 & 65.71 & 53.29 & 49.54 & 63.88 & 53.25 & 59.42 & 66.38 & 66.79 & 65.92 & 64.83 & 67.79 & 65.71 & 63.19 & 67.33 & 52.06 & 65.56 & 58.00 & 68.35 & 68.08 & 67.00 & 67.00 & 67.00 \\
57 & 57.26 & 58.88 & 59.87 & 59.26 & 59.59 & 58.82 & 60.09 & 59.61 & 59.17 & 58.74 & 59.00 & 59.86 & 59.51 & 58.95 & 59.57 & 58.09 & 58.84 & 59.14 & 59.25 & 59.48 & 59.81 & 59.41 & 59.00 & 59.77 & 59.85 & 60.55 & 59.02 & 59.52 & 58.90 \\
58 & 65.67 & 45.58 & 68.94 & 69.28 & 62.14 & 68.33 & 63.42 & 61.76 & 52.30 & 49.26 & 61.34 & 65.65 & 63.70 & 62.50 & 63.83 & 57.67 & 54.99 & 73.92 & 69.51 & 66.79 & 69.43 & 68.44 & 55.93 & 69.80 & 66.06 & 69.99 & 58.24 & 66.95 & 52.83 \\
60 & 79.10 & 78.05 & 79.37 & 79.51 & 77.96 & 79.34 & 79.53 & 79.29 & 79.10 & 78.32 & 79.38 & 79.80 & 79.60 & 79.72 & 79.22 & 79.68 & 77.32 & 79.84 & 79.58 & 79.80 & 79.56 & 79.40 & 77.35 & 79.82 & 79.74 & 79.04 & 77.42 & 77.42 & 77.42 \\
61 & 93.70 & 85.77 & 90.78 & 93.47 & 92.18 & 91.29 & 89.10 & 88.32 & 91.62 & 83.78 & 88.01 & 88.94 & 87.11 & 88.21 & 88.15 & 87.65 & 90.50 & 87.98 & 91.34 & 89.08 & 91.15 & 85.46 & 90.78 & 87.42 & 92.30 & 91.06 & 90.98 & 91.06 & 90.78 \\
63 & 84.16 & 80.64 & 89.71 & 89.21 & 87.30 & 89.45 & 87.37 & 87.21 & 84.43 & 86.20 & 86.37 & 88.50 & 87.19 & 87.46 & 87.18 & 85.30 & 78.41 & 89.47 & 89.70 & 88.53 & 88.88 & 88.25 & 83.37 & 89.10 & 87.73 & 88.52 & 87.39 & 88.19 & 83.63 \\
64 & 89.50 & 89.87 & 92.32 & 92.07 & 89.05 & 84.27 & 93.01 & 94.63 & 72.64 & 93.38 & 91.54 & 93.35 & 92.83 & 93.43 & 93.97 & 91.21 & 36.89 & 96.50 & 96.85 & 93.29 & 95.45 & 93.39 & - & 94.90 & 95.45 & 86.58 & 86.58 & 86.58 & 86.58 \\
65 & 75.75 & 75.75 & 75.75 & 75.75 & 75.75 & 75.75 & 75.73 & 75.74 & 75.76 & 75.70 & 75.74 & 75.70 & 75.74 & 75.66 & 75.69 & 75.73 & 75.71 & 75.73 & 75.68 & 75.71 & 75.71 & 75.70 & 75.75 & 75.72 & 75.76 & 75.66 & 75.75 & 75.66 & 75.80 \\
66 & 92.59 & 92.55 & 93.80 & 94.72 & 94.03 & 94.18 & 94.41 & 94.28 & 94.59 & 93.38 & 94.45 & 94.26 & 94.34 & 94.52 & 94.31 & 94.02 & 94.08 & 94.45 & 94.18 & 94.27 & 94.29 & 94.16 & 93.70 & 94.35 & 94.24 & 93.47 & 94.00 & 93.62 & 93.61 \\
67 & 88.70 & 88.08 & 88.27 & 88.41 & 88.41 & 88.46 & 88.75 & 88.31 & 88.02 & 87.94 & 88.25 & 88.60 & 88.39 & 87.85 & 88.37 & 88.17 & 88.25 & 87.67 & 88.62 & 88.43 & 88.58 & 88.17 & 88.41 & 87.85 & 88.35 & 88.25 & 88.37 & 88.41 & 88.33 \\
68 & 98.60 & 92.82 & 98.03 & 98.30 & 96.05 & 98.21 & 98.98 & 99.35 & 94.81 & 97.25 & 98.76 & 98.69 & 98.79 & 98.79 & 98.89 & 96.58 & 98.10 & 99.59 & 99.56 & 98.70 & 99.40 & 98.31 & 97.27 & 98.99 & 98.56 & 99.54 & 98.78 & 99.48 & 96.89 \\
70 & 71.70 & 74.85 & 80.77 & 80.79 & 80.01 & 80.21 & 77.06 & 77.17 & 80.40 & 76.76 & 76.35 & 80.28 & 76.50 & 77.10 & 76.72 & 73.55 & 74.63 & 79.72 & 79.58 & 78.61 & 79.25 & 80.25 & 77.58 & 80.33 & 81.34 & 76.63 & 76.78 & 76.63 & 76.63 \\
71 & 69.79 & 73.39 & 81.38 & 81.03 & 80.48 & 80.34 & 75.52 & 75.02 & 77.41 & 76.84 & 75.00 & 80.38 & 76.51 & 75.65 & 76.40 & 74.75 & 71.68 & 79.41 & 79.63 & 79.81 & 79.43 & 80.70 & 76.82 & 80.27 & 81.14 & 76.74 & 75.45 & 76.74 & 75.45 \\
77 & 89.51 & 99.08 & 99.12 & 98.82 & 96.60 & 99.15 & 97.79 & 98.60 & 99.43 & 98.44 & 98.40 & 98.87 & 98.36 & 99.01 & 98.56 & 95.31 & 95.67 & 98.99 & 98.99 & 99.11 & 99.45 & 99.00 & 98.53 & 99.65 & 99.76 & 98.76 & 98.76 & 98.76 & 98.76 \\
78 & 84.85 & 87.88 & 89.37 & 88.71 & 87.35 & 91.56 & 88.06 & 88.71 & 84.91 & 84.06 & 86.87 & 89.68 & 87.78 & 89.09 & 88.85 & 89.09 & 86.97 & 88.48 & 87.82 & 88.81 & 87.96 & 88.95 & - & 90.14 & 89.70 & 89.15 & 89.15 & 89.15 & 89.15 \\
80 & 84.65 & 79.34 & 87.63 & 87.87 & 86.90 & 87.88 & 87.24 & 87.61 & 87.01 & 86.98 & 86.91 & 87.59 & 87.03 & 87.18 & 86.88 & 83.09 & 86.70 & 88.13 & 85.96 & 87.05 & 87.86 & 87.58 & 86.69 & 87.83 & 88.24 & 87.28 & 87.21 & 87.19 & 86.84 \\
81 & 87.95 & 87.96 & 88.17 & 88.05 & 87.22 & 88.05 & 87.71 & 87.65 & 86.15 & 84.94 & 88.12 & 86.88 & 88.26 & 86.43 & 88.85 & 87.72 & 86.95 & 87.95 & 90.19 & 88.18 & 88.04 & 88.04 & 88.88 & 88.50 & 89.08 & 88.60 & 89.55 & 88.85 & 88.60 \\
84 & 67.25 & 82.92 & 90.80 & 93.57 & 87.02 & 91.08 & 91.57 & 92.65 & 89.53 & 88.07 & 91.75 & 93.67 & 93.73 & 95.50 & 95.42 & 57.23 & 84.50 & 96.23 & 98.12 & 96.97 & 96.27 & 95.27 & 96.77 & 96.88 & 97.17 & 97.07 & 97.17 & 97.07 & 97.17 \\
88 & 82.89 & 83.19 & 84.23 & 84.27 & 85.07 & 83.64 & 83.03 & 82.56 & 84.25 & 83.77 & 83.30 & 83.63 & 83.83 & 83.00 & 83.04 & 83.08 & 83.20 & 83.44 & 83.26 & 83.49 & 83.85 & 83.39 & 83.38 & 83.41 & 83.70 & 83.38 & 83.38 & 83.51 & 83.38 \\
92 & 72.39 & 69.23 & 71.17 & 72.91 & 72.04 & 70.90 & 72.11 & 73.38 & 72.19 & 66.47 & 68.86 & 72.14 & 72.09 & 73.06 & 71.59 & 67.86 & 70.10 & 70.95 & 72.89 & 69.53 & 71.59 & 70.75 & 75.52 & 72.59 & 75.45 & 75.35 & 76.22 & 75.35 & 75.45 \\
93 & 96.61 & 93.98 & 97.22 & 96.79 & 96.39 & 97.19 & 97.27 & 97.40 & 94.86 & 94.95 & 97.33 & 96.76 & 96.94 & 96.96 & 97.09 & 97.30 & 96.20 & 97.74 & 97.67 & 97.33 & 97.25 & 96.98 & 94.95 & 97.47 & 96.47 & 97.58 & 97.19 & 96.72 & 95.28 \\
96 & 71.43 & 74.68 & 76.45 & 77.75 & 77.27 & 78.14 & 76.06 & 74.94 & 77.06 & 73.03 & 72.51 & 75.58 & 76.62 & 73.64 & 76.49 & 68.61 & 76.02 & 75.97 & 75.24 & 75.76 & 74.03 & 76.58 & 75.58 & 77.01 & 77.40 & 75.45 & 76.23 & 76.06 & 74.85 \\
98 & 86.40 & 85.87 & 87.16 & 87.25 & 87.40 & 87.47 & 86.91 & 86.77 & 86.07 & 86.13 & 85.08 & 86.72 & 87.07 & 87.17 & 86.49 & 86.39 & 86.33 & 86.87 & 86.99 & 87.01 & 86.96 & 86.85 & 87.45 & 87.25 & 87.27 & 87.01 & 87.43 & 87.01 & 87.24 \\
99 & 89.10 & 84.42 & 85.06 & 87.30 & 87.42 & 84.08 & 86.67 & 86.67 & 69.76 & 79.87 & 86.86 & 86.95 & 85.12 & 87.08 & 87.71 & 86.26 & 85.69 & 86.03 & 84.55 & 87.27 & 85.88 & 86.89 & 87.52 & 84.93 & 87.84 & 87.27 & 87.58 & 87.39 & 87.27 \\
100 & 83.31 & 82.21 & 85.28 & 86.89 & 76.55 & 85.18 & 85.85 & 85.73 & 84.74 & 83.74 & 84.33 & 85.66 & 85.50 & 85.91 & 85.40 & 85.02 & 81.78 & 86.09 & - & 86.27 & 86.96 & 85.05 & 83.12 & 86.78 & 86.35 & 83.12 & 83.12 & 83.12 & 83.12 \\
102 & 72.72 & 93.55 & 97.67 & 97.52 & 81.77 & 97.60 & 94.58 & 95.27 & 97.28 & 97.25 & 97.10 & 97.03 & 95.02 & 96.83 & 96.62 & 91.96 & 76.52 & 97.24 & 97.38 & 97.22 & 97.21 & 97.40 & 96.53 & 97.21 & 97.16 & 91.29 & 91.29 & 91.29 & 91.29 \\
103 & 62.18 & 64.22 & 66.83 & 67.00 & 66.30 & 66.75 & 65.93 & 67.43 & 66.68 & 66.38 & 65.39 & 68.48 & 66.60 & 67.64 & 67.10 & 64.21 & 64.80 & 67.98 & 67.08 & 67.17 & 67.50 & 67.98 & 65.68 & 67.74 & 68.49 & 67.20 & 66.46 & 66.76 & 65.08 \\
106 & 85.46 & 86.10 & 95.50 & 95.17 & 94.22 & 95.80 & 87.94 & 89.09 & 91.29 & 89.32 & 87.36 & 95.19 & 92.58 & 91.96 & 91.29 & 85.74 & 90.85 & 94.09 & 93.73 & 93.58 & 94.16 & 93.20 & 93.30 & 94.78 & 95.32 & 92.62 & 91.96 & 93.33 & 91.81 \\
108 & 98.95 & 98.66 & 98.95 & 98.95 & 98.95 & 98.95 & 98.92 & 98.82 & 98.95 & 98.81 & 98.93 & 98.93 & 98.95 & 98.92 & 98.92 & 98.82 & 98.91 & 98.88 & 98.79 & 98.90 & 98.89 & 98.93 & 98.95 & 98.94 & 98.95 & 98.86 & 98.95 & 98.94 & 98.95 \\
109 & 63.87 & 60.98 & 61.44 & 62.50 & 62.93 & 62.86 & 63.01 & 62.80 & 65.05 & 60.66 & 62.58 & 63.10 & 62.13 & 62.58 & 61.15 & 62.35 & 61.32 & 62.63 & 62.80 & 62.33 & 61.64 & 62.23 & 61.54 & 61.68 & 62.69 & 63.20 & 62.26 & 62.31 & 62.81 \\
110 & 67.23 & 67.96 & 65.88 & 65.10 & 65.27 & 65.80 & 68.74 & 67.93 & 70.56 & 58.15 & 68.60 & 65.35 & 66.25 & 68.29 & 68.49 & 67.54 & 69.05 & 65.97 & 70.03 & 66.30 & 69.80 & 60.06 & 74.45 & 62.16 & 74.51 & 74.41 & 74.51 & 74.41 & 74.51 \\
111 & 96.58 & 96.65 & 98.83 & 98.79 & 98.36 & 98.69 & 98.85 & 98.42 & 94.61 & 97.11 & 98.46 & 98.49 & 98.94 & 98.51 & 98.35 & 97.61 & 98.04 & 98.72 & 99.07 & 98.63 & 98.80 & 98.57 & 99.14 & 98.22 & 99.12 & 99.02 & 99.08 & 99.20 & 99.08 \\
114 & 59.57 & 61.87 & 61.52 & 61.84 & 61.90 & 60.97 & 63.15 & 63.57 & 59.62 & 61.18 & 63.29 & 63.01 & 62.73 & 62.62 & 62.49 & 60.68 & 62.11 & 62.71 & 63.04 & 62.54 & 63.25 & 62.93 & 61.98 & 62.95 & 63.51 & 63.17 & 63.17 & 63.17 & 63.17 \\
116 & 82.89 & 83.61 & 85.53 & 85.66 & 84.76 & 85.12 & 84.34 & 84.45 & 82.75 & 83.02 & 84.07 & 84.80 & 83.39 & 84.05 & 83.08 & 84.46 & 82.86 & 84.83 & 84.97 & 84.73 & 84.35 & 84.72 & 84.42 & 85.73 & 85.61 & 84.67 & 84.34 & 84.34 & 84.34 \\
117 & 79.74 & 83.07 & 83.62 & 84.51 & 84.54 & 83.82 & 83.48 & 82.92 & 81.34 & 80.25 & 82.63 & 83.69 & 83.64 & 83.25 & 83.39 & 81.80 & 82.38 & 83.57 & 83.21 & 83.96 & 83.37 & 83.87 & 84.05 & 84.05 & 84.25 & 83.85 & 83.85 & 83.85 & 83.85 \\
118 & 80.94 & 84.34 & 85.03 & 85.00 & 83.86 & 84.49 & 82.98 & 82.56 & 81.55 & 82.67 & 82.21 & 84.83 & 84.45 & 84.76 & 83.64 & 81.48 & 81.38 & 84.62 & 83.88 & 83.72 & 84.15 & 83.69 & 84.00 & 84.05 & 84.57 & 84.00 & 84.00 & 84.00 & 84.00 \\
119 & 84.07 & 82.72 & 86.93 & 87.31 & 86.23 & 87.41 & 85.19 & 85.16 & 86.06 & 85.25 & 85.35 & 85.88 & 85.61 & 85.69 & 85.72 & 84.65 & 85.68 & 86.90 & 87.28 & 86.39 & 86.10 & 86.02 & 84.44 & 87.19 & 86.46 & 85.51 & 84.62 & 84.62 & 84.62 \\
121 & 63.50 & 60.50 & 61.67 & 62.88 & 63.17 & 61.27 & 57.20 & 58.05 & 58.65 & 55.22 & 59.02 & 60.32 & 60.05 & 60.97 & 61.43 & 63.63 & 59.88 & 62.42 & 60.08 & 62.17 & 60.93 & 59.13 & 58.40 & 62.37 & 60.80 & 60.20 & 60.20 & 60.20 & 60.20 \\
122 & 96.16 & 95.81 & 98.03 & 97.78 & 96.83 & 97.60 & 96.49 & 96.50 & 95.87 & 96.58 & 96.44 & 97.09 & 97.26 & 96.30 & 96.94 & 96.20 & 96.10 & 97.04 & 97.13 & 97.16 & 97.13 & 97.46 & 97.29 & 97.28 & 97.65 & 97.26 & 97.20 & 97.38 & 97.20 \\
123 & 97.09 & 96.77 & 98.22 & 97.99 & 97.73 & 98.53 & 96.67 & 97.02 & 96.69 & 96.59 & 97.01 & 97.73 & 97.43 & 97.03 & 97.62 & 96.62 & 96.27 & 97.93 & 97.79 & 97.81 & 97.67 & 97.60 & 96.99 & 98.56 & 98.13 & 97.48 & 97.30 & 97.01 & 97.01 \\
124 & 98.82 & 99.41 & 98.78 & 98.58 & 98.66 & 99.09 & 98.70 & 97.71 & 97.99 & 90.53 & 98.62 & 97.24 & 98.42 & 98.93 & 98.74 & 98.38 & 98.19 & 98.86 & 99.41 & 98.78 & 99.33 & 97.95 & 99.37 & 99.13 & 99.37 & 99.37 & 99.41 & 99.41 & 99.41 \\
128 & 83.32 & 32.13 & 87.72 & 88.96 & 76.57 & 88.51 & 78.10 & 77.14 & 60.77 & 63.88 & 70.36 & 80.80 & 76.12 & 77.91 & 77.27 & 61.29 & 71.83 & 90.38 & 91.67 & 91.32 & 83.27 & 77.76 & 63.04 & 92.65 & 65.56 & 79.42 & 66.13 & 73.84 & 64.39 \\
130 & 54.40 & 60.53 & 65.72 & 66.49 & 61.07 & 64.33 & 58.39 & 58.47 & 51.04 & 45.23 & 57.85 & 62.77 & 59.17 & 57.65 & 52.81 & 58.01 & 46.51 & 57.91 & 50.77 & 55.87 & 62.71 & 57.64 & 50.15 & 61.11 & 65.39 & 54.31 & 54.31 & 54.31 & 54.31 \\
131 & 35.00 & 31.21 & 35.45 & 35.79 & 35.48 & 39.24 & 34.00 & 35.58 & 34.73 & 23.36 & 33.88 & 35.55 & 32.36 & 36.36 & 32.67 & 34.15 & 24.27 & 35.64 & 34.45 & 32.27 & 33.61 & 30.09 & 34.21 & 34.58 & 36.76 & 34.88 & 36.67 & 34.88 & 34.88 \\
133 & 89.42 & 90.19 & 90.60 & 90.82 & 89.98 & 90.63 & 90.52 & 90.60 & 90.35 & 89.99 & 90.45 & 90.95 & 90.69 & 90.50 & 90.70 & 89.99 & 90.25 & 90.87 & 90.74 & 90.71 & 90.95 & 90.92 & 89.26 & 90.89 & 90.97 & 90.22 & 90.25 & 90.19 & 90.19 \\
136 & 55.64 & 53.09 & 48.12 & 51.85 & 47.32 & 56.41 & 53.33 & 52.87 & 53.31 & 51.71 & 55.13 & 54.76 & 55.22 & 54.35 & 52.12 & 54.55 & 52.75 & 54.04 & 52.75 & 53.21 & 53.24 & 53.62 & 55.64 & 51.81 & 55.64 & 55.54 & 55.64 & 55.54 & 55.64 \\
137 & 92.54 & 95.52 & 93.98 & 93.88 & 94.60 & 94.20 & 92.54 & 93.13 & 91.64 & 92.31 & 92.89 & 93.76 & 92.76 & 93.58 & 93.86 & 90.17 & 94.15 & 94.73 & 92.74 & 94.10 & 93.68 & 93.33 & 93.33 & 94.13 & 94.45 & 94.23 & 94.23 & 94.23 & 94.23 \\
139 & 88.73 & 89.50 & 98.09 & 99.23 & 95.92 & 98.90 & 98.15 & 98.48 & 75.57 & 89.85 & 98.94 & 98.21 & 98.65 & 97.98 & 98.50 & 99.50 & 67.61 & 99.34 & 99.46 & 98.75 & 99.54 & 97.32 & 97.51 & 98.02 & 99.17 & 99.98 & 98.13 & 99.02 & 98.25 \\
141 & 89.30 & 86.25 & 94.88 & 95.75 & 94.39 & 94.65 & 93.25 & 93.84 & 93.32 & 89.97 & 93.05 & 96.34 & 95.51 & 94.79 & 94.79 & 89.29 & 92.15 & 95.75 & 95.49 & 95.15 & 96.30 & 95.13 & 93.53 & 96.57 & 96.47 & 93.57 & 93.57 & 93.57 & 93.57 \\
142 & 49.15 & 54.58 & 56.14 & 57.02 & 55.82 & 57.51 & 55.89 & 56.29 & 55.03 & 46.87 & 54.46 & 56.72 & 56.61 & 57.29 & 56.47 & 47.71 & 47.25 & 56.27 & 55.66 & 54.78 & 56.88 & 57.27 & 57.31 & 55.21 & 57.47 & 56.47 & 58.08 & 58.21 & 56.47 \\
146 & 96.48 & 96.74 & 96.98 & 96.89 & 96.99 & 97.01 & 96.79 & 96.64 & 97.09 & 96.74 & 96.85 & 96.65 & 96.87 & 96.84 & 96.75 & 96.87 & 96.71 & 96.83 & 97.89 & 96.63 & 96.81 & 96.88 & 97.00 & 96.71 & 96.92 & 96.82 & 96.92 & 97.00 & 96.92 \\
147 & 75.88 & 68.13 & 78.45 & 76.37 & 74.69 & 75.08 & 75.93 & 76.60 & 69.81 & 69.07 & 74.69 & 76.64 & 75.92 & 73.59 & 72.64 & 75.01 & 65.59 & 86.64 & 85.17 & 77.09 & 79.04 & 71.45 & 69.36 & 78.79 & 73.73 & 83.83 & 74.05 & 71.71 & 70.04 \\
149 & 77.14 & 65.95 & 85.17 & 85.17 & 73.82 & 83.97 & 82.40 & 83.40 & 83.81 & 81.51 & 84.90 & 85.55 & 85.48 & 86.18 & 86.02 & - & 87.33 & 86.67 & 86.53 & 82.77 & 86.49 & 84.89 & 66.04 & 87.33 & 82.58 & 83.69 & 75.55 & 77.01 & 75.55 \\
150 & 49.83 & 51.30 & 58.80 & 60.16 & 56.63 & 60.52 & 57.42 & 56.66 & 61.06 & 50.98 & 55.25 & 47.50 & 58.24 & 59.53 & 56.81 & 50.80 & 52.09 & 59.39 & 57.24 & 59.30 & 58.31 & 56.81 & 59.16 & 61.31 & 61.33 & 61.08 & 59.28 & 59.30 & 57.45 \\
153 & 67.87 & 70.98 & 78.17 & 77.96 & 77.95 & 77.82 & 75.80 & 75.40 & 62.25 & 74.69 & 75.54 & 75.88 & 75.61 & 75.56 & 75.28 & 74.67 & 74.43 & 77.55 & 72.98 & 77.23 & 77.82 & 75.78 & 76.93 & 77.43 & 77.65 & 77.44 & 77.44 & 77.44 & 77.44 \\
154 & 72.50 & 75.50 & 76.83 & 77.67 & 76.70 & 74.50 & 73.30 & 75.77 & 78.07 & 68.37 & 75.33 & 79.13 & 74.33 & 78.97 & 77.40 & 75.83 & - & 75.40 & 78.50 & 74.23 & 74.43 & 78.43 & 74.80 & 78.13 & 78.63 & 78.30 & 78.70 & 78.30 & 78.30 \\
155 & 93.43 & 90.23 & 96.07 & 96.31 & 93.36 & 96.06 & 96.00 & 95.99 & 95.57 & 95.43 & 95.73 & 96.28 & 96.10 & 95.82 & 95.82 & 95.53 & 95.05 & 96.34 & 96.17 & 96.06 & 96.08 & 95.95 & 91.32 & 96.44 & 95.71 & 91.65 & 95.58 & 91.65 & 91.65 \\
156 & 60.81 & 63.01 & 64.20 & 64.62 & 62.18 & 64.17 & 62.97 & 63.14 & 63.91 & 62.53 & 63.13 & 63.87 & 64.06 & 64.10 & 64.22 & 62.47 & 63.76 & 63.86 & 63.96 & 63.95 & 63.93 & 63.79 & 61.45 & 63.84 & 63.60 & 62.30 & 62.30 & 62.30 & 62.30 \\
160 & 94.67 & 94.31 & 95.02 & 94.82 & 94.89 & 94.85 & 94.49 & 94.73 & 94.44 & 94.35 & 94.69 & 94.71 & 94.73 & 94.67 & 94.42 & 90.68 & 94.60 & 94.82 & 94.93 & 94.69 & 94.52 & 94.60 & 94.72 & 94.95 & 94.98 & 94.72 & 94.85 & 94.74 & 94.72 \\
162 & 98.41 & 98.41 & 98.76 & 98.66 & 98.41 & 98.70 & 98.52 & 98.41 & 98.41 & 98.37 & 98.35 & 98.76 & 98.48 & 98.39 & 98.50 & 98.41 & 98.40 & 98.76 & 98.68 & 98.42 & 98.87 & 98.58 & 98.41 & 98.71 & 99.00 & 98.56 & 98.41 & 98.49 & 98.43 \\
163 & 89.66 & 96.08 & 96.30 & 96.52 & 94.97 & 96.99 & 92.24 & 96.33 & 90.79 & 81.72 & 93.39 & 93.50 & 95.57 & 95.02 & 95.97 & 95.61 & - & 94.85 & 95.65 & 93.73 & 96.54 & 96.51 & - & 96.70 & 96.76 & 93.33 & 93.33 & 93.33 & 93.33 \\
164 & 40.58 & 39.26 & 38.59 & 38.44 & 39.35 & 39.49 & 38.51 & 37.15 & 40.27 & 37.97 & 35.90 & 39.86 & 39.06 & 36.92 & 39.40 & 38.94 & 36.82 & 39.26 & 39.89 & 37.79 & 38.13 & 39.45 & 40.32 & 39.58 & 40.32 & 40.22 & 40.32 & 40.32 & 40.32 \\
165 & 91.48 & 91.14 & 92.10 & 92.18 & 91.29 & 92.29 & 92.46 & 92.54 & 91.07 & 91.67 & 92.24 & 92.53 & 92.47 & 92.49 & 92.38 & 92.18 & 35.09 & 92.74 & 92.58 & 92.42 & 92.64 & 92.41 & 92.19 & 92.45 & 92.37 & 92.66 & 92.31 & 92.27 & 92.31 \\
166 & 92.22 & 62.48 & 94.49 & 95.68 & 88.41 & 94.60 & 95.69 & 93.61 & 81.95 & 63.12 & 94.56 & 95.35 & 95.26 & 95.64 & 95.73 & 91.70 & 81.56 & 98.19 & 99.11 & 95.42 & 98.95 & 93.67 & 92.84 & 96.90 & 95.49 & 95.68 & 93.88 & 95.14 & 91.04 \\
167 & 84.27 & 76.01 & 92.71 & 91.69 & 85.26 & 92.31 & 85.37 & 85.52 & 83.18 & 81.92 & 84.95 & 87.51 & 86.05 & 85.55 & 85.15 & 79.49 & 82.84 & 96.39 & 95.62 & 86.82 & 87.70 & 87.44 & 79.45 & 88.66 & 84.99 & 88.97 & 84.88 & 87.02 & 78.78 \\
169 & 84.63 & 73.74 & 90.06 & 94.69 & 93.18 & 96.08 & 93.24 & 93.74 & 75.43 & 92.10 & 94.29 & 97.13 & 95.24 & 95.19 & 96.75 & 86.70 & 80.80 & 98.03 & 98.28 & 97.42 & 95.92 & 96.83 & 96.78 & 97.49 & 97.75 & 97.78 & 96.64 & 97.08 & 96.64 \\
170 & 59.46 & 64.19 & 67.66 & 67.70 & 67.66 & 65.72 & 65.81 & 68.83 & 70.90 & 48.29 & 61.85 & 67.12 & 67.43 & 66.71 & 66.13 & 61.44 & - & 68.24 & 70.77 & 68.87 & 67.97 & 69.73 & 69.77 & 68.06 & 72.39 & 68.96 & 70.00 & 68.96 & 68.96 \\
171 & 61.24 & 49.56 & 72.64 & 71.88 & 64.62 & 72.85 & 62.77 & 63.71 & 60.82 & 54.80 & 59.63 & 71.59 & 61.15 & 62.87 & 63.33 & 54.58 & 43.34 & 96.98 & 98.55 & 80.07 & 76.66 & 59.58 & 55.06 & 92.71 & 75.26 & 69.37 & 58.00 & 60.76 & 54.44 \\
172 & 58.02 & 55.58 & 63.25 & 61.52 & 60.57 & 63.93 & 57.08 & 57.54 & 56.11 & 56.45 & 57.45 & 59.45 & 58.31 & 57.41 & 57.64 & 56.73 & 55.79 & 65.95 & 90.88 & 63.30 & 59.63 & 60.28 & 57.84 & 59.87 & 63.16 & 58.45 & 58.04 & 58.04 & 58.04 \\
174 & 59.89 & 48.45 & 61.23 & 61.11 & 61.66 & 60.11 & 57.63 & 58.42 & 43.57 & 50.95 & 57.10 & 57.51 & 57.03 & 57.12 & 57.76 & 51.26 & 55.80 & 58.91 & 59.95 & 56.48 & 57.49 & 55.24 & 57.75 & 56.08 & 58.14 & 59.93 & 57.51 & 58.97 & 54.55 \\
177 & 99.46 & 98.96 & 99.42 & 99.49 & 99.40 & 99.47 & 99.41 & 99.50 & 98.80 & 98.39 & 99.35 & 99.20 & 99.45 & 99.40 & 99.34 & 99.26 & 48.31 & 99.48 & 99.49 & 99.57 & 99.66 & 99.24 & - & 99.63 & 99.48 & 98.18 & 98.18 & 98.18 & 98.18 \\
179 & 84.02 & 90.87 & 92.91 & 91.05 & 91.45 & 91.99 & 88.31 & 89.92 & 91.84 & 89.04 & 88.77 & 92.88 & 94.40 & 89.92 & 94.52 & 84.35 & 88.31 & 93.55 & 91.29 & 90.96 & 90.68 & 91.29 & 90.81 & 93.00 & 93.06 & 90.29 & 91.75 & 90.72 & 90.29 \\
182 & 71.45 & 71.25 & 72.33 & 72.30 & 71.98 & 72.21 & 71.81 & 71.75 & 72.05 & 71.21 & 71.94 & 72.17 & 71.95 & 70.87 & 71.63 & 71.87 & 71.40 & 72.17 & 72.13 & 72.40 & 72.17 & 72.45 & 71.75 & 72.21 & 72.29 & 71.57 & 71.84 & 71.65 & 71.36 \\
183 & 55.56 & 78.68 & 57.78 & 51.66 & 54.98 & 60.27 & 65.49 & 77.64 & 49.19 & 49.82 & 61.26 & 50.81 & 54.87 & 59.53 & 62.69 & 89.44 & 58.41 & 70.04 & 78.11 & 53.44 & 79.64 & 54.54 & 51.63 & 52.67 & 63.43 & 71.47 & 52.59 & 58.96 & 57.67 \\
184 & 86.32 & 81.46 & 88.26 & 88.47 & 87.88 & 88.08 & 87.17 & 86.97 & 87.23 & 85.03 & 86.86 & 87.47 & 87.21 & 86.85 & 86.62 & 86.26 & 86.32 & 87.02 & 87.04 & 86.83 & 87.44 & 87.22 & 87.11 & 87.41 & 87.75 & 87.51 & 87.51 & 87.51 & 87.51 \\
191 & 97.99 & 97.98 & 97.89 & 97.94 & 98.03 & 97.81 & 97.83 & 97.86 & 97.54 & 97.83 & 97.80 & 97.89 & 97.83 & 97.86 & 97.82 & 97.85 & 97.91 & 97.85 & 98.00 & 97.94 & 97.85 & 97.93 & 97.80 & 97.97 & 97.95 & 97.85 & 97.98 & 97.85 & 97.98 \\
192 & 85.03 & 99.95 & 100.0 & 100.0 & 100.0 & 100.0 & 95.06 & 96.17 & 84.49 & 92.79 & 93.67 & 100.0 & 93.65 & 97.80 & 97.23 & 92.27 & 86.12 & 99.98 & 100.0 & 99.93 & 99.84 & 100.0 & 99.86 & 100.0 & 100.0 & 99.95 & 99.34 & 99.16 & 98.66 \\
193 & 69.53 & 67.09 & 77.10 & 76.22 & 74.42 & 76.85 & 76.29 & 76.05 & 72.25 & 67.77 & 75.52 & 75.70 & 74.31 & 74.90 & 75.70 & 76.02 & 73.99 & 76.83 & 75.90 & 75.57 & 76.65 & 74.60 & 64.39 & 77.18 & 75.57 & 74.63 & 70.41 & 70.41 & 70.41 \\
194 & 93.10 & 79.54 & 92.98 & 93.08 & 93.11 & 93.09 & 80.12 & 76.80 & 79.76 & 73.56 & 79.84 & 79.73 & 79.94 & 80.02 & 79.87 & 74.12 & 76.18 & 93.09 & 92.92 & 80.45 & 93.10 & 79.94 & 81.59 & 80.26 & 80.18 & 80.08 & 80.18 & 80.08 & 80.18 \\
196 & 78.39 & 77.05 & 80.29 & 79.72 & 81.42 & 80.89 & 75.02 & 76.64 & 78.22 & 76.68 & 75.31 & 78.92 & 77.98 & 76.87 & 77.54 & - & 75.63 & 80.04 & 78.75 & 78.05 & 76.91 & 78.88 & - & 79.65 & 79.30 & 79.13 & 79.13 & 79.13 & 79.13 \\
197 & 81.67 & 81.29 & 81.44 & 80.84 & 81.11 & 80.89 & 81.25 & 80.95 & 81.31 & 80.94 & 80.78 & 81.41 & 81.35 & 81.09 & 81.00 & 81.31 & 81.24 & 81.65 & 80.90 & 81.28 & 81.04 & 80.86 & 81.32 & 81.53 & 81.27 & 81.78 & 81.81 & 81.25 & 81.14 \\
198 & 88.14 & 68.01 & 86.10 & 86.47 & 82.62 & 86.59 & 95.27 & 86.96 & 85.51 & 87.84 & 92.11 & 97.61 & 95.52 & 96.74 & 90.57 & 71.44 & 64.80 & 98.98 & 99.50 & 99.07 & 99.58 & 97.34 & 82.10 & 94.41 & 96.99 & 95.52 & 85.63 & 82.34 & 82.34 \\
199 & 86.02 & 85.07 & 86.27 & 86.24 & 86.98 & 87.44 & 85.81 & 85.94 & 85.02 & 84.11 & 85.40 & 85.34 & 86.26 & 85.53 & 84.88 & 86.07 & 85.96 & 84.74 & 84.53 & 85.26 & 86.10 & 84.71 & 86.62 & 86.03 & 86.73 & 86.60 & 86.71 & 86.76 & 86.60 \\
200 & 83.72 & 85.81 & 88.36 & 87.74 & 88.27 & 88.70 & 84.96 & 85.16 & 83.28 & 84.69 & 84.89 & 84.87 & 84.58 & 85.07 & 85.07 & 84.96 & 84.23 & 87.60 & 88.14 & 87.87 & 87.73 & 84.82 & 85.99 & 87.41 & 86.03 & 86.96 & 85.98 & 86.63 & 85.98 \\
203 & 93.59 & 97.94 & 99.84 & 99.84 & 99.27 & 99.84 & 99.40 & 99.52 & 99.77 & 96.02 & 99.43 & 99.69 & 99.68 & 99.67 & 99.54 & 99.43 & 99.38 & 99.78 & 99.76 & 99.60 & 99.69 & 99.51 & 99.24 & 99.44 & 99.80 & 99.70 & 99.26 & 99.68 & 99.08 \\
205 & 93.82 & 100.0 & 100.0 & 100.0 & 100.0 & 100.0 & 99.99 & 100.0 & 100.0 & 99.93 & 100.0 & 98.02 & 100.0 & 100.0 & 100.0 & 94.17 & - & 100.0 & 100.0 & 100.0 & 100.0 & 94.39 & 100.0 & 100.0 & 100.0 & 100.0 & 100.0 & 99.98 & 100.0 \\
206 & 72.34 & 73.44 & 73.01 & 73.49 & 73.11 & 72.41 & 73.78 & 73.74 & 73.49 & 70.73 & 72.78 & 73.31 & 73.54 & 72.59 & 73.26 & 73.62 & 73.39 & 73.83 & 73.18 & 72.78 & 73.41 & 73.70 & 73.55 & 73.41 & 73.47 & 73.37 & 73.93 & 73.61 & 73.93 \\
207 & 72.97 & 72.03 & 74.21 & 74.31 & 74.50 & 74.52 & 73.97 & 74.16 & 73.93 & 72.33 & 73.50 & 73.61 & 73.89 & 73.48 & 73.60 & 73.58 & 73.97 & 73.66 & 73.88 & 73.78 & 73.33 & 73.45 & 73.41 & 73.94 & 73.99 & 73.98 & 73.54 & 73.54 & 73.54 \\
209 & 67.83 & 60.35 & 86.32 & 86.43 & 74.12 & 86.38 & 61.52 & 59.38 & 76.45 & 55.30 & 58.56 & 66.91 & 57.31 & 59.17 & 61.41 & 59.52 & 50.91 & 72.40 & 83.32 & 62.48 & 84.67 & 58.27 & - & 64.13 & 82.60 & 59.63 & 59.63 & 59.63 & 59.63 \\
210 & 98.66 & 98.19 & 98.67 & 98.83 & 98.52 & 98.76 & 98.67 & 98.71 & 98.21 & 98.31 & 98.78 & 98.71 & 98.70 & 98.65 & 98.70 & 98.17 & 98.43 & 98.72 & 98.56 & 98.71 & 98.81 & 98.64 & 98.65 & 98.69 & 98.75 & 98.72 & 98.73 & 98.62 & 98.62 \\
211 & 82.76 & 62.56 & 82.13 & 81.28 & 82.04 & 83.32 & 72.41 & 76.29 & 73.92 & 66.40 & 72.91 & 71.23 & 68.01 & 66.77 & 71.66 & 66.83 & 81.05 & 83.25 & 82.99 & 81.25 & 82.46 & 73.60 & 77.50 & 80.43 & 80.76 & 81.81 & 78.46 & 81.25 & 78.46 \\
213 & 96.25 & 97.75 & 96.60 & 96.80 & 96.55 & 96.88 & 96.15 & 97.25 & 93.83 & 91.23 & 96.02 & 96.42 & 95.93 & 95.77 & 96.32 & 96.63 & 95.80 & 96.93 & 97.22 & 96.37 & 97.10 & 96.53 & - & 96.57 & 96.55 & 96.45 & 96.47 & 96.45 & 96.47 \\
214 & 86.50 & 83.50 & 85.72 & 87.18 & 86.02 & 86.23 & 84.45 & 85.97 & 83.30 & 77.42 & 84.07 & 85.23 & 83.90 & 84.67 & 83.88 & 81.88 & 84.25 & 86.53 & 86.48 & 85.23 & 88.08 & 85.55 & 83.63 & 86.72 & 87.73 & 88.48 & 85.70 & 86.27 & 85.27 \\
215 & 96.00 & 94.25 & 94.00 & 95.95 & 94.72 & 95.02 & 95.63 & 96.50 & 93.77 & 84.10 & 95.42 & 95.55 & 95.27 & 95.35 & 94.83 & 94.72 & 94.48 & 95.57 & 96.53 & 95.75 & 96.42 & 94.95 & 94.62 & 96.07 & 96.58 & 96.32 & 95.08 & 96.03 & 95.08 \\
216 & 76.00 & 74.92 & 74.88 & 74.28 & 74.90 & 75.33 & 76.43 & 76.42 & 69.00 & 74.53 & 76.92 & 76.68 & 76.37 & 75.92 & 76.17 & 76.25 & 73.92 & 76.33 & 75.53 & 76.15 & 75.43 & 74.45 & 76.10 & 76.00 & 76.67 & 76.57 & 77.30 & 76.57 & 76.67 \\
217 & 92.25 & 93.75 & 96.13 & 95.20 & 95.90 & 95.92 & 95.47 & 95.65 & 94.55 & 86.85 & 94.52 & 94.90 & 95.85 & 96.00 & 95.60 & - & 96.00 & 96.37 & 96.50 & 95.08 & 95.88 & 95.00 & - & 96.00 & 94.08 & 93.50 & 93.50 & 93.50 & 93.50 \\
218 & 80.50 & 82.38 & 78.07 & 78.55 & 76.03 & 79.45 & 84.77 & 83.62 & 67.93 & 72.63 & 83.47 & 83.33 & 83.17 & 82.98 & 82.50 & 81.72 & 81.03 & 87.27 & 79.02 & 82.18 & 83.03 & 79.77 & 81.58 & 83.90 & 81.97 & 87.67 & 82.32 & 84.03 & 81.80 \\
219 & 100.0 & 98.61 & 96.30 & 99.07 & 97.96 & 98.61 & 99.23 & 99.94 & 98.80 & 85.31 & 99.14 & 98.02 & 98.86 & 99.48 & 99.32 & 99.35 & 99.78 & 99.88 & - & 98.12 & 100.0 & 97.96 & 99.60 & 98.33 & 100.0 & 99.94 & 99.94 & 100.0 & 99.94 \\
220 & 58.63 & 56.80 & 62.03 & 61.89 & 61.12 & 61.92 & 63.00 & 62.93 & 61.16 & 61.42 & 62.20 & 62.76 & 62.18 & 62.90 & 62.22 & 59.67 & 46.39 & 63.28 & 63.08 & 62.77 & 63.79 & 62.59 & 60.60 & 62.39 & 63.70 & 62.13 & 61.53 & 59.62 & 59.62 \\
221 & 96.68 & 83.69 & 98.05 & 98.03 & 97.01 & 97.95 & 98.32 & 98.04 & 96.13 & 97.19 & 98.07 & 98.17 & 98.21 & 98.26 & 98.25 & 88.56 & 83.55 & 98.93 & 98.47 & 98.33 & 98.47 & 98.14 & 93.40 & 98.53 & 98.27 & 99.23 & 98.34 & 96.69 & 93.83 \\
222 & 91.12 & 77.85 & 94.59 & 94.66 & 94.56 & 94.58 & 91.94 & 91.35 & 92.73 & 92.61 & 91.48 & 92.76 & 90.90 & 92.50 & 82.88 & 85.09 & 91.53 & 94.26 & 94.28 & 93.86 & 93.24 & 92.79 & 91.19 & 93.47 & 93.10 & 92.88 & 91.20 & 92.87 & 91.20 \\
224 & 96.56 & 95.81 & 96.80 & 96.95 & 96.34 & 96.97 & 96.87 & 96.94 & 96.19 & 96.55 & 96.93 & 96.91 & 96.92 & 96.81 & 96.54 & 97.00 & 96.96 & 97.02 & 96.90 & 96.98 & 96.93 & 96.40 & 96.31 & 97.03 & 96.86 & 96.95 & 96.17 & 96.17 & 96.17 \\
225 & 78.92 & 98.68 & 100.0 & 100.0 & 100.0 & 100.0 & 94.53 & 96.45 & 99.79 & 98.90 & 94.18 & 99.65 & 99.37 & 98.45 & 98.30 & 91.93 & 90.72 & 99.40 & 99.89 & 99.88 & 99.53 & 97.73 & 99.50 & 100.0 & 100.0 & 98.00 & 98.00 & 99.43 & 98.70 \\
226 & 72.68 & 74.25 & 75.12 & 75.08 & 72.27 & 74.75 & 74.50 & 74.28 & 73.09 & 73.36 & 74.58 & 75.00 & 74.55 & 74.81 & 74.83 & 73.62 & 71.85 & 74.81 & 74.75 & 74.71 & 74.79 & 74.07 & 73.12 & 74.78 & 74.98 & 73.17 & 72.97 & 73.02 & 72.97 \\
230 & 84.59 & 87.79 & 90.51 & 90.63 & 86.33 & 90.19 & 88.03 & 88.71 & 88.75 & 89.08 & 88.64 & 90.19 & 89.76 & 90.32 & 89.69 & 88.68 & 88.29 & 89.65 & 90.53 & 90.21 & 90.11 & 89.98 & 89.69 & 89.49 & 89.89 & 89.38 & 88.66 & 89.94 & 89.06 \\
231 & 97.95 & 95.82 & 98.24 & 98.24 & 98.01 & 98.31 & 98.25 & 98.64 & 95.71 & 94.33 & 98.05 & 97.79 & 97.98 & 98.35 & 98.28 & 98.07 & 98.19 & 98.71 & 99.13 & 97.85 & 98.70 & 96.98 & 97.89 & 98.62 & 98.35 & 98.60 & 97.74 & 98.91 & 97.74 \\
232 & 95.07 & 94.08 & 94.79 & 94.40 & 95.00 & 94.81 & 93.74 & 93.74 & 93.69 & 93.46 & 93.03 & 94.83 & 94.53 & 94.31 & 94.00 & 93.52 & 94.46 & 94.42 & 94.66 & 93.37 & 94.02 & 94.52 & 95.13 & 94.94 & 95.12 & 95.02 & 95.16 & 95.36 & 95.11 \\
233 & 98.23 & 96.65 & 98.14 & 97.81 & 98.23 & 97.87 & 97.43 & 97.68 & 98.12 & 98.06 & 97.26 & 97.53 & 97.57 & 97.47 & 97.52 & 98.08 & - & 97.91 & 97.76 & 96.73 & 97.30 & - & 97.83 & 98.07 & 98.04 & 97.94 & 98.23 & 97.94 & 98.23 \\
234 & 96.16 & 95.65 & 97.45 & 96.67 & 97.03 & 97.12 & 96.69 & 96.68 & 92.95 & 95.27 & 96.94 & 96.71 & 97.11 & 96.63 & 96.61 & 96.40 & 96.38 & 96.77 & 96.68 & 96.85 & 97.11 & 96.32 & 97.27 & 96.74 & 97.23 & 97.13 & 97.23 & 97.55 & 97.23 \\
235& 93.69 & 93.24 & 93.81 & 93.51 & 93.57 & 93.75 & 92.73 & 92.10 & 93.24 & 92.19 & 93.36 & 93.09 & 93.24 & 91.83 & 92.73 & 93.24 & 92.43 & 93.36 & 93.42 & 93.39 & 92.70 & 91.83 & 93.69 & 93.72 & 93.70 & 93.60 & 93.69 & 93.72 & 93.70 \\
236 & 89.14 & 88.71 & 88.56 & 88.80 & 89.41 & 87.73 & 88.20 & 88.67 & 89.78 & 88.84 & 89.07 & 89.78 & 89.52 & 88.67 & 89.29 & 89.76 & 89.63 & 88.46 & 88.75 & 88.24 & 87.92 & 89.69 & 89.50 & 88.60 & 89.48 & 89.38 & 89.65 & 89.38 & 89.39 \\
237 & 89.04 & 90.27 & 91.51 & 90.16 & 89.13 & 91.10 & 89.89 & 88.52 & 87.79 & 87.47 & 90.05 & 89.59 & 90.34 & 89.75 & 90.16 & 89.68 & 89.75 & 89.89 & 89.43 & 89.86 & 89.91 & 89.45 & 90.78 & 90.21 & 90.64 & 90.54 & 90.64 & 90.57 & 90.65 \\
238 & 99.09 & 92.81 & 94.14 & 99.13 & 98.79 & 95.09 & 99.50 & 99.48 & 98.07 & 98.74 & 99.29 & 99.35 & 99.35 & 99.42 & 99.16 & 99.24 & 22.76 & 99.43 & 99.40 & 99.29 & 99.60 & 99.37 & 98.88 & 99.41 & 99.44 & 99.68 & 99.04 & 98.78 & 98.78 \\
239 & 73.86 & 70.47 & 76.73 & 75.44 & 75.79 & 76.03 & 71.87 & 71.16 & 75.27 & 69.87 & 71.01 & 69.81 & 72.48 & 70.27 & 71.59 & 71.18 & 71.97 & 74.11 & 76.81 & 72.31 & 73.82 & 73.03 & - & 73.88 & 74.09 & 72.33 & 72.33 & 72.33 & 72.33 \\
240 & 86.86 & 72.94 & 87.66 & 87.97 & 88.52 & 88.07 & 86.84 & 87.14 & 85.38 & 84.66 & 86.10 & 88.09 & 86.22 & 87.10 & 85.93 & 80.70 & 86.43 & 88.92 & 89.71 & 86.98 & 87.84 & 86.54 & 86.24 & 87.06 & 87.10 & 88.59 & 87.41 & 88.22 & 85.56 \\
241 & 95.34 & 88.15 & 98.24 & 98.60 & 97.32 & 98.18 & 98.54 & 98.85 & 98.08 & 97.70 & 98.53 & 99.14 & 99.05 & 98.82 & 98.78 & 97.58 & 71.42 & 98.87 & 98.85 & 99.04 & 98.59 & 98.63 & 97.48 & 98.96 & 99.21 & 98.47 & 98.22 & 96.92 & 97.11 \\
244 & 67.91 & 73.50 & 75.68 & 77.82 & 74.79 & 76.58 & 74.60 & 74.78 & 74.92 & 71.08 & 74.61 & 75.47 & 74.57 & 75.85 & 75.80 & 75.40 & 73.24 & 74.93 & 73.45 & 74.90 & 76.55 & 75.32 & 74.93 & 74.87 & 77.04 & 75.20 & 75.20 & 75.20 & 76.05 \\
249 & 92.39 & 92.13 & 91.89 & 92.23 & 92.18 & 92.34 & 91.79 & 92.10 & 91.91 & 91.66 & 91.63 & 92.04 & 92.05 & 91.92 & 92.23 & 92.44 & 91.36 & 92.16 & 92.98 & 91.75 & 92.20 & 91.92 & 92.98 & 91.74 & 92.07 & 92.55 & 92.90 & 92.94 & 92.90 \\
250 & 77.57 & 77.05 & 97.15 & 97.11 & 93.67 & 97.41 & 97.41 & 97.66 & 96.88 & 94.85 & 81.80 & 97.65 & 96.95 & 96.96 & 97.30 & 50.79 & 97.64 & 97.95 & 98.02 & 97.79 & 97.61 & 97.74 & 98.04 & 98.06 & 97.98 & 97.88 & 97.98 & 97.95 & 97.98 \\
251 & 65.49 & 63.08 & 79.93 & 78.58 & 72.95 & 78.01 & 65.62 & 65.92 & 64.75 & 62.81 & 66.65 & 71.96 & 68.22 & 64.88 & 64.90 & 62.82 & 62.58 & 87.79 & 83.75 & 75.03 & 74.72 & 71.68 & 67.38 & 76.52 & 80.44 & 73.62 & 67.09 & 67.61 & 66.44 \\
252 & 90.75 & 83.67 & 91.42 & 91.35 & 90.86 & 91.62 & 89.93 & 90.44 & 86.12 & 87.90 & 89.79 & 89.59 & 89.70 & 90.45 & 90.41 & 86.31 & 89.81 & 91.31 & 91.04 & 90.40 & 91.56 & 89.47 & 90.77 & 89.63 & 91.27 & 92.00 & 91.68 & 92.26 & 89.74 \\
254 & 88.96 & 82.97 & 91.49 & 92.02 & 91.96 & 91.77 & 91.99 & 91.86 & 79.57 & 87.92 & 91.85 & 91.37 & 90.39 & 91.88 & 89.06 & 91.39 & 24.04 & 91.90 & 92.77 & 89.86 & 91.33 & 90.59 & 92.34 & 91.18 & 92.62 & 92.52 & 92.61 & 93.26 & 92.61 \\
255 & 93.04 & 93.23 & 92.73 & 92.93 & 93.31 & 93.28 & 93.05 & 92.71 & 93.40 & 92.87 & 92.51 & 92.80 & 92.73 & 92.50 & 92.82 & 92.64 & 92.92 & 92.96 & 93.42 & 92.96 & 92.66 & 92.82 & 93.42 & 92.74 & 93.42 & 93.32 & 93.42 & 93.41 & 93.42 \\
256 & 90.28 & 86.21 & 91.31 & 92.04 & 91.54 & 91.66 & 91.68 & 92.56 & 87.29 & 56.47 & 91.66 & 90.80 & 90.70 & 92.50 & 90.89 & 91.45 & 92.79 & 91.68 & 92.20 & 89.78 & 93.88 & 90.80 & - & 92.29 & 90.87 & 90.39 & 90.39 & 90.39 & 90.39 \\
258 & 90.59 & 86.96 & 90.00 & 90.21 & 90.26 & 89.88 & 90.92 & 90.93 & - & 90.78 & 90.89 & 90.66 & 90.69 & 90.82 & 90.95 & 89.51 & 90.55 & 91.07 & 90.78 & 90.73 & 90.79 & 90.29 & 89.82 & 90.74 & 90.86 & 90.84 & 90.11 & 90.41 & 89.99 \\
259 & 83.00 & 83.90 & 86.68 & 86.69 & 86.40 & 86.80 & 85.72 & 86.18 & 86.73 & 85.27 & 85.94 & 86.55 & 86.37 & 86.09 & 86.12 & 84.78 & 85.52 & 86.74 & 86.63 & 86.15 & 86.28 & 86.27 & 86.12 & 86.75 & 87.02 & 86.03 & 86.29 & 85.42 & 85.42 \\
260 & 99.91 & 92.68 & 99.98 & 99.97 & 99.97 & 85.39 & 99.91 & 99.91 & 99.97 & 99.73 & 99.93 & 99.96 & 99.96 & 99.89 & 99.92 & 99.62 & 94.77 & 99.94 & 99.96 & 99.95 & 99.96 & 99.95 & - & 99.96 & 99.95 & 99.87 & 99.88 & 99.56 & 99.56 \\
263 & 91.75 & 91.91 & 95.23 & 94.67 & 93.30 & 94.61 & 93.86 & 93.91 & 93.99 & 91.34 & 93.74 & 93.93 & 93.56 & 93.83 & 92.96 & 93.81 & 93.23 & 93.88 & 93.20 & 93.04 & 94.35 & 93.46 & 94.73 & 93.40 & 94.77 & 94.67 & 94.77 & 94.77 & 94.77 \\
264 & 86.83 & 94.04 & 96.17 & 95.29 & 95.51 & 96.31 & 93.41 & 94.12 & 77.30 & 86.73 & 68.06 & 94.93 & 93.05 & 94.56 & 94.90 & 94.87 & 51.46 & 93.34 & 94.37 & 92.22 & 95.78 & 94.87 & 85.50 & 95.60 & 95.62 & 93.93 & 93.93 & 93.93 & 93.93 \\
265 & 83.00 & 83.13 & 82.40 & 83.40 & 82.37 & 83.77 & 81.27 & 82.40 & 70.17 & 77.10 & 83.43 & 83.30 & 82.57 & 82.23 & 82.17 & 82.93 & 83.27 & 82.53 & 88.43 & 83.73 & 83.53 & 80.53 & 84.00 & 83.30 & 84.13 & 84.20 & 83.97 & 84.17 & 83.97 \\
266 & 71.00 & 75.00 & 74.03 & 74.30 & 72.97 & 75.13 & 72.20 & 70.30 & 72.27 & 68.73 & 69.73 & 71.23 & 72.50 & 71.23 & 71.57 & 70.77 & 68.60 & 72.40 & 72.50 & 70.73 & 71.57 & 73.03 & 73.10 & 72.40 & 72.27 & 73.30 & 70.77 & 72.77 & 70.90 \\
268 & 77.12 & 72.41 & 79.71 & 79.11 & 77.77 & 76.90 & 75.17 & 76.35 & 57.48 & 72.44 & 74.96 & 76.32 & 73.64 & 67.69 & 75.65 & 74.76 & 28.76 & 78.63 & 78.11 & 74.67 & 77.53 & 75.39 & 76.28 & 77.99 & 78.61 & 76.37 & 76.37 & 76.37 & 76.75 \\
272 & 80.32 & 80.59 & 83.99 & 82.92 & 83.08 & 82.68 & 81.39 & 82.68 & 83.21 & 78.77 & 81.50 & 81.04 & 79.54 & 83.78 & 80.72 & 81.98 & 80.29 & 81.90 & 82.81 & 83.64 & 82.33 & 80.59 & 82.52 & 83.78 & 83.16 & 82.41 & 82.38 & 82.38 & 82.38 \\
273 & 88.20 & 91.61 & 94.54 & 95.24 & 93.16 & 94.24 & 93.09 & 93.29 & 91.54 & 92.31 & 93.29 & 94.71 & 92.70 & 93.57 & 89.94 & 93.50 & 91.61 & 96.38 & 95.88 & 94.63 & 95.04 & 93.30 & 93.20 & 94.87 & 95.62 & 93.39 & 93.42 & 93.40 & 92.75 \\
274 & 97.14 & 96.77 & 97.02 & 97.18 & 97.17 & 97.12 & 96.85 & 96.67 & 96.78 & 96.81 & 96.86 & 96.74 & 96.74 & 96.62 & 96.64 & 96.86 & 96.60 & 96.78 & 96.74 & 96.62 & 96.67 & 96.69 & 96.92 & 96.90 & 96.98 & 96.88 & 96.98 & 96.92 & 96.97 \\
275 & 77.15 & 79.59 & 80.64 & 80.57 & 79.65 & 80.60 & 79.36 & 79.56 & 79.88 & 78.54 & 78.90 & 80.05 & 79.82 & 80.02 & 79.57 & 78.20 & 79.02 & 80.13 & 79.69 & 80.31 & 79.95 & 80.05 & 79.70 & 80.21 & 80.25 & 79.90 & 79.96 & 80.26 & 79.69 \\
277 & 95.00 & 93.82 & 99.52 & 99.49 & 99.64 & 99.53 & 98.67 & 98.50 & 97.56 & 98.44 & 98.42 & 98.94 & 98.87 & 98.40 & 98.28 & 93.47 & 97.13 & 98.96 & 99.21 & 99.10 & 99.48 & 99.15 & 98.56 & 99.18 & 99.30 & 98.59 & 98.62 & 98.44 & 98.44 \\
278 & 95.63 & 95.71 & 99.15 & 99.52 & 99.29 & 99.74 & 98.42 & 98.08 & 93.53 & 97.43 & 98.14 & 98.58 & 98.47 & 98.07 & 98.24 & 98.25 & 96.12 & 98.61 & 98.73 & 98.71 & 99.19 & 98.58 & 98.23 & 99.43 & 99.34 & 98.36 & 98.36 & 98.36 & 98.36 \\
279 & 67.14 & 69.95 & 68.93 & 69.24 & 69.29 & 69.69 & 68.68 & 68.76 & 63.50 & 66.35 & 68.93 & 69.32 & 67.00 & 69.24 & 69.60 & 68.98 & 68.75 & 67.39 & 66.92 & 68.83 & 69.06 & 69.38 & 69.06 & 69.06 & 69.10 & 69.68 & 69.76 & 69.17 & 69.60 \\
282 & 49.14 & 46.96 & 50.66 & 51.70 & 51.46 & 51.47 & 50.70 & 50.90 & 47.51 & 48.51 & 50.43 & 51.53 & 49.80 & 50.70 & 51.28 & 46.92 & 47.50 & 50.85 & 52.43 & 51.00 & 51.63 & 50.33 & 49.94 & 50.96 & 52.38 & 50.36 & 50.36 & 50.36 & 50.36 \\
283 & 97.97 & 97.91 & 97.63 & 97.74 & 97.45 & 97.57 & 97.73 & 97.45 & 97.91 & 96.29 & 97.83 & 97.82 & 97.62 & 96.76 & 97.81 & 50.40 & 64.04 & 97.62 & 97.78 & 97.82 & 97.85 & 97.40 & 97.91 & 97.77 & 97.78 & 97.68 & 97.90 & 97.98 & 97.90 \\
285 & 72.94 & 75.96 & 76.04 & 75.29 & 72.78 & 78.12 & 82.35 & 84.94 & 78.20 & 65.84 & 80.43 & 78.24 & 80.63 & 84.39 & 81.10 & 81.76 & 75.76 & 85.76 & 80.24 & 81.33 & 82.31 & 77.84 & 84.35 & 80.90 & 84.47 & 84.71 & 84.71 & 84.63 & 84.47 \\
288 & 92.49 & 68.68 & 99.58 & 99.65 & 99.30 & 99.82 & 93.41 & 93.94 & 90.96 & 92.02 & 92.74 & 97.97 & 95.36 & 92.82 & 92.72 & 92.23 & 90.33 & 97.73 & 98.77 & 98.89 & 99.30 & 98.39 & 94.94 & 99.22 & 99.12 & 94.44 & 94.44 & 95.74 & 94.44 \\
289 & 90.56 & 89.50 & 90.25 & 89.98 & 90.01 & 90.41 & 90.78 & 91.05 & 90.06 & 89.21 & 90.72 & 90.13 & 90.40 & 90.56 & 90.25 & 90.52 & 90.33 & 90.50 & 89.90 & 90.72 & 90.92 & 89.78 & 89.56 & 89.60 & 90.45 & 90.50 & 90.25 & 89.71 & 89.23 \\
290 & 85.10 & 86.28 & 85.41 & 85.96 & 85.10 & 84.95 & 85.57 & 86.23 & 86.25 & 82.67 & 86.02 & 86.11 & 85.45 & 85.88 & 85.32 & 85.54 & 40.77 & 86.04 & 86.24 & 86.24 & 86.70 & 86.09 & 85.91 & 86.29 & 86.50 & 85.96 & 85.96 & 86.65 & 85.96 \\
291 & 84.10 & 86.20 & 83.69 & 84.38 & 82.91 & 81.73 & 85.87 & 86.38 & 85.69 & 80.36 & 85.69 & 86.42 & 85.57 & 85.61 & 85.67 & 85.57 & 85.52 & 86.26 & 87.02 & 85.51 & 86.27 & 85.04 & 86.53 & 85.59 & 86.09 & 85.99 & 86.77 & 86.35 & 86.77 \\
293 & 86.72 & 86.72 & 89.86 & 90.43 & 89.42 & 88.17 & 87.31 & 88.73 & 89.62 & 83.91 & 88.95 & 90.14 & 86.77 & 90.87 & 86.81 & 89.72 & 87.45 & 89.50 & 87.53 & 88.14 & 89.64 & 87.23 & 89.94 & 86.96 & 90.99 & 90.89 & 91.00 & 90.89 & 91.00 \\
295 & 77.10 & 70.45 & 75.00 & 75.34 & 73.86 & 73.39 & 72.68 & 72.59 & 75.17 & 72.04 & 71.90 & 72.34 & 71.62 & 72.11 & 72.05 & 68.28 & 74.48 & 73.93 & 74.82 & 72.98 & 72.90 & 72.20 & 73.41 & 71.81 & 73.96 & 73.86 & 73.75 & 73.59 & 73.59 \\
297 & 65.62 & 54.81 & 63.62 & 60.73 & 63.29 & 63.60 & 58.67 & 58.08 & 58.23 & 54.69 & 50.17 & 58.19 & 55.67 & 56.17 & 58.52 & 51.92 & 61.75 & 61.27 & 63.62 & 60.65 & 58.75 & 55.15 & 58.13 & 58.29 & 60.52 & 58.67 & 59.54 & 59.25 & 57.81 \\
298 & 63.37 & 53.06 & 62.42 & 63.15 & 62.35 & 62.95 & 57.79 & 57.90 & 54.30 & 49.56 & 54.98 & 55.17 & 60.93 & 55.86 & 58.84 & 53.08 & 31.65 & 62.65 & 63.27 & 59.42 & 60.81 & 53.27 & 57.95 & 53.49 & 60.56 & 63.31 & 59.80 & 62.52 & 57.25 \\
299 & 59.60 & 58.59 & 59.84 & 60.83 & 62.47 & 58.25 & 60.31 & 60.34 & 59.57 & 52.55 & 59.12 & 59.82 & 59.06 & 59.28 & 57.53 & 59.19 & 16.66 & 59.84 & 60.34 & 59.87 & 59.26 & 56.72 & 59.57 & 60.88 & 61.53 & 60.36 & 60.36 & 60.36 & 60.36 \\
\bottomrule
\end{longtable}
}}

\begin{table}[h]
\caption{Classification accuracy (mean $\pm$ standard deviation) across 20 high-dimensional datasets. Each method is evaluated over 5 splits and 3 random seeds, and the best-performing method for each dataset is highlighted. ``MNCA'', ``XGB'', ``RF'', ``PG'' denote ``ModernNCA'', ``XGBoost'', ``RandomForest'', ``ProtoGate'', respectively.}
\label{tab:performance_subscript}
\small{
\begin{tabular}{lcccccccc}
\toprule
Dataset & \name & MNCA & TabM & MLP & KNN & XGB & RF & PG \\
\midrule
CLL\_SUB\_111 & 69.86$_{\pm 6.24}$ & 62.90$_{\pm 9.10}$ & 72.17$_{\pm 3.82}$ & 72.46$_{\pm 6.28}$ & 57.39$_{\pm 6.96}$ & \textbf{73.04$_{\pm 7.65}$} & 70.14$_{\pm 6.32}$ & 65.51$_{\pm 7.69}$ \\
BASEHOCK & 96.56$_{\pm 0.77}$ & 96.31$_{\pm 0.83}$ & 96.89$_{\pm 0.81}$ & \textbf{97.01$_{\pm 0.51}$} & 71.88$_{\pm 4.39}$ & 95.29$_{\pm 0.79}$ & 96.73$_{\pm 1.14}$ & 96.32$_{\pm 2.70}$ \\
Prostate\_GE & 88.89$_{\pm 6.43}$ & 81.27$_{\pm 8.60}$ & \textbf{90.16$_{\pm 5.35}$} & 86.98$_{\pm 6.61}$ & 80.00$_{\pm 3.56}$ & 89.52$_{\pm 6.07}$ & 87.94$_{\pm 7.75}$ & 84.13$_{\pm 6.19}$ \\
PCMAC & 86.77$_{\pm 1.26}$ & 88.21$_{\pm 1.44}$ & 88.02$_{\pm 1.13}$ & 88.53$_{\pm 1.26}$ & 66.48$_{\pm 5.23}$ & 91.64$_{\pm 1.45}$ & \textbf{92.20$_{\pm 1.09}$} & 88.21$_{\pm 3.79}$ \\
GLI\_85 & \textbf{86.67$_{\pm 9.48}$} & 81.57$_{\pm 8.00}$ & 79.22$_{\pm 6.02}$ & 85.49$_{\pm 7.08}$ & 76.47$_{\pm 10.52}$ & 82.35$_{\pm 6.08}$ & 83.92$_{\pm 7.58}$ & 81.96$_{\pm 9.48}$ \\
RELATHE & 90.19$_{\pm 1.45}$ & 88.18$_{\pm 1.42}$ & 89.91$_{\pm 1.29}$ & \textbf{90.54$_{\pm 1.70}$} & 75.03$_{\pm 4.75}$ & 87.11$_{\pm 1.19}$ & 87.30$_{\pm 1.28}$ & 89.92$_{\pm 2.50}$ \\
SMK\_CAN\_187 & \textbf{72.28$_{\pm 7.05}$} & 63.51$_{\pm 6.36}$ & 67.02$_{\pm 6.99}$ & 66.84$_{\pm 8.64}$ & 69.47$_{\pm 5.42}$ & 66.49$_{\pm 6.62}$ & 70.70$_{\pm 4.17}$ & 70.71$_{\pm 5.13}$ \\
warpPIE10P & \textbf{100.00$_{\pm 0.00}$} & 98.41$_{\pm 1.66}$ & \textbf{100.00$_{\pm 0.00}$} & 99.05$_{\pm 1.17}$ & 92.38$_{\pm 3.81}$ & 94.92$_{\pm 3.46}$ & 98.57$_{\pm 1.17}$ & 97.79$_{\pm 6.68}$ \\
leukemia & 95.11$_{\pm 3.82}$ & 90.22$_{\pm 5.90}$ & 91.11$_{\pm 4.66}$ & 95.11$_{\pm 2.95}$ & 86.67$_{\pm 5.96}$ & \textbf{97.78$_{\pm 3.14}$} & 92.00$_{\pm 2.67}$ & 94.00$_{\pm 10.55}$ \\
orlraws10P & 96.67$_{\pm 3.50}$ & 97.67$_{\pm 2.49}$ & \textbf{99.00$_{\pm 2.00}$} & 98.33$_{\pm 2.36}$ & 92.0$_{\pm 2.45}$ & 84.33$_{\pm 6.80}$ & \textbf{99.00$_{\pm 2.00}$} & 92.67$_{\pm 5.73}$ \\
GLIOMA & \textbf{72.67$_{\pm 9.29}$} & 58.00$_{\pm 11.08}$ & 62.00$_{\pm 11.08}$ & 60.67$_{\pm 12.36}$ & 68.00$_{\pm 9.80}$ & 66.67$_{\pm 8.69}$ & 64.00$_{\pm 14.51}$ & 69.91$_{\pm 15.78}$ \\
lung\_discrete & 72.00$_{\pm 10.10}$ & 68.00$_{\pm 8.84}$ & 76.00$_{\pm 10.83}$ & 71.56$_{\pm 10.46}$ & 73.33$_{\pm 5.96}$ & \textbf{77.33$_{\pm 9.04}$} & 71.56$_{\pm 8.60}$ & 70.01$_{\pm 7.38}$ \\
warpAR10P & 91.54$_{\pm 7.05}$ & 83.08$_{\pm 6.71}$ & \textbf{96.15$_{\pm 4.21}$} & 85.64$_{\pm 6.95}$ & 53.08$_{\pm 6.15}$ & 81.28$_{\pm 9.09}$ & 87.18$_{\pm 5.73}$ & 90.04$_{\pm 13.95}$ \\
TOX\_171 & \textbf{91.24$_{\pm 3.53}$} & 76.00$_{\pm 6.33}$ & 87.81$_{\pm 3.04}$ & 88.19$_{\pm 4.53}$ & 70.86$_{\pm 5.54}$ & 78.10$_{\pm 6.40}$ & 78.67$_{\pm 5.10}$ & 85.52$_{\pm 7.41}$ \\
lung & \textbf{95.77$_{\pm 1.08}$} & 91.54$_{\pm 3.43}$ & 93.01$_{\pm 2.8}$ & 95.45$_{\pm 1.22}$ & 93.66$_{\pm 1.95}$ & 93.66$_{\pm 2.64}$ & 92.68$_{\pm 2.52}$ & 95.43$_{\pm 4.13}$ \\
ALLAML & \textbf{96.89$_{\pm 4.12}$} & 87.56$_{\pm 6.38}$ & 92.00$_{\pm 2.67}$ & 95.56$_{\pm 3.14}$ & 81.33$_{\pm 7.77}$ & 96.00$_{\pm 4.07}$ & 96.44$_{\pm 3.33}$ & 91.14$_{\pm 10.00}$ \\
colon & \textbf{85.13$_{\pm 5.94}$} & 78.46$_{\pm 11.65}$ & 82.56$_{\pm 7.14}$ & 78.97$_{\pm 9.09}$ & 76.92$_{\pm 12.87}$ & 74.87$_{\pm 8.17}$ & 82.56$_{\pm 6.57}$ & 78.46$_{\pm 8.04}$ \\
gisette & 97.32$_{\pm 0.21}$ & - & 97.15$_{\pm 0.23}$ & \textbf{97.57$_{\pm 0.32}$} & 95.04$_{\pm 0.38}$ & 97.55$_{\pm 0.54}$ & 96.82$_{\pm 0.61}$ & 97.18$_{\pm 0.72}$ \\
madelon & 58.29$_{\pm 2.11}$ & 64.97$_{\pm 5.79}$ & 56.97$_{\pm 2.44}$ & 56.44$_{\pm 1.82}$ & 53.58$_{\pm 2.74}$ & \textbf{75.77$_{\pm 2.53}$} & 70.42$_{\pm 2.30}$ & 72.13$_{\pm 5.54}$ \\
arcene & 86.67$_{\pm 4.05}$ & 81.67$_{\pm 3.84}$ & 78.83$_{\pm 6.38}$ & 85.50$_{\pm 3.89}$ & 84.50$_{\pm 3.32}$ & 75.00$_{\pm 8.37}$ & \textbf{86.83$_{\pm 5.44}$} & 85.33$_{\pm 4.82}$ \\
\bottomrule
\end{tabular}
}
\end{table}

\end{document}